\documentclass[10pt,twocolumn,letterpaper]{article}
\newcommand{\ARXIV}[2]{#1} %
\usepackage[pagenumbers]{templates/3DV/cvpr} %

\usepackage[dvipsnames]{xcolor}

\usepackage[pagebackref,breaklinks,colorlinks]{hyperref}

\def\paperID{19} %
\def\confName{3DV\xspace}
\def\confYear{2024\xspace}

\PassOptionsToPackage{hyphens}{url}
\hypersetup{pagebackref=true,breaklinks=true,colorlinks,bookmarks=false}

\usepackage{times}
\usepackage{bm}
\usepackage{epsfig}
\usepackage{amsmath}
\usepackage{amssymb}
\usepackage{times}
\usepackage{graphicx}
\usepackage{mathtools}
\usepackage{amssymb}
\usepackage{booktabs}
\usepackage{multirow}
\usepackage{tabularx}
\usepackage{xcolor}
\usepackage{lipsum}
\usepackage{soul}
\usepackage{xfrac}
\usepackage{filecontents}

\usepackage{balance}
\usepackage{placeins}
\usepackage{subcaption}
\usepackage{enumitem}
\usepackage{arydshln}
\usepackage{algorithm}
\usepackage{algpseudocode}
\usepackage{tikz}
\usepackage{pgfplots}
\usepackage{styles}
\usepackage{microtype}
\usetikzlibrary{shapes, plotmarks, positioning, calc, intersections, arrows.meta, patterns, patterns.meta}

\usepackage[hang,flushmargin]{footmisc}
\usepackage[accsupp]{axessibility}  %
\tikzset{
  hatch distance/.store in=\hatchdistance,
  hatch distance=10pt,
  hatch thickness/.store in=\hatchthickness,
  hatch thickness=2pt
}

\pgfdeclarepattern{name=north west lines loose,
  bottom left={\pgfpoint{-1pt}{-1pt}},
  top right={\pgfpoint{100pt}{100pt}},
  tile size={\pgfpoint{\hatchdistance}{\hatchdistance}},
  parameters={\hatchdistance,\hatchthickness},
  code={
    \pgfsetlinewidth{\hatchthickness}
    \pgfpathmoveto{\pgfpoint{0pt}{0pt}}
    \pgfpathlineto{\pgfpoint{100.2pt}{100.2pt}}
    \pgfusepath{stroke}
  }
}

\begin{document}

\title{\TITLE}

\author{
    Damien Robert\textsuperscript{1, 2}\\
    {\tt\small damien.robert@ign.fr}
    \and
    Hugo Raguet\textsuperscript{3}\\
    {\tt\small hugo.raguet@insa-cvl.fr}
    \and
    Loic Landrieu\textsuperscript{2,4}\\
    {\tt\small loic.landrieu@enpc.fr}
    \and
    {\textsuperscript{1}CSAI, ENGIE Lab CRIGEN, France}\\
    {\textsuperscript{2} LASTIG, IGN, ENSG, Univ Gustave Eiffel, France}\\
    {\textsuperscript{3}INSA Centre Val-de-Loire Univ de Tours, LIFAT, France}\\
    {{\textsuperscript{4}LIGM, Ecole des Ponts, Univ Gustave Eiffel, CNRS, France}}\\
}

\maketitle

\begin{abstract}
We introduce a highly efficient method for panoptic segmentation of large 3D point clouds by redefining  this task as a scalable graph clustering problem.
This approach can be trained using only local auxiliary tasks, thereby eliminating the resource-intensive instance-matching step during training.
Moreover, our formulation can easily be adapted to the superpoint paradigm, further increasing its efficiency. This allows our model to process scenes with millions of points and thousands of objects in a single inference. 
Our method, called SuperCluster, achieves a new state-of-the-art panoptic segmentation performance for two indoor scanning datasets: 
$50.1$ PQ ($+7.8$) for S3DIS Area~5, 
and $58.7$ PQ ($+25.2$) for ScanNetV2.
We also set the first state-of-the-art for two large-scale mobile mapping benchmarks: KITTI-360 and DALES.
With only $209$k parameters, our model is over $30$ times smaller than the best-competing method and trains up to $15$ times faster.
Our code and pretrained models are available at \GITHUB.
\end{abstract}
\setlength{\parskip}{-0.13em}

\section{Introduction}
Understanding large-scale 3D environments is pivotal for numerous high-impact applications such as the creation of ``digital twins'' of extensive industrial facilities \cite{quattrocchi2022panoptic,niederer2021scaling,jiang2021industrial} or even the digitization of entire cities \cite{lafarge2012creating,xue2020lidar,nochta2021socio}.
Extensive and comprehensive 3D analysis methods also benefit large-scale geospatial analysis, \eg for land \cite{yan2015urban,smeeckaert2013large} or forest surveys \cite{xu2021lidar,hauglin2021large}, as well as building inventory \cite{wang2020urban} for country-scale mapping.
These problems call for scalable models that can process large point clouds with millions of 3D points, accurately predict the semantics of each point, and recover all instances of specific objects, a task referred to as 3D panoptic segmentation \cite{kirillov2019panoptic}.

Most existing 3D panoptic segmentation methods focus on sparse LiDAR scans for autonomous navigation \cite{aygun20214d,fong2022panoptic,zhou2021panoptic}.
Given the relevance of large-scale analysis for industry and practitioners, there is surprisingly little work on large-scale 3D panoptic segmentation \cite{xiang2023review}.
Although they contain non-overlapping instance labels, S3DIS \cite{armeni20163d} and ScanNet \cite{dai2017scannet} only have a few panoptic segmentation entries, and KITTI-360 \cite{liao2022kitti360} and DALES \cite{singer2021dalesobjects} currently have none.

\begin{figure}
    \centering
    \includegraphics[width=\linewidth]{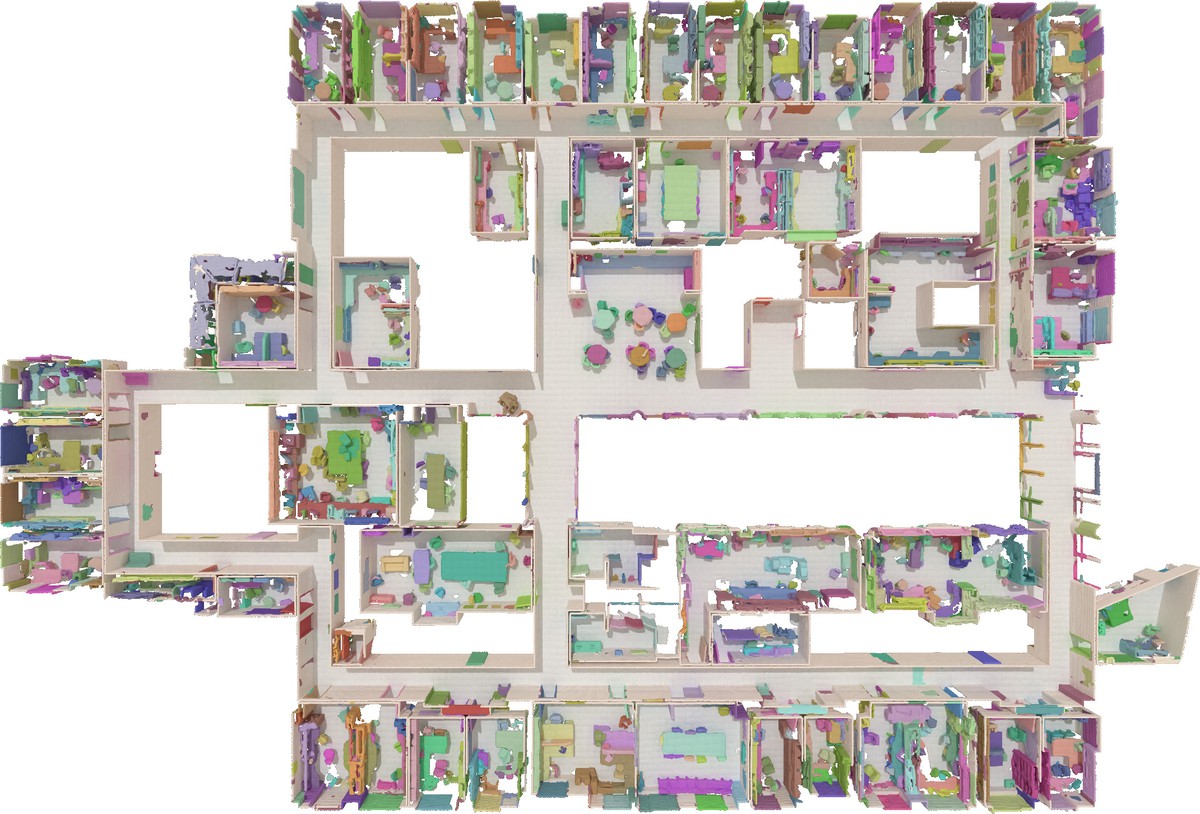}
    \caption{{\bf Large-Scale Panoptic Segmentation.} 
    We present the results of SuperCluster for the entire Area 5 of S3DIS \cite{armeni20193d} (ceiling removed for visualization) with $9.2$M points ($78$M pre-subsampling) and $1863$ true ``things'' objects. 
    Our model can process such large scan in one inference on a single V100-32GB GPU in $3.3$ seconds and reach a state-of-the-art
    PQ of $50.1$.
    }
    \label{fig:teaser}
\end{figure}

Large-scale 3D panoptic segmentation is particularly challenging due to
the scale of scenes, often featuring millions of 3D points, and the diversity in objects---ranging from a few to thousands and with extreme size variability.
Current methods typically rely on large backbone networks with millions of parameters, restricting their analysis to small scenes or portions of scenes due to their high memory consumption.
Furthermore, training these models requires resource-intensive procedures, such as non-maximum suppression and instance matching. These costly operations prevent the analysis of large scenes with many points or objects.
Most methods also require a pre-set limit on the number of detectable objects, introducing unnecessary complexity and the risk of missing objects in large scenes. 
Although recent mask-based instance segmentation methods \cite{schult2023mask3d} have demonstrated high performance and versatility, they fail to scale effectively to large scenes, as they predict a binary mask that covers the entire scene for each proposed instance.

To address these limitations, we present Super-Cluster, a novel approach for large-scale and efficient 3D panoptic segmentation. Our model differs from existing methods in three main ways:
\begin{itemize}
     \item {\bf Scalable graph clustering: } We view the panoptic segmentation task as a scalable graph clustering problem, which can be resolved efficiently at a large scale without setting the number of predicted objects in advance.
    \item {\bf Local supervision: } We use a neural network to predict the parameters of the graph clustering problem and supervise with auxiliary losses that do not require an actual segmentation. This allows us to avoid resource-intensive non-maximum suppression or instance matching steps.
    \item {\bf Superpoint-only segmentation: } Our approach can easily be adapted to a superpoint-based approach. Feature computation,  supervision, and prediction are conducted entirely at the superpoint level and never individual points, starkly decreasing their complexity. %
\end{itemize}
These features make SuperCluster particularly resource-efficient, fast, and scalable, while ensuring high precision, as shown in \figref{fig:teaser}. %
Our primary contributions are as follows:
\begin{itemize}
    \item {\bf Large-scale panoptic segmentation: }
    SuperCluster significantly improves the panoptic segmentation state-of-the-art for two indoor scanning datasets: 
    $50.1$ PQ (+$7.8$) on S3DIS Fold5~\cite{armeni20193d}, 
    and $58.7$ PQ (+$25.2$) on ScanNetV2~\cite{dai2017scannet}. 
    We also set the first panoptic state-of-the-art for S3DIS $6$-fold and two large-scale benchmarks (KITTI-360~\cite{liao2022kitti360} and DALES~\cite{singer2021dalesobjects}).
    \item {\bf Fast and scalable segmentation: }
    SuperCluster contains only $209$k trainable parameters ($205$k in the backbone), yet outperforms networks over $30$ times larger.
    SuperCluster inference is also as fast as the fastest instance segmentation methods and trains up to $15$ times faster: $4$~h for one S3DIS fold and $6$~h for ScanNet.
\end{itemize}

\section{Related Work}
The panoptic segmentation of point clouds with millions of points has received little attention from the 3D computer vision community.
In this paper, we aim to address this gap.

\paragraph{3D Panoptic and Instance Segmentation.}

Over the last few years, deep learning approaches for 3D point clouds have garnered considerable interest \cite{guo2020deep}.
Autonomous driving, in particular, has been the focus of numerous studies, resulting in multiple proposed approaches for object detection \cite{zamanakos2021comprehensive,alaba2022survey}, as well as semantic \cite{zhu2021cylindrical,loiseau2022online,zhang2020polarnet}, instance \cite{zhou2020joint, zhao2021technical} and panoptic segmentation \cite{aygun20214d,fong2022panoptic,zhou2021panoptic,mohan2020efficientps}.
However, these methods consider sequences of sparse LiDAR acquisition, and focus on a small set of classes (pedestrians, cars).

For the panoptic segmentation of dense LiDAR point clouds, the volume of research is surprisingly small \cite{xiang2023review}.
A limited number of studies have addressed the panoptic segmentation of indoor spaces using RGB-D images \cite{wu2021scenegraphfusion,narita2019panopticfusion}.
Dense scans have primarily been used in the context of instance segmentation \cite{schult2023mask3d,ngo2023isbnet,He2021dyco3d,yang2019learning,jiang2020pointgroup,vu2022softgroup}.
However, while this task is related to panoptic segmentation, these methods often adopt specific strategies to maximize instance segmentation metrics \cite{xiang2023review,cheng2021mask2former}.
Moreover, many methods require specifying the maximum number of predicted instances in advance, a constraint that proves inefficient for small scenes and results in missing objects in large scenes.
Additionally, when implementing a sliding-window strategy, the predicted instances must be stitched together using either heuristic techniques or resource-intensive post-processing.
Lastly, the best-performing methods \cite{schult2023mask3d,ngo2023isbnet} rely on a computationally expensive matching step between the predicted and true instances \cite{carion2020end, yang2019learning,jia2023detrs}.
This process often depends on the Hungarian algorithm, which has cubic complexity in the number of objects and, therefore, cannot scale to large scenes.

\paragraph{Superpoint-Based 3D Analysis.}
The strategy of partitioning large 3D point clouds into groups of adjacent and homogeneous points, called superpoints, has been used successfully for point cloud oversegmentation \cite{landrieu2019ssp,papon2013voxel,lin2018toward}, 
semantic segmentation \cite{landrieu2018spg,robert2023efficient,hui2021superpoint}, and object detection \cite{han2020occuseg,engelmann20203d}.
Our approach shares similarities with some superpoint-based approaches for 3D instance segmentation \cite{sun2023superpoint,liang2021instance}.
However, these methods are limited in scalability due to their reliance on point-wise encoders.
Furthermore, the work by Sun \etal \cite{sun2023superpoint} employs a Hungarian-type instance matching scheme and allocates a binary mask to each predicted instance, covering the entire scene and drastically limiting the number of detected instances.
Liang \etal \cite{liang2021instance} resort to quadratic-complexity agglomerative clustering to merge superpoints, and heavy postprocessing to refine and score superpoints.
In contrast, our method employs a fast graph clustering approach \cite{landrieu2017cut,kolmogorov2004energy}, which does not require any instance matching or post-processing steps.

\section{Method}
\begin{figure*}
    \centering
    \scriptsize
    \input{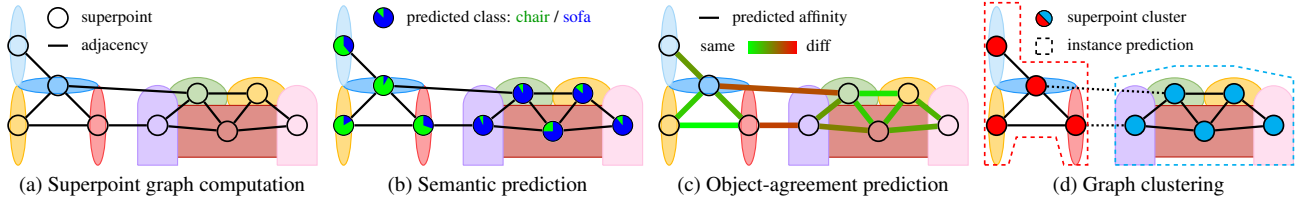}
    \caption{{\bf SuperCluster.}
        We illustrate the sequence of operations of SuperCluster for a simplified scene with two objects: a chair and a sofa.
        \textbf{Sub-figure (a)} showcases the first stage of our process, where the point cloud is partitioned into connected \emph{superpoints} with simple geometric shapes.
        In \textbf{Sub-figure (b)}, we predict a semantic class distribution for each superpoint. 
        In \textbf{Sub-figure (c)}, we predict the object agreement for each pair of adjacent superpoints, indicating the likelihood that they belong to the same object.
        Finally, \textbf{Sub-figure (d)} showcases the output of a graph clustering problem which merges superpoints with compatible class distribution and object agreement while cutting edges at the transition between objects.
        The resulting superpoint clusters define the instances of a panoptic 3D segmentation.
    }
    \label{fig:enter-label}
\end{figure*}

Our objective is to perform panoptic segmentation of a large 3D point cloud $\cP$ with potentially numerous and 
broad objects.
For clarity, we first present our graph clustering formulation at the point level.
We then explain how our approach can be supervised solely with 
per-node and pre-edge targets,
making its training particularly efficient.
Finally, we detail how our method can be easily generalized to superpoints to further increase its scalability.

\paragraph{Problem Statement.} Consistently with the image panoptic segmentation setup~\cite{kirillov2019panoptic}, each point $p \in \cP$ is associated with its position, a semantic label $\cls(p)\in [1,C]$ with $C$ the total number of classes, and an object index $\obj(p) \in \mathbb{N}$.
Points identified with a ``thing'' label (\eg, chair, car) are given an index that uniquely identifies this object.
Conversely, points with a ``stuff'' label (\eg, road, wall) are assigned an index \emph{shared by all points with the same class} within $\cP$.
Our goal is to recover the class and object index of all points in $\cP$.

\subsection{Panoptic Segmentation as Graph Clustering}

\label{sec:graph_clustering}

We propose viewing the panoptic segmentation task as grouping adjacent points with compatible class and object predictions.
We formulate this task as an optimization problem structured by a graph.
Specifically, we connect the points of $\cP$ to their $K$-nearest neighbors, forming a graph $\cG=(\cP,\cE)$ where $\cE \subset \cP \times \cP$ denotes these connections.

\paragraph{Spatial-Semantic Regularization.}
We use a neural network to associate each point $p$ with a 
probabilistic class prediction $x^\text{class}_p\in [0,1]^C$.
The architecture and supervision of this network are detailed in \secref{sec:supervision}.
A simple way to obtain a panoptic segmentation would be to group spatially adjacent points with the same class prediction $\argmax_c x^\text{class}_{p,c}$.
However, this approach neglects object structure, potentially causing two problems: erroneously merging adjacent same-class objects and unwanted object fragmentation due to the probabilistic nature of the prediction $x^\text{class}$.

To address this last issue, we aim to enforce the spatial consistency of object prediction. We introduce the signal $x$, defined for each point $p$ as the channelwise concatenation of its position $x^\text{pos}_p$ and its semantic prediction: $x_p=[x^\text{class}_p,x^\text{pos}_p]$.
We propose to compute a piecewise-constant approximation $y^\star$ of $x$ with an energy minimization problem regularized by the graph cut \cite{boykov2001fast} between its constant components \cite{landrieu2017structured}.
This approach aligns with well-established practices in 2D \cite{leclerc1989constructing,mumford1989optimal} and 3D \cite{labatut2009robust} analyses, and leads to the following optimization problem:
\begin{align}
    \!\!y^\star = \!\!\!\!\argmin_{y \in  \bR^{(C+3) \times \vert \cP \vert }}
    \sum_{p \in \cP} d(x_p,y_p)
    + 
    \lambda \!\!\!\! \sum_{(p,q) \in \cE} \!\!\!\!
    w_{p,q} [y_p \neq y_q]~,
     \label{eq:gmpp}
\end{align}
where $[a \neq b]\eqdef0$ if $a=b$ and $1$ otherwise, $\lambda > 0$ is a parameter controlling the regularization strength, and $w_{p,q}$ is a nonnegative weight associated with edge $(p,q)$, see below.

The dissimilarity function $d$ takes into account both the spatial and semantic nature of $x$:
\begin{align}
   \!\! d(x_p,y_p)\! =\!
   H(y^\text{class}_p, x^\text{class}_p)
   + \eta
    \Vert x^\text{pos}_p-y^\text{pos}_p\Vert^2
    ~,
     \label{eq:dist}
\end{align}
where $y^\text{class}_p$ is the first $C$ coordinates of $y_p$ and $y^\text{pos}_p$ the last $3$, and $\eta \geq 0$ a parameter. The term $H(x,y)$ denotes the cross-entropy between two distributions: 
$H(x,y)=-\sum_{c=1}^C y_c \log(x_c)$. 

\paragraph{Object-Guided Edge Weights.}
The edge weight $w_{p,q}$ determines the cost of predicting an object transition between $p$ and $q$. Designing appropriate edge weights is critical to differentiate between objects of the same class that are spatially adjacent, such as rows of chairs or cars in traffic. Edge weights should encourage cuts along probable object transitions and prevent cuts within objects.

To facilitate this, 
{we propose to train a neural network to predict}
an object agreement $a_{p,q} \in [0,1]$ for each edge $(p,q)$ in $\cE$. This value represents the probability that both points belong to the same object. We then determine the edge weight $w_{p,q} \in [0,\infty]$ as follows:
\begin{align}
    \label{eq:edgeweight}
    w_{p,q} =a_{p,q}/(1-a_{p,q}+\epsilon)~,
\end{align}
with $\epsilon>0$ a fixed parameter.
High values of $w_{p,q}$ discourage cuts between points $p$ and $q$ that are confidently predicted to belong to the same object. Conversely, a smaller $w_{p,q}$ means that cuts between edges with probable transition $a_{p,q}$ are not heavily penalized.

\paragraph{Graph Clustering.} 
The constant components of the solution $y^\star$ of \eqref{eq:gmpp} define a clustering $\cK$ of $\cP$.
Clusters $\cK$ contain spatially adjacent points with compatible semantics, and their contours should follow predicted object transitions.

\paragraph{Converting to a Panoptic Segmentation.} 
We can derive a panoptic segmentation from the clusters $\cK$. 
For each cluster, we calculate the average point distribution of its constituent points and select the class with the highest probability.
We then associate a unique object index to each cluster $k$ predicted as a ``thing'' class. Likewise, we assign to each cluster classified as ``stuff'' an index shared by all clusters predicted as the same class. 
Finally, each individual point is labeled with the class and object index of its respective cluster.

\paragraph{Optimization.}
The optimization problem expressed in \eqref{eq:gmpp} is widely explored in the graph optimization literature.
Referred to as the \emph{generalized minimal partition problem} \cite{landrieu2017cut}, this problem is related to the Potts models \cite{potts1952some} and image partitioning techniques \cite{leclerc1989constructing,mumford1989optimal}. 
We adapt the parallel $\ell_0$-cut pursuit algorithm \cite{raguet2019parallel,landrieu2017structured} to the dual spatial-semantic nature of the regularized signal. The resulting algorithm is particularly scalable and can handle graphs with hundreds of millions of edges on a standard workstation. 
This allows us to process large point clouds in one inference without the need for tiling and instance stitching post-processing.

\subsection{Local Supervision}

\label{sec:supervision}
A major benefit of our approach is that it can be entirely supervised with local auxiliary tasks: all losses described in this section are sums of simple functions depending on one or two points at the time.
In particular, we bypass the computationally expensive step of matching true instances with their predicted counterparts.

Recall from \secref{sec:graph_clustering} that we can obtain a panoptic segmentation by predicting the parameters of a graph clustering problem: the semantic predictions $x^\text{class}_{p}$ and the object agreements $a_{p,q}$.
These quantities are both derived from a common pointwise embedding $\{e_p\}_{p \in \cP}$, computed by a neural network.

\paragraph{Predicting Semantics.}
We predict the class distribution $x^\text{class}_p = \softmax(\phi^{\text{class}}(e_p))$ with $\phi^{\text{class}}$ a Multi-Layer Perceptron (MLP). This distribution is supervised by its cross-entropy against the true class $\cls(p)$:
\begin{align}
\cL^\text{class}_{p}
=
H(x^\text{class}_p,\mathbf{1}(\cls(p)))~,
\label{eq:cls}
\end{align}
with $\mathbf{1}(c) \in \{0,1\}^C$ the one-hot embedding of class $c$.

\paragraph{Predicting Object Agreement.}  
To predict the object agreement ${a}_{p,q}$ between two adjacent points $(p,q) \in \cE$, we employ an MLP $\phi^{\text{object}}$ whose input is a symmetric combination of the points' embedding vectors:
\begin{align}
    {a}_{p,q} =
    \sigmoid 
    \left(
    \phi^{\text{object}}
    \left(
        \left[
            (e_p  + e_q) / 2
            ,
            \mid e_p  - e_q \mid
        \right]
    \right)
    \right)~,
\end{align}
where $\mid \cdot \mid$ refers to the termwise absolute value.
The true object agreement $\hat{a}_{p,q}$ is assigned the value of $1$ if $\obj(p)=\obj(q)$ and $0$ otherwise.
The prediction of ${a}_{s,t}$ can be seen as \emph{a binary edge classification problem} as inter- and intra-object edges \cite{landrieu2019ssp}, and is supervised with the cross-entropy between true and predicted object agreements:
\begin{align}
\cL^\text{object}_{p,q}
=
H(\bernoulli(a_{p,q}),\bernoulli(\hat{a}_{p,q}))~,
\label{eq:affinity}
\end{align}
where $\bernoulli(a)$ denote the Bernoulli distribution parametrized by $a \in [0,1]$.

\paragraph{Loss Function.} We combine the two losses above into a single objective $\cL$:
\begin{align}
    \cL =     \frac1{\mid\cP\mid  }
    \sum_{p\in \cP}
\cL^\text{class}_p
    +
    \frac1{\mid\cE\mid  }\sum_{(p,q) \in \cE} {\cL^\text{object}_{p,q}}~,
\end{align}
with $\mid \cE \mid$ and $\mid \cP \mid$ the total number of edges and 3D points, respectively.
\subsection{Extension to Superpoints}
In this section, we discuss the extension of our method to a superpoint-based approach for enhanced scalability.

\paragraph{Motivation.}
We aim to design a panoptic segmentation method that can scale to large 3D point clouds. While the formulation presented in the previous section is advantageous, it still requires computing embeddings and predictions for each individual point, which can be memory intensive and limits the volume of data that can be processed simultaneously.
We propose to group adjacent points with similar local geometry and color into \emph{superpoints}, and to only compute embeddings and predictions for superpoints and not individual points. By doing so, we drastically reduce the computational and memory requirements of our method, enabling it to handle larger 3D point clouds at once.

\paragraph{Computing Superpoints.}
We partition the point cloud $\cP$ into a set of non-overlapping superpoints $\cS$. %
We use the partition method implemented by Robert \etal in SPT \cite{robert2023efficient}, which defines superpoints as the constant components of a low-surface piecewise constant approximation of geometric and radiometric point features.

Although the superpoints $\cS$ form a high-purity oversegmentation of $\cP$, some superpoints can span multiple objects. To account for this, we associate each superpoint $s$ with its \emph{majority-object} $\obj(s)$ defined as the most common object index within its points: $\obj(s) = \mode \{\obj(p) \mid p  \in s \}$. 
Likewise, we define $\cls(s)=\mode \{\cls(p) \mid p  \in s \}$. 

\paragraph{Adapting Graph Clustering.}
Our clustering step can be directly adapted by substituting the point set $\cP$ with the superpoint set $\cS$, and defining the graph $\cG$ by connecting superpoints with adjacent points following the approach of SPT \cite{robert2023efficient}.
We replace the point position $x^\text{pos}_p$ by the coordinates of the superpoints' centroids $x^\text{pos}_s$.
All other steps remain unchanged.

\paragraph{Superpoint Embedding.} 
We use a superpoint-embedding network to compute the superpoint features $e_s$ for $s\in\cS$.
We employ the Superpoint Transformer model \cite{robert2023efficient} for its efficiency and ability to leverage large spatial context. 
More details on this design choice are provided in the Appendix.

\paragraph{Superpoint Semantic Supervision.}
We supervise the semantic superpoint prediction $x^\text{class}_s$ with \eqref{eq:cls} where we replace $\cls(p)$ with $\cls(s)$.

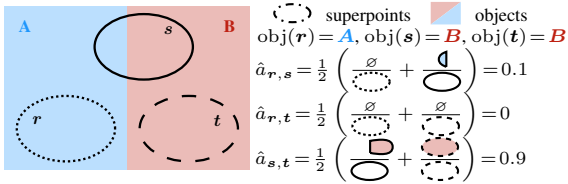
\begin{figure}[t!]
    \centering
    \scriptsize
     \begin{tabular}{l@{}l@{}l@{}l}
    \multirow{2}{*}{
    \resizebox{0.4\linewidth}{!}{
    \begin{tikzpicture}

        \fill[fill=OBJECTACOLOR!30, draw=none] (0,0) rectangle (3,4);
        \fill[fill=OBJECTBCOLOR!30,draw=none] (3,0) rectangle (6,4);

        \draw [ultra thick, draw=black, dotted] (1.5,1) ellipse (1.2cm and 0.8cm);
        \draw [ultra thick, draw=black] (3.4,3) ellipse (1.2cm and 0.8cm);
        \draw [ultra thick, draw=black, dashed, dash pattern=on 10pt off 10pt] (4.5,1) ellipse (1.2cm and 0.8cm);

        \node [draw=none] at (0.8, 1.2) {\bf \large $\bm{r}$};
        \node [draw=none] at (4.0, 3.4) {\bf \large $\bm{s}$};
        \node [draw=none] at (5.2, 1.2) {\bf \large $\bm{t}$};

        \node [draw=none, text=OBJECTACOLOR, font=\bfseries\large] at (0.5, 3.5) {A};
        \node [draw=none, text=OBJECTBCOLOR, font=\bfseries\large] at (5.5, 3.5) {B};

    \end{tikzpicture}
    }
    }
    &
    \multicolumn{3}{l}{
    \begin{tikzpicture}[scale=0.2,baseline=-1mm]
         \fill [thick, draw=black, fill=none, dashdotted] (0,0) ellipse (1.2cm and 0.6cm);
    \end{tikzpicture}
    \; superpoints
    \;\;
    \begin{tikzpicture}[scale=0.2,baseline=0.5mm]
            \fill[color=OBJECTBCOLOR!30,draw=none] (0,0) -- (0,1.5) -- (2.0,1.5) -- cycle;
            \fill[color=OBJECTACOLOR!30,draw=none] (0,0) -- (2,0) -- (2.0,1.5) -- cycle;
    \end{tikzpicture}
     \; objects
    }
    \\
    &
    $\obj(\textcolor{black}{{\bm{r}}}) \!=\! \textcolor{OBJECTACOLOR}{\bm{A}}$,\; 
    &
    $\obj(\textcolor{black}{ {\bm{s}}}) \!=\! \textcolor{OBJECTBCOLOR}{\bm B}$,\; 
    &
    $\obj(\textcolor{black}{{\bm{t}}}) \!=\!  \textcolor{OBJECTBCOLOR}{\bm B}$
    \\
    &
    \multicolumn{3}{l}{
    \hspace{-3mm}
    $\hat{a}_{{\bm{r}},{\bm{s}}}
    \!=\!
    \frac12
    \left(
        \frac{
    \varnothing
    }
    {
    \begin{tikzpicture}[scale=0.2]
         \fill [thick, draw=black, fill=none, dotted, dash pattern = on 1pt off 1pt] (0,0) ellipse (1.2cm and 0.6cm);
    \end{tikzpicture}
    } 
    +
    \frac{
    \begin{tikzpicture}[scale=0.2]
         \fill [thick, draw=black, fill=OBJECTACOLOR!30] (0,0) -- (0,1) to[out=200, in=90] (-0.5,0.5) to[out=-90, in=160] (0,0) ;
    \end{tikzpicture}
    }
    {
    \begin{tikzpicture}[scale=0.2]
         \draw [thick, draw=black] (0,0) ellipse (1.2cm and 0.6cm);
    \end{tikzpicture}
    }
    \right)\!=\!0.1$
    }
    \\
    &
    \multicolumn{3}{l}{
    \hspace{-3mm}
    $\hat{a}_{\bm r, \bm  t} \!=\! \frac12
    \left(
    \frac{
    \varnothing
    }
    {
    \begin{tikzpicture}[scale=0.2]
         \fill [thick, draw=black, fill=none, dotted, dash pattern = on 1pt off 1pt] (0,0) ellipse (1.2cm and 0.6cm);
    \end{tikzpicture}
    }
    +
    \frac{
    \varnothing
    }
    {
    \begin{tikzpicture}[scale=0.2]
         \fill [thick, draw=black, fill=none, dashed, dash pattern = on 2pt off 2pt] (0,0) ellipse (1.2cm and 0.6cm);
    \end{tikzpicture}
    }
    \right)\!=\!0$
    }
    \\
    &
    \multicolumn{3}{l}{
    \hspace{-3mm}
    $\hat{a}_{\bm s, \bm t} \!=\! \frac12
    \left(
    \frac{
         \begin{tikzpicture}[scale=0.2]
         \fill [fill=OBJECTBCOLOR!30, thick, draw=black]  (0,0) -- (0,0.8) to[out=10, in=90] (1.5,0.4) to[out=-90, in=-10] (0,0) ;
    \end{tikzpicture}
    }
    {
    \begin{tikzpicture}[scale=0.2]
         \draw [thick, draw=black] (0,0) ellipse (1.2cm and 0.6cm);
    \end{tikzpicture}
    }
    +
    \frac{
        \begin{tikzpicture}[scale=0.2]
         \fill [thick, draw=black, fill=OBJECTBCOLOR!30, dotted, dash pattern = on 2pt off 2pt] (0,0) ellipse (1.2cm and 0.6cm);
    \end{tikzpicture}
    }
    {
    \begin{tikzpicture}[scale=0.2]
         \draw [thick, draw=black, dotted, dash pattern = on 2pt off 2pt] (0,0) ellipse (1.2cm and 0.6cm);
    \end{tikzpicture}
    }
    \right)\!=\!0.9$
    }

    \end{tabular}
    \caption{{\bf Superpoint Object Agreement.} We compute for each pair of adjacent superpoint $(s,t)$ an object agreement score $\hat{a}_{s,t}$. This value is defined by the average overlap ratio between $s$ and $t$ and their majority-objects $\obj(t)$ and $\obj(s)$, see \eqref{eq:agreement}.}
    \label{fig:affinity}
\end{figure}

\paragraph{Superpoint Object Agreement Supervision.} 
While the true object agreement $\hat{a}_{p,q}$ between two points is binary, the agreement between superpoints spans a continuum. As illustrated in \figref{fig:affinity}, we quantify this agreement as: 
\begin{align}
\label{eq:agreement}
\hat{a}_{s,t} = \frac12
\left(
\frac{\mid s \cap \cP_{\mid\obj(t)} \mid}{\mid s\mid }
+
\frac{\mid t \cap \cP_{\mid\obj(s)} \mid}{\mid t\mid }
\right)
~,
\end{align}
where 
$ \cP_{\mid o} \eqdef  \{ p \in \cP \mid \obj(p) =  o \}$ is the set of points of $\cP$ with the object index $o$, and $\mid s \mid$ is the count of 3D points in $s$. We can now supervise the predicted object agreement $a_{s,t}$ with \eqref{eq:affinity} unchanged.

\section{Experiments}

We first present the datasets and metrics used for evaluation in \secref{sec:datasets}, then our main results and their analysis in \secref{sec:results}, and finally an ablation study in \secref{sec:ablation}.

\subsection{Datasets and Metrics}
\label{sec:datasets}
\paragraph{Datasets.} We present the four datasets used in this paper.
\begin{itemize}
    \item {\bf S3DIS} \cite{armeni20193d}.
    This indoor scanning dataset consists of $274$ million points distributed across $271$ rooms in $6$ building floors---or areas.
    We do not use the provided room partition, as they require significant manual processing and may not translate well to other environments such as open offices, industrial sites, or mobile mapping.
    Instead, we merge all rooms in the same area and treat each floor as one single large-scale acquisition \cite{thomas2019kpconv,chaton2020torch}.
    
    We follow the standard evaluation protocol, using the area $5$ as a test set and implementing $6$-fold cross-validation.
    In line with Xiang \etal's \cite{xiang2023review} proposal, we treat all $13$ classes as ``thing''.
    However, certain classes, such as \emph{walls}, \emph{ceiling}, and \emph{floors}, are susceptible to arbitrary division due to room splitting, making their evaluation somewhat inconsistent.
    As a result, we also present panoptic metrics in which these three classes are considered as ``stuff''.
    
    \item {\bf ScanNet} \cite{dai2017scannet}.
    This dataset consists of  $237$M 3D points organized in $1501$ medium-scale indoor scenes.
    We evaluate SuperCluster on ScanNet's open test set, as the hidden test set is not evaluated for panoptic segmentation.
    We use for ``things'' the class evaluated in the instance segmentation setting: \emph{bathtub}, \emph{bed}, \emph{bookshelf}, \emph{cabinet}, \emph{chair}, \emph{counter}, \emph{curtain}, \emph{desk}, \emph{door}, \emph{other furniture}, \emph{picture}, \emph{refrigerator}, \emph{shower curtain}, \emph{sink}, \emph{sofa}, \emph{table}, \emph{toilet}, and \emph{window}.
    The \emph{walls} and \emph{floor} class are designated as ``stuff''.

    \item {\bf KITTI-360} \cite{liao2022kitti360}.
    Containing over $100$k mobile mapping laser scans from an outdoor urban environment, we utilize the \emph{accumulated point clouds} format, which aggregates multiple sensor rotations to form $300$ extensive scenes with an average of more than $3$ million points.
    We train on $239$ scenes and evaluate it on the remaining $61$.
    \emph{Building} and \emph{cars} classes are treated as ``thing'' while the remaining $13$ are classified as ``stuff''.

    \item {\bf DALES} \cite{varney2020dales}.
    This large-scale aerial scan data set spans $10$ km$^2$ and contains $500$ millions of 3D points organized along $40$ urban and rural scenes, of which we use $12$ for evaluation.
    The ``thing'' classes are \emph{buildings}, \emph{cars}, \emph{trucks}, \emph{power lines}, \emph{fences}, and \emph{poles}.
    \emph{Ground} and \emph{vegetation} evaluated as ``stuff''.
\end{itemize}

\begin{table}
 \caption{{\bf S3DIS Area~5.} We report the semantic (\emph{SS}) and panoptic segmentation results of the top-performing semantic segmentation methods on the fifth area of S3DIS, as well as panoptic segmentation approaches implemented by Xiang \etal~\cite{xiang2023review}. We provide two panoptic metrics by considering all classes as ``things'' (\emph{PS - no ``stuff''}) and with \emph{wall}, \emph{ceiling} and \emph{floor} as ``stuff'' (\emph{PS}).}
 \label{tab:s3dis:f5}
 \resizebox{\columnwidth}{!}{
\begin{tabular}{lllllllll}
    \toprule
    & \multicolumn{1}{c}{Size}
    & \multicolumn{1}{c}{SS} & \multicolumn{3}{c}{PS - no ``stuff''} & \multicolumn{3}{c}{PS}\\  
    \cmidrule(r{4pt}){2-2} \cmidrule(r{4pt}){3-3} \cmidrule(r{4pt}){4-6} \cmidrule(r{4pt}){7-9}
    &\multicolumn{1}{c}{\small{$\times 10^6$}} &\multicolumn{1}{c}{mIoU}  & PQ & RQ & SQ & PQ & RQ & SQ  \\
     \midrule
    \multicolumn{8}{l}{   Semantic segmentation models}
    \\\midrule
    SPT \cite{robert2023efficient} & \bf 0.21 & 68.9  & - & - & - & - & - & - \\ 
    Point Trans.\cite{zhao2021point} &  7.8 & 70.4 & - & - & - & - & - & - \\
    PointNeXt-XL \cite{qian2022pointnext}& 41.6 &71.1 & - & - & - & - & - & -\\
    Strat. Trans. \cite{lai2022stratified,Wang2022WindowNE}& 8.0 & \bf 72.0 & - & - & - & - & - & - \\
    \midrule
   \multicolumn{8}{l}{   Panoptic segmentation models}
    \\
    \midrule
    Xiang \etal \cite{xiang2023review} & 0.13\\
    \;\;+\;PointNet++ \cite{qi2017pointnetpp} & +3.0 &58.7 & 24.6 & 32.6 & 68.2 & - & - & - \\
    \;\;+\;Minkowski \cite{choy20194minko} & +37.9 &63.8 & 39.2 & 48.0 & 74.9 & - & - & - \\
    \;\;+\;KPConv \cite{thomas2019kpconv} & +14.1 &65.3 & 41.8 & 51.5 & 74.7 & - & - & -\\
    PointGroup \cite{jiang2020pointgroup} in \cite{xiang2023review} & 7.7 & 64.9 & 42.3 & 52.0 & 74.7 & - & - & -\\
    \bf SuperCluster (ours) & \bf 0.21 & 68.1 & \bf 50.1 & \bf 60.1 & \bf 76.6 & \bf 58.4 & \bf 68.4 & \bf  77.8 \\

    \bottomrule
\end{tabular}
}
\end{table}

\begin{table}
 \caption{{\bf S3DIS 6-Fold.} We report the 6-Fold cross-validated semantic and panoptic segmentation results on S3DIS. No panoptic methods were evaluated in this setting to the best of our knowledge.}
 \label{tab:s3dis:6f}
 \resizebox{\columnwidth}{!}{
\begin{tabular}{lllllllll}
    \toprule
    & \multicolumn{1}{c}{Size}& \multicolumn{1}{c}{SS} & \multicolumn{3}{c}{PS - no ``stuff''} & \multicolumn{3}{c}{PS}\\  
    \cmidrule(r{4pt}){2-2}\cmidrule(r{4pt}){3-3} \cmidrule(r{4pt}){4-6} \cmidrule(r{4pt}){7-9}
    & \multicolumn{1}{c}{\small{$\times 10 ^6$}}&\multicolumn{1}{c}{mIoU}  & PQ & RQ & SQ & PQ & RQ & SQ  \\
     \midrule
    \multicolumn{8}{l}{Semantic segmentation models}
    \\\midrule
    DeepViewAgg \cite{robert2022learning}& 41.2 & 74.7  & - & - & - & - & - & - \\
    Strat. Trans. \cite{lai2022stratified,Wang2022WindowNE} & 8.0 & 74.9 & - & - & - & - & - & - \\
    PointNeXt-XL \cite{qian2022pointnext}& 41.6 & 74.9 & - & - & - & - & - & -\\
    SPT \cite{robert2023efficient} & \bf 0.21  & \bf  76.0  & - & - & - & - & - & - \\ 
    \midrule
    \multicolumn{8}{l}{  Panoptic segmentation models}
    \\
    \midrule
    \bf SuperCluster (ours) & \bf  0.21 & 75.3 & \bf 55.9 & \bf 66.3 & \bf 83.8 & \bf  62.7 & \bf  73.2 & \bf 84.8 \\

    \bottomrule
\end{tabular}
}
\end{table}

\paragraph{Evaluation Metrics.}
Recognition Quality (RQ) assesses object identification and classification.
Segmentation Quality (SQ) evaluates the alignment between target and predicted object segmentations.
Panoptic Quality (PQ) combines both measures.
We also compute the semantic segmentation performance by associating points with their superpoint's class and computing the mean Intersection over Union (mIoU).

\paragraph{Model Parameterization.}
Our backbone for the S3DIS and DALES datasets is a small SPT-64 model~\cite{robert2023efficient} with $205$k parameters.
We use a larger SPT-128 ($791$k parameters) for KITTI-360 and a slightly modified model for ScanNet ($1$M) parameters.
SuperCluster adds two small MLP $\phi^\text{class}$ and $\phi^\text{object}$ for a total of $4.4$k parameters for S3DIS and DALES, and $8.8$k parameters for KITTI-360 and ScanNet.

Our training batches are composed of 4 randomly sampled cylinders with a radius of $7$~m for S3DIS, $50$~m for KITTI and DALES, and entire scenes for ScanNet.
Partition parameters are adjusted so that $\cS/\cP\sim 30$ for S3DIS, DALES, and KITTI-360, and $20$ for ScanNet.

We can tune the graph clustering parameters after training to optimize the PQ on the training set: $\lambda$ in \eqref{eq:gmpp},  $\eta$ in \eqref{eq:dist}, and $\epsilon$ in \eqref{eq:edgeweight}.
As the clustering step is particularly efficient, we can evaluate tens of values in a few minutes.
More details are provided in the Appendix.

\subsection{Results and Analysis}
\label{sec:results}
\begin{figure*}
    \centering
     \begin{tabular}{lc@{}c@{}c@{}c}
     & S3DIS & ScanNet & KITTI-360 & DALES\\
    \rotatebox{90}{ \qquad \quad Input}
    &  \includegraphics[width=.23\textwidth, height=.11\textheight]{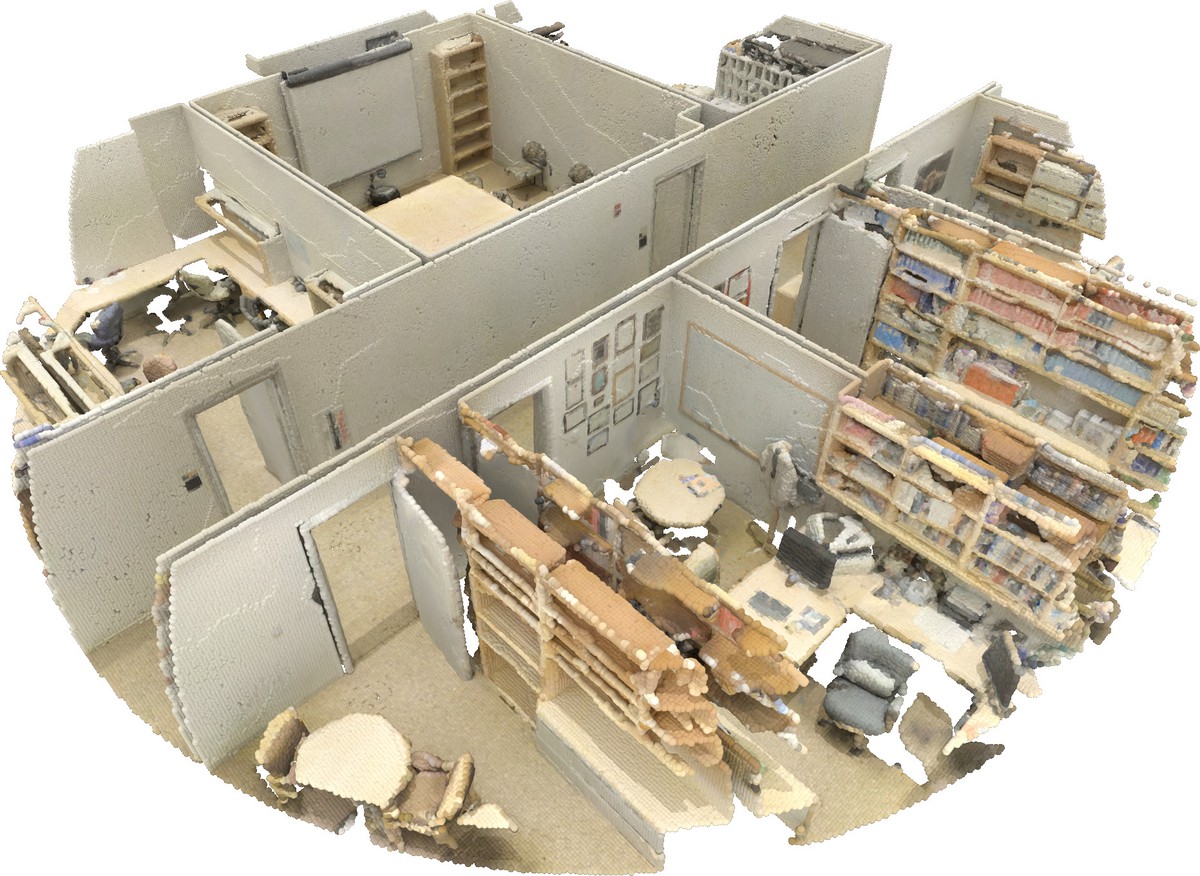}
    &  \includegraphics[width=.23\textwidth, height=.11\textheight]{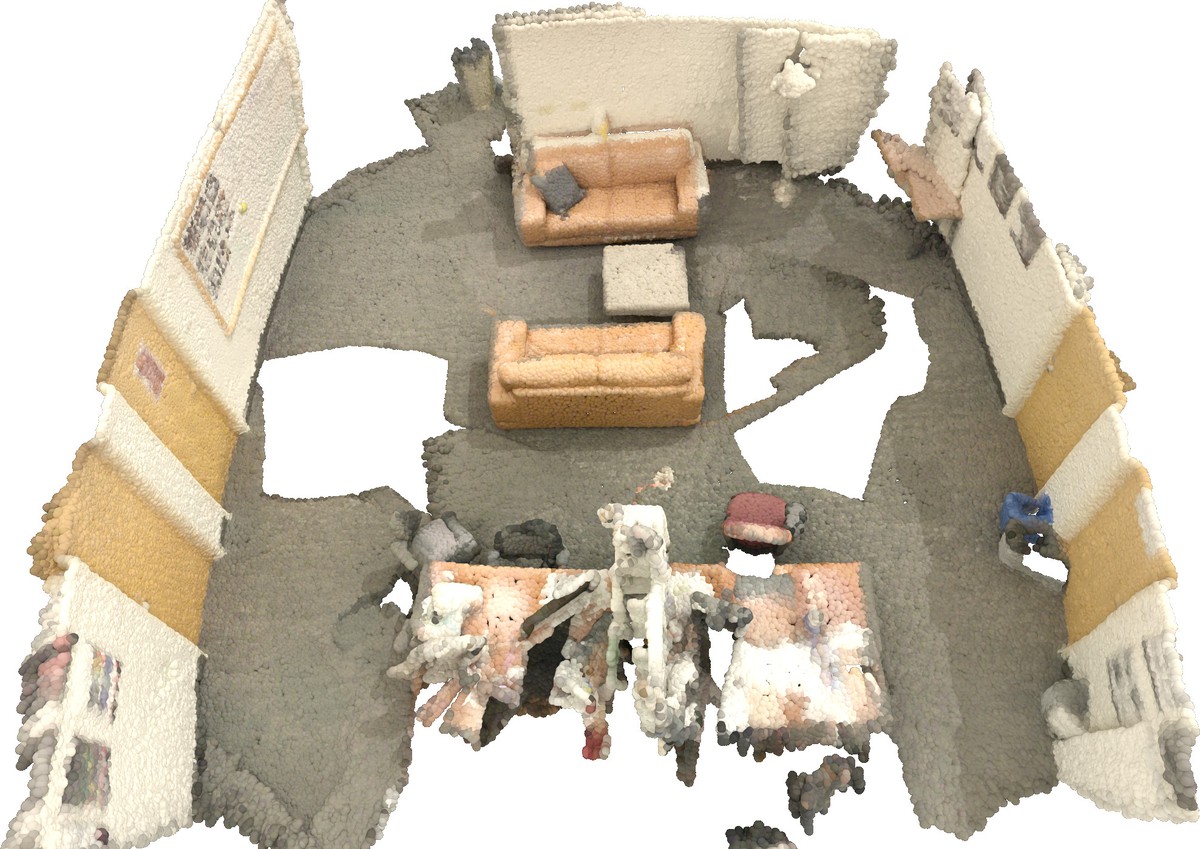}
    &  \includegraphics[width=.23\textwidth, height=.11\textheight]{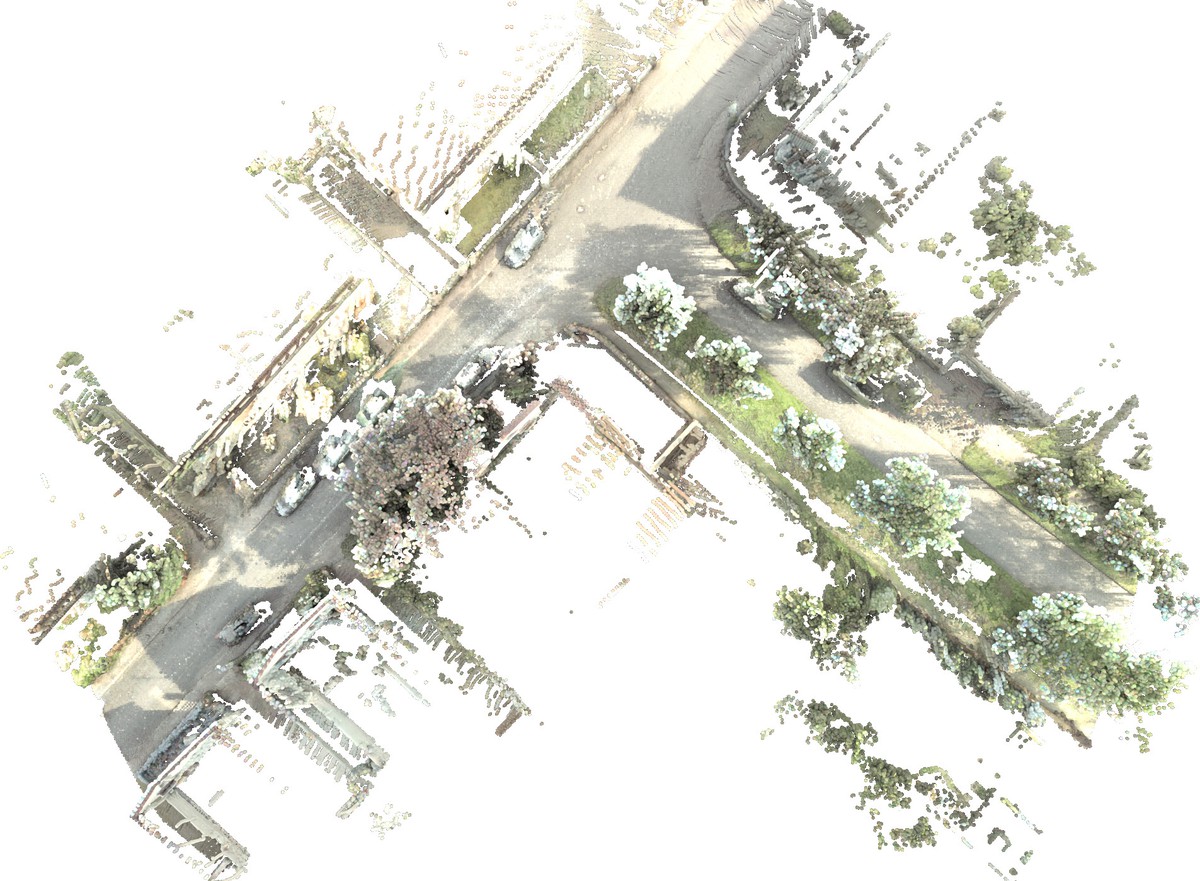}
    &  \includegraphics[width=.23\textwidth, height=.11\textheight]{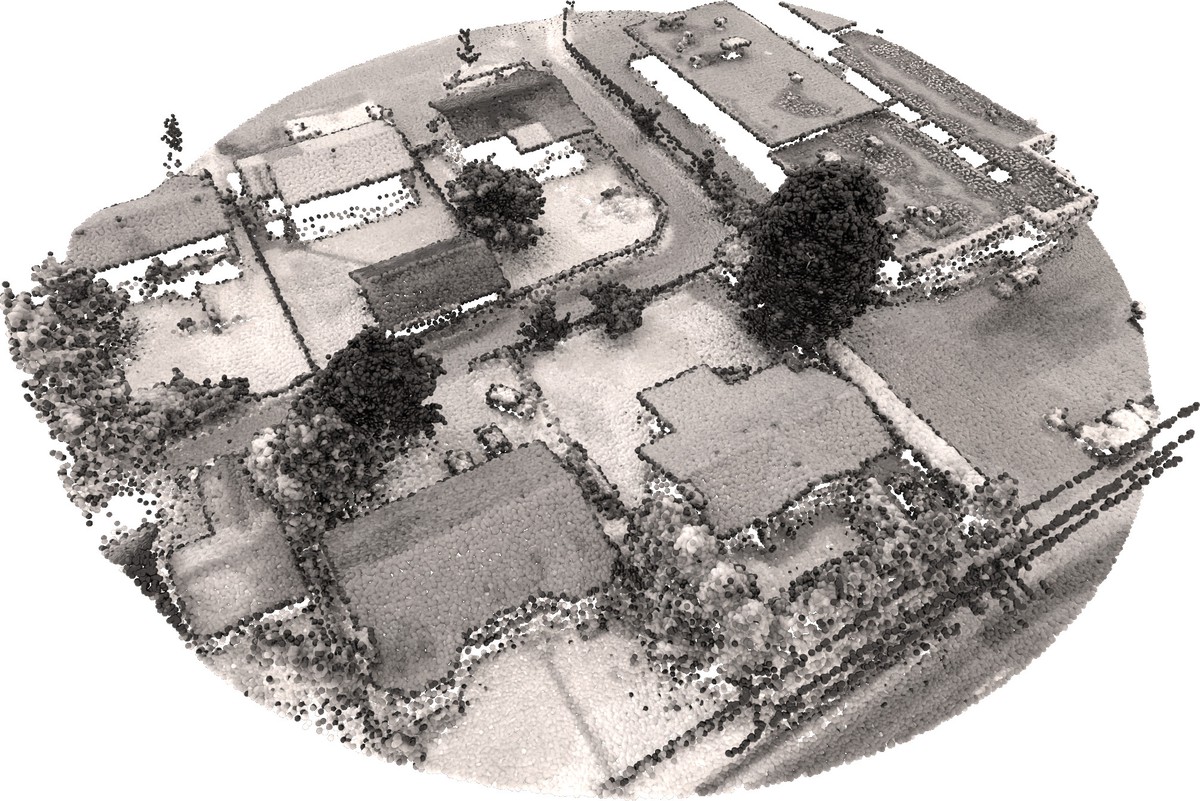}
    \\ 
    \rotatebox{90}{ \quad \;\; True classes} 
    &  \includegraphics[width=.23\textwidth, height=.11\textheight]{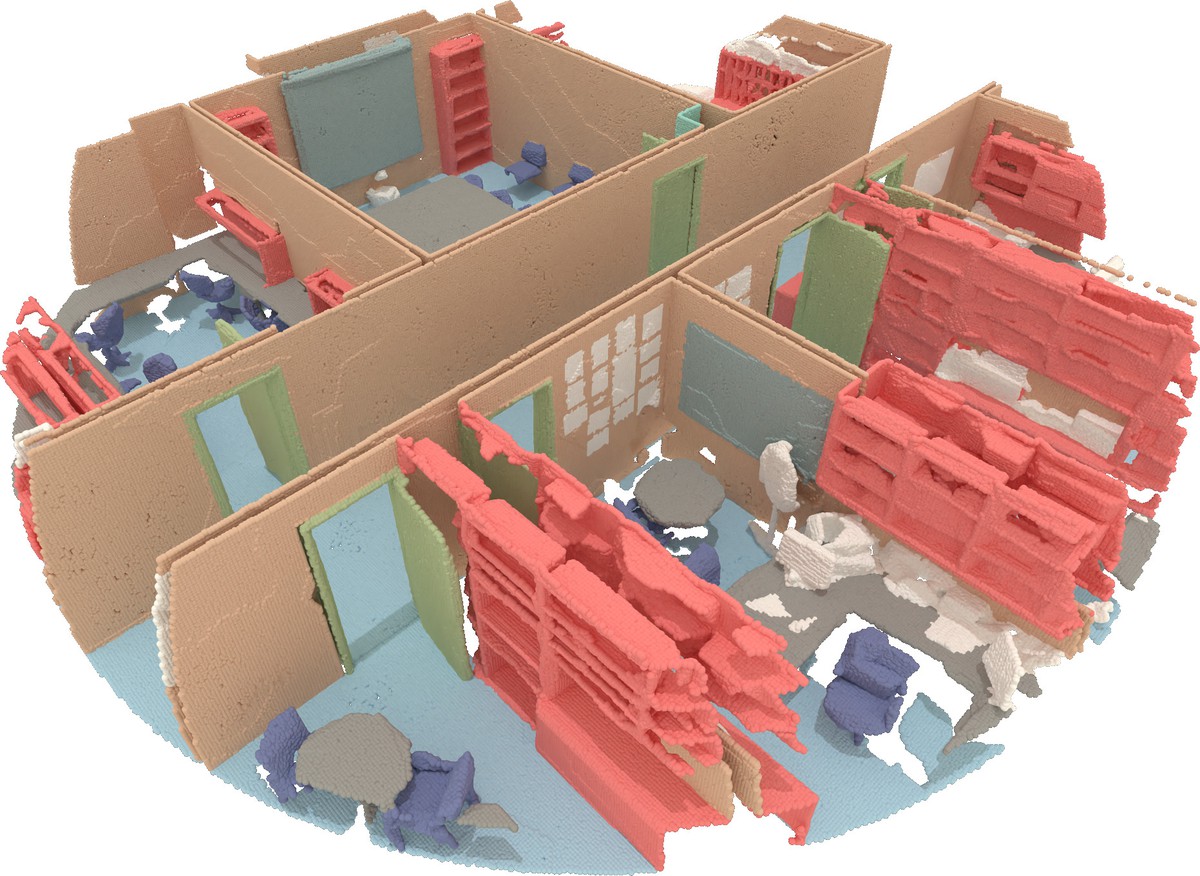}
    &  \includegraphics[width=.23\textwidth, height=.11\textheight]{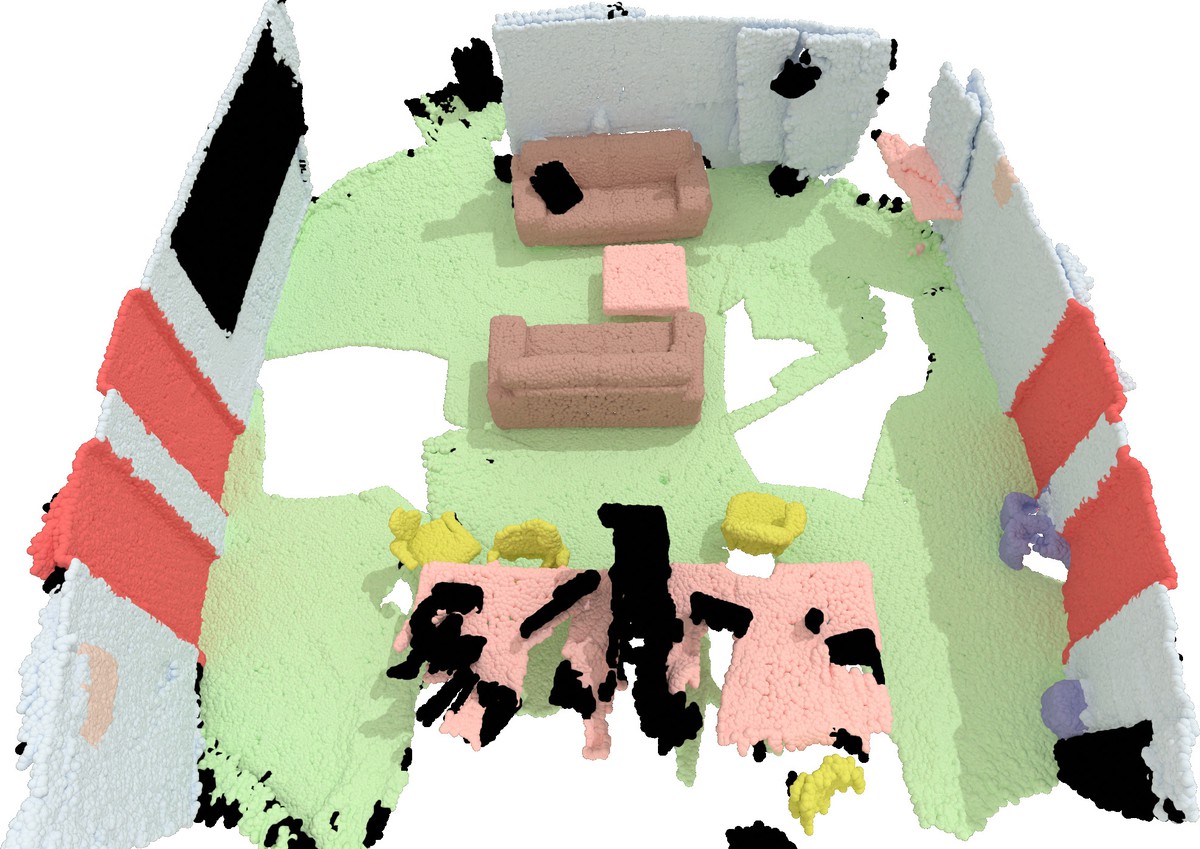}
    &  \includegraphics[width=.23\textwidth, height=.11\textheight]{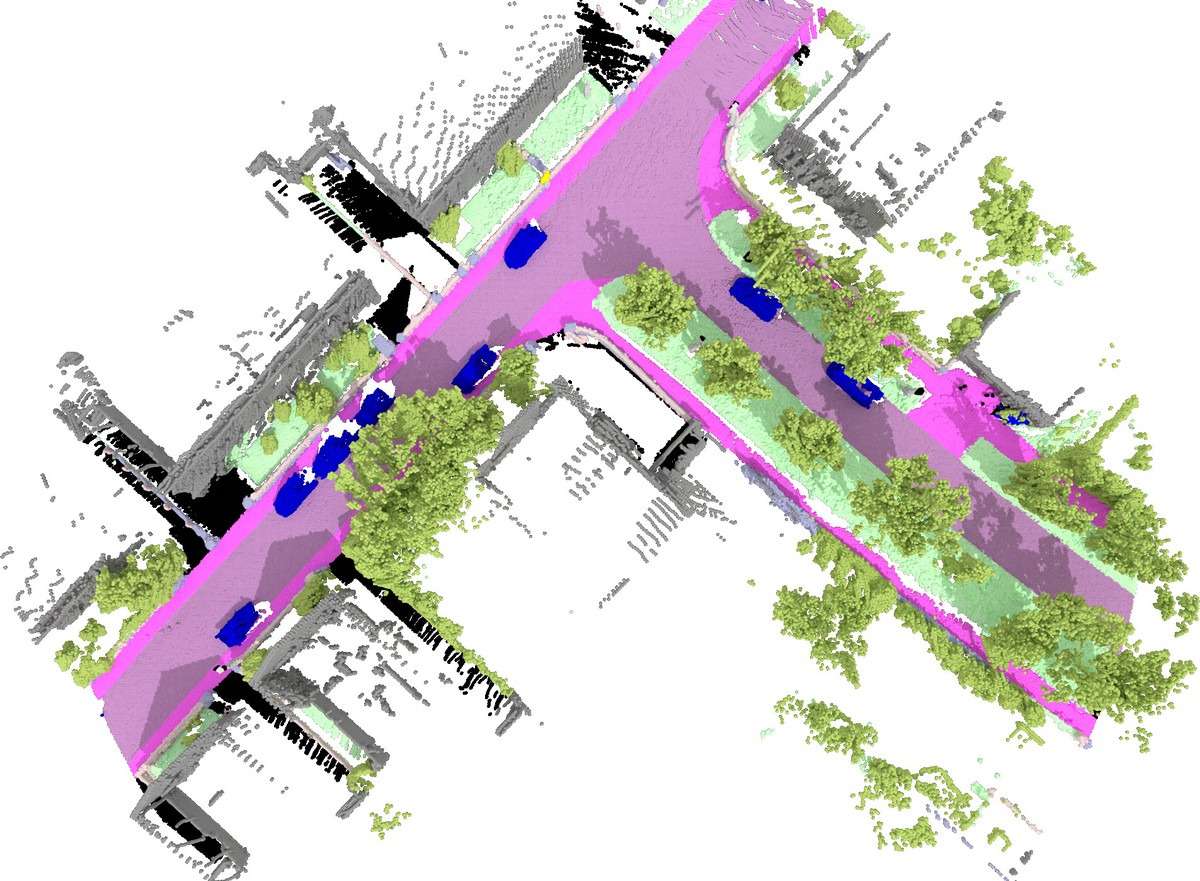}
    &  \includegraphics[width=.23\textwidth, height=.11\textheight]{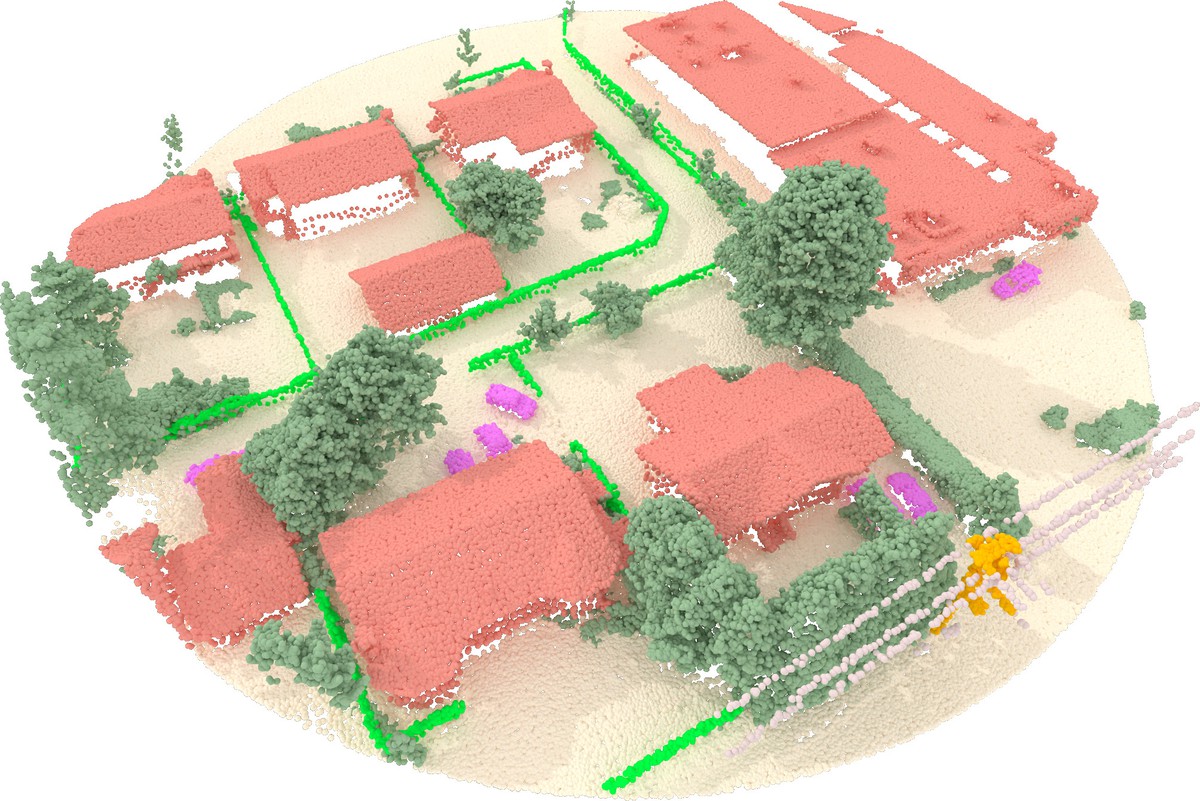}\\
    \rotatebox{90}{ \quad \;\; Pred. classes} 
    &  \includegraphics[width=.23\textwidth, height=.11\textheight]{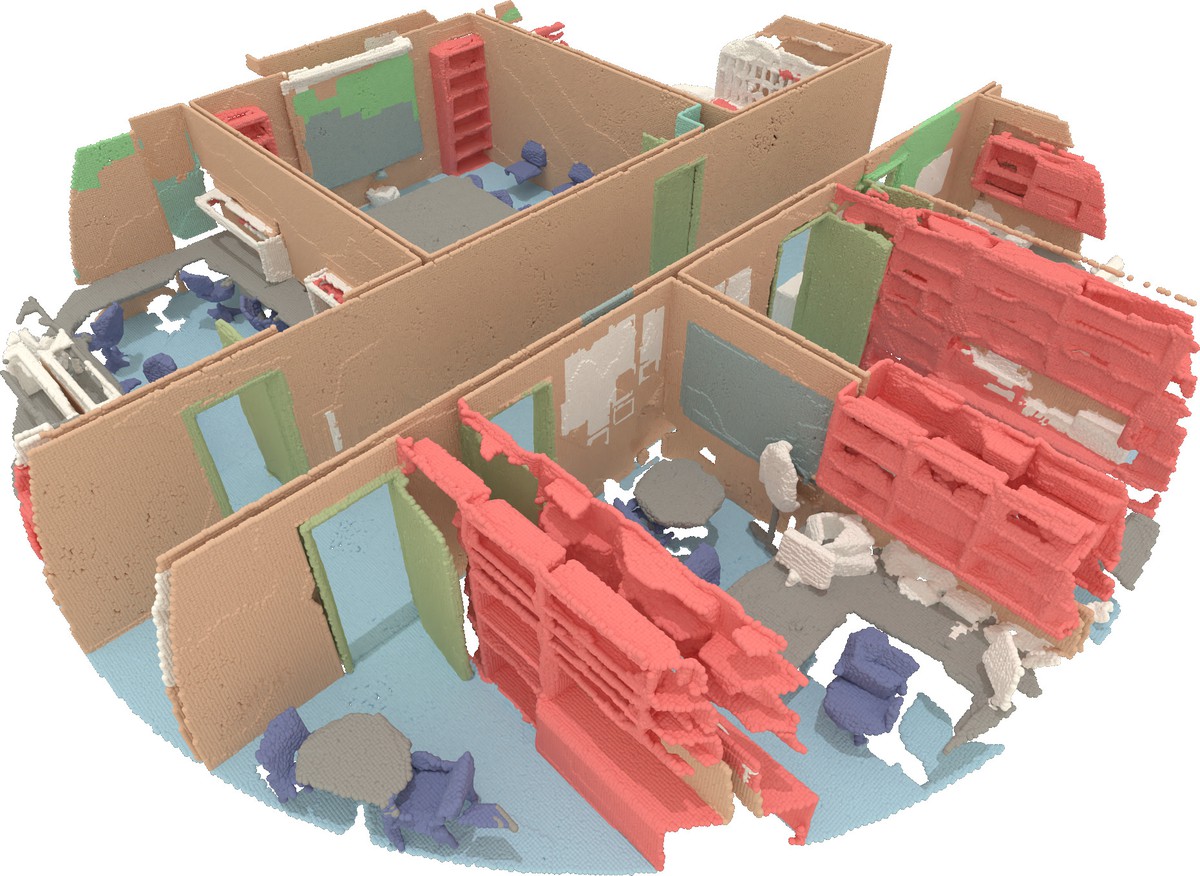}
    &  \includegraphics[width=.23\textwidth, height=.11\textheight]{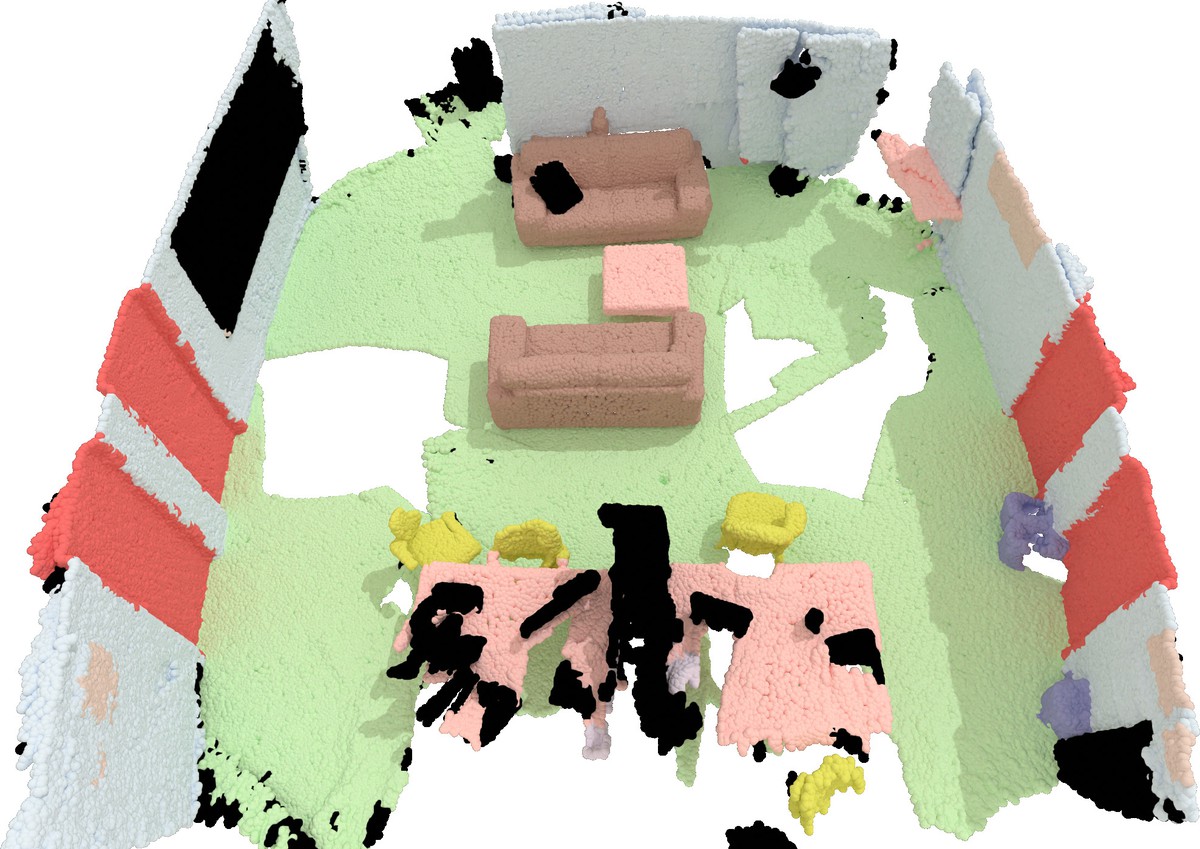}
    &  \includegraphics[width=.23\textwidth, height=.11\textheight]{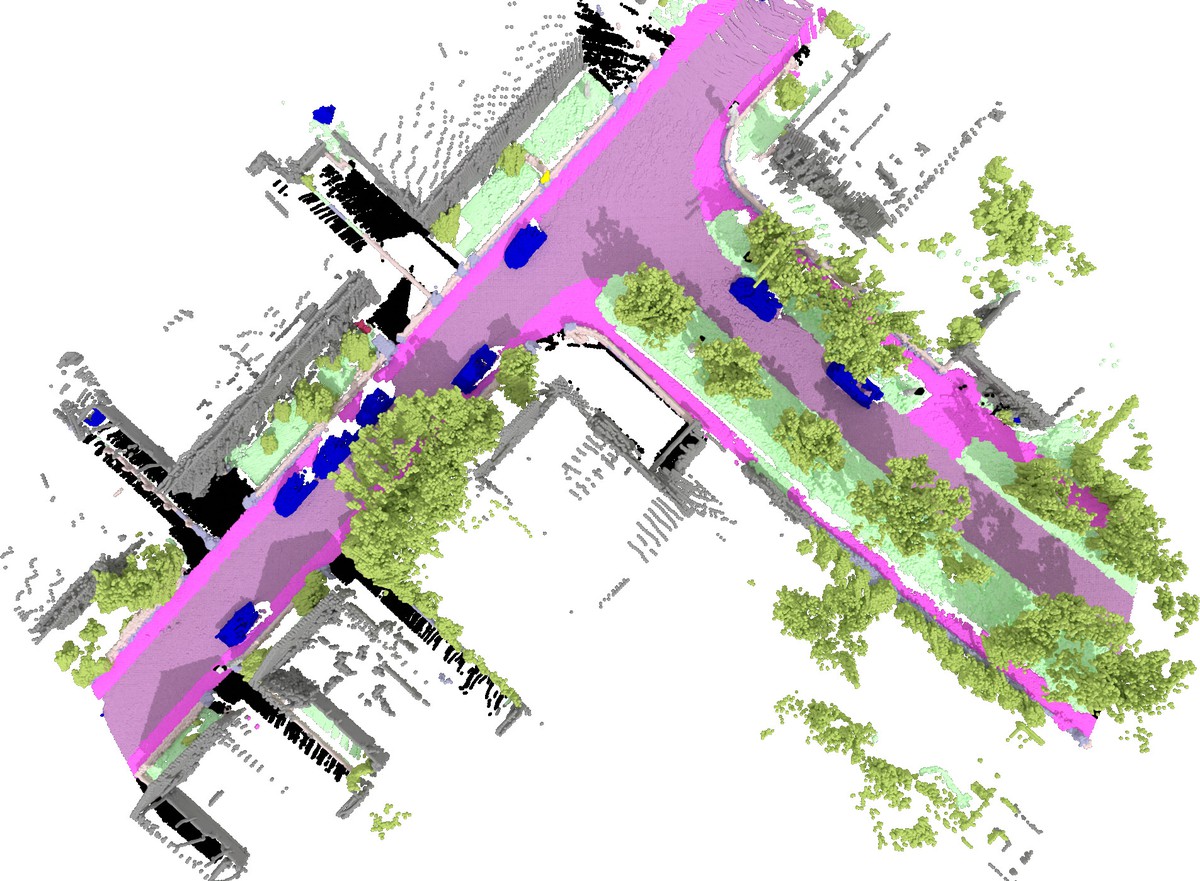}
    &  \includegraphics[width=.23\textwidth, height=.11\textheight]{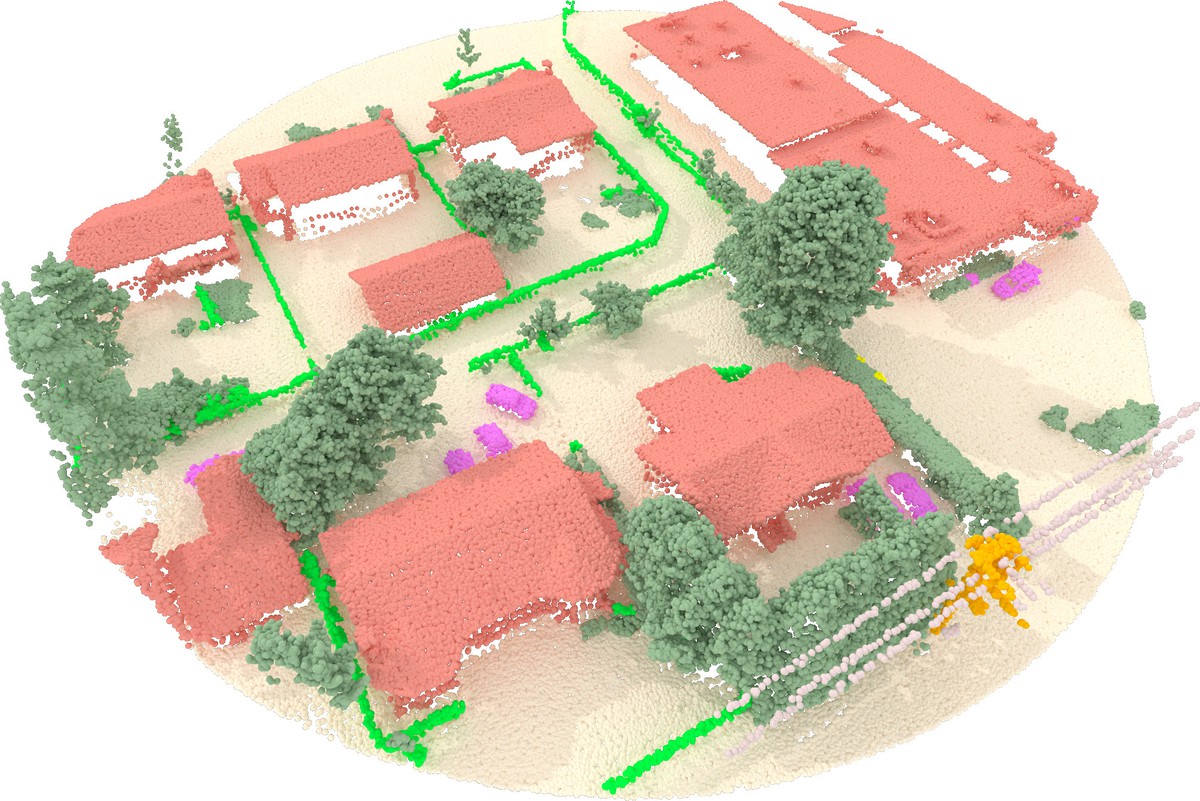}
    \\
    \rotatebox{90}{ \quad True instances} 
    &  \includegraphics[width=.23\textwidth, height=.11\textheight]{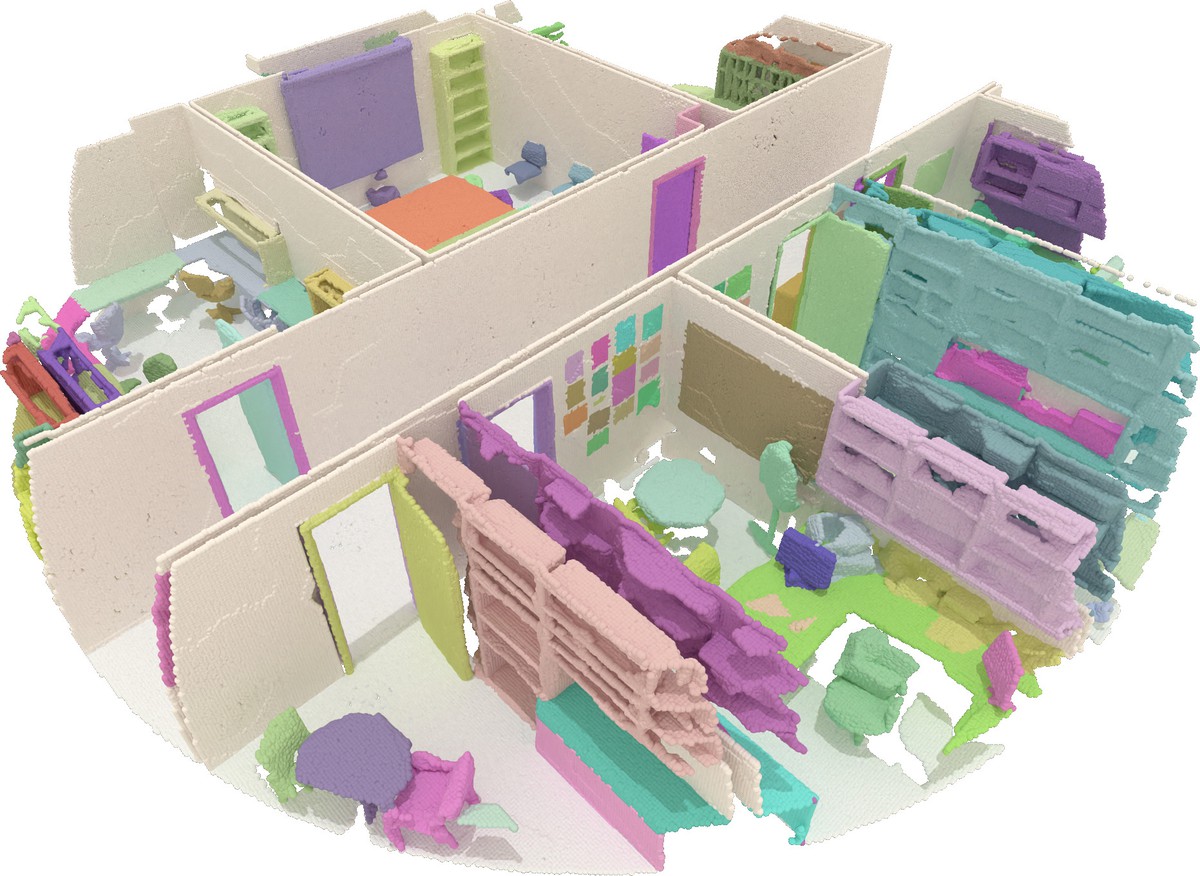}
    &  \includegraphics[width=.23\textwidth, height=.11\textheight]{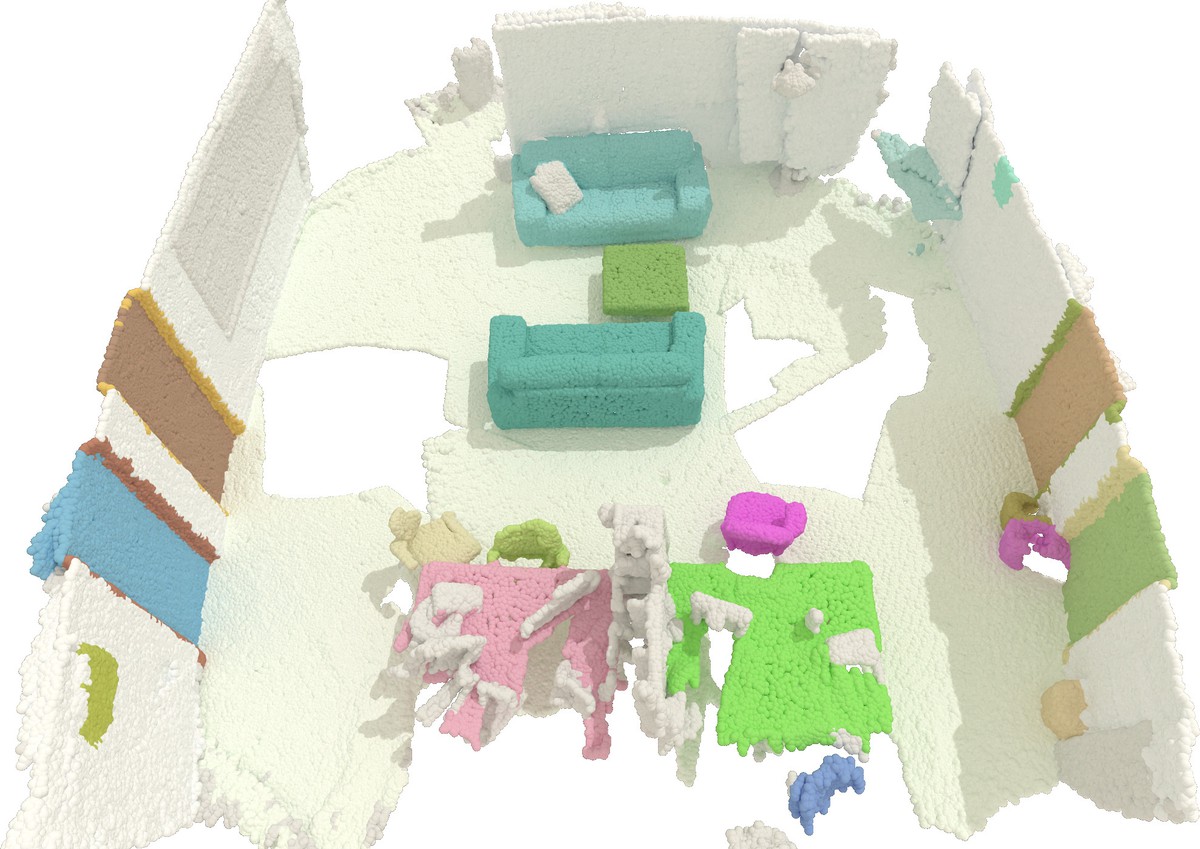}
    &  \includegraphics[width=.23\textwidth, height=.11\textheight]{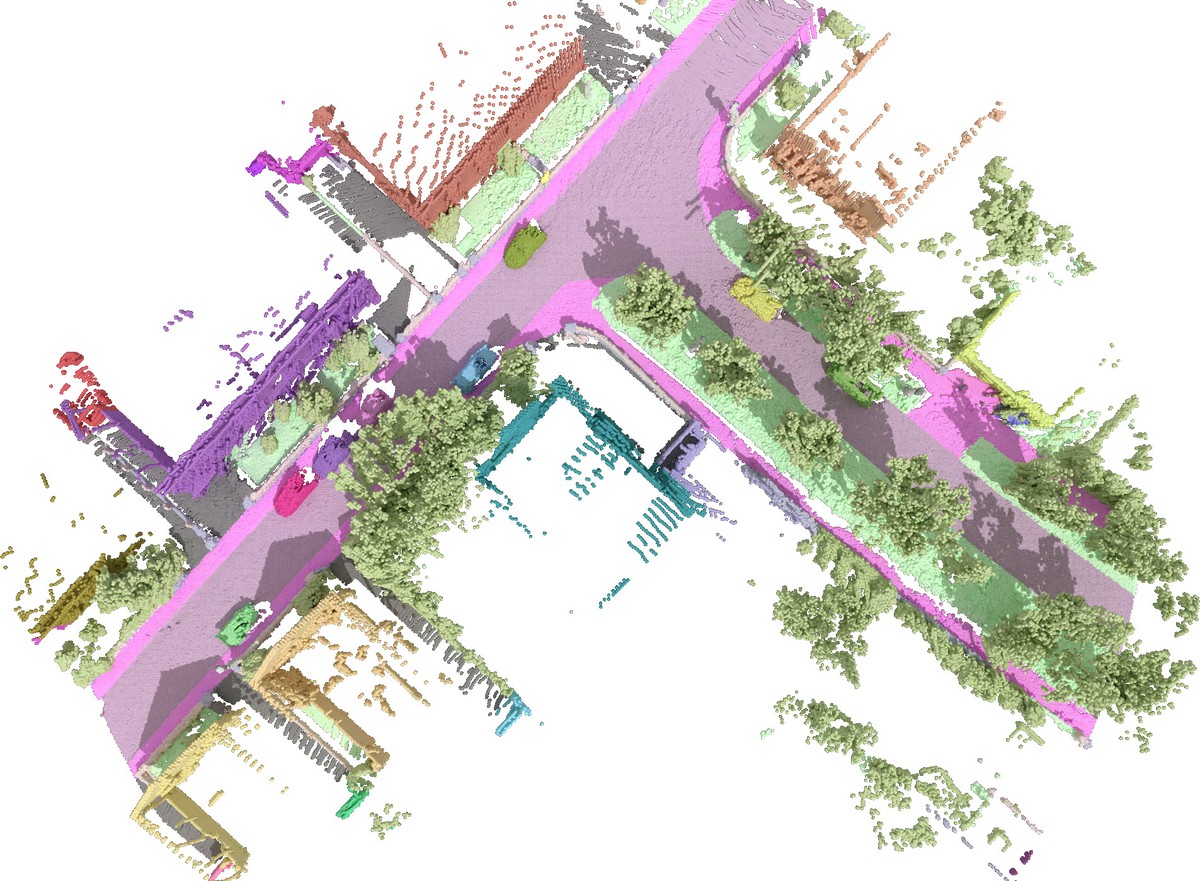}
    &  \includegraphics[width=.23\textwidth, height=.11\textheight]{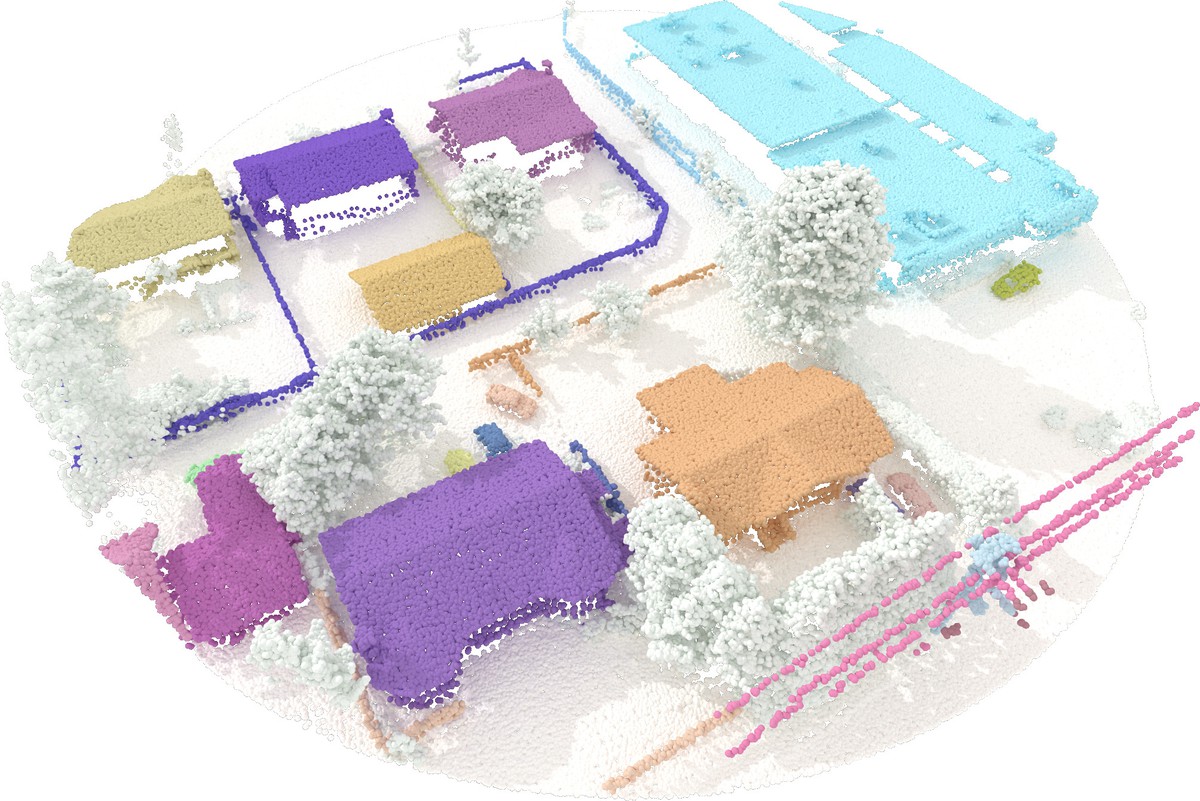}
    \\
    \rotatebox{90}{ \quad Pred. instances} 
    &  \includegraphics[width=.23\textwidth, height=.11\textheight]{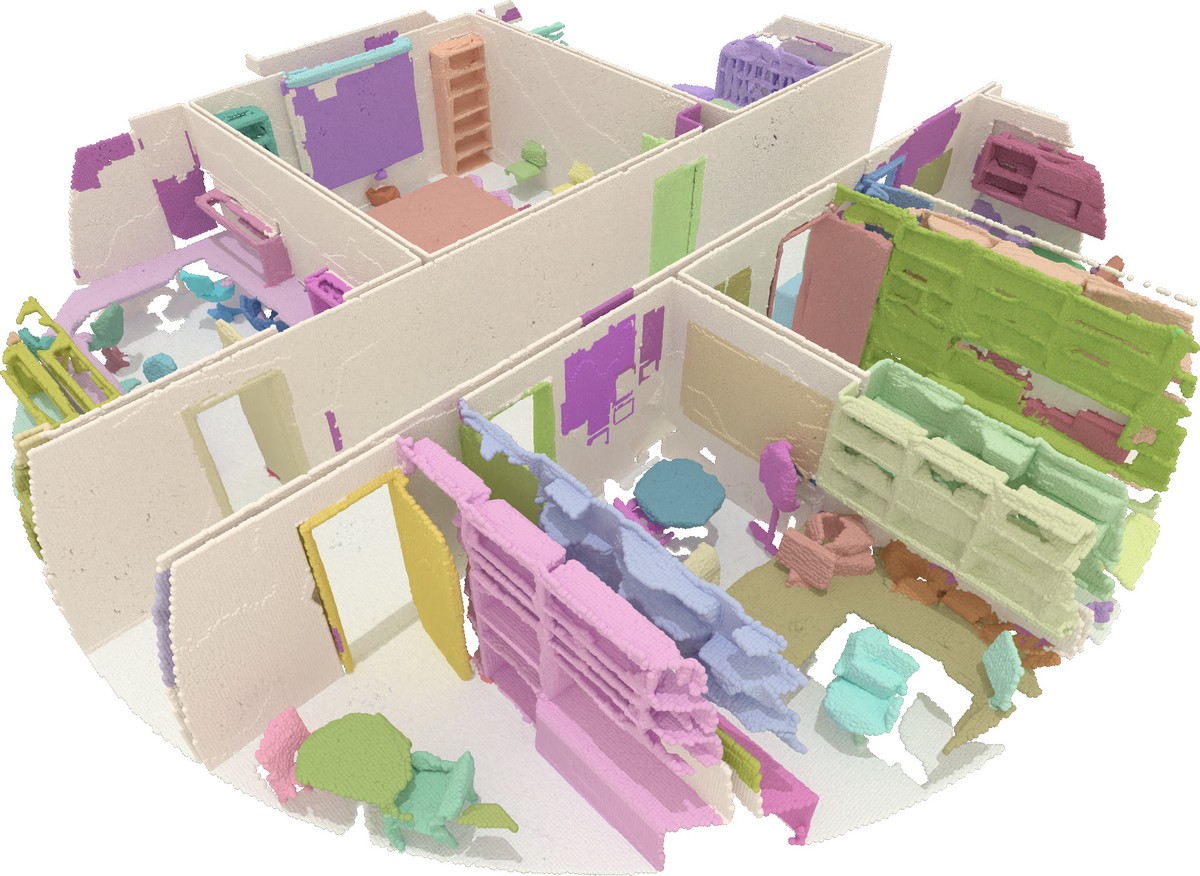}
    &  \includegraphics[width=.23\textwidth, height=.11\textheight]{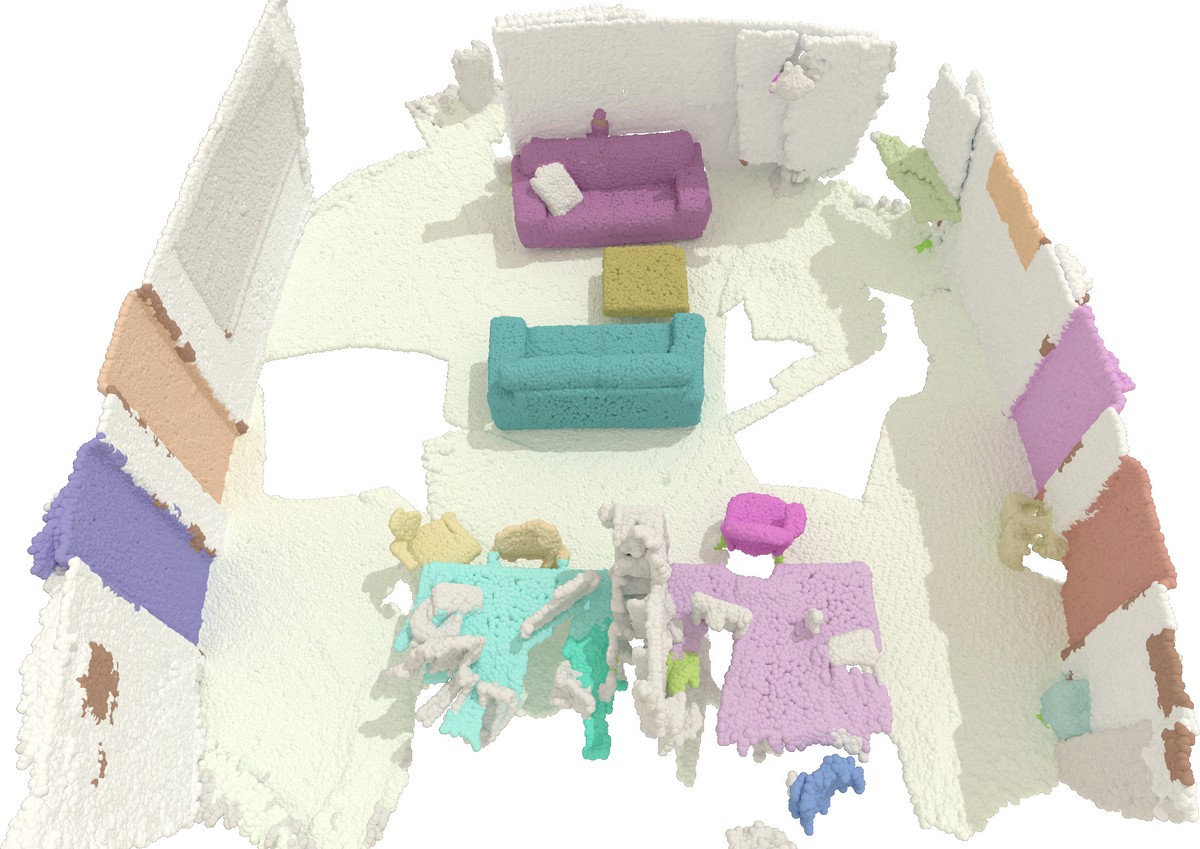}
    &  \includegraphics[width=.23\textwidth, height=.11\textheight]{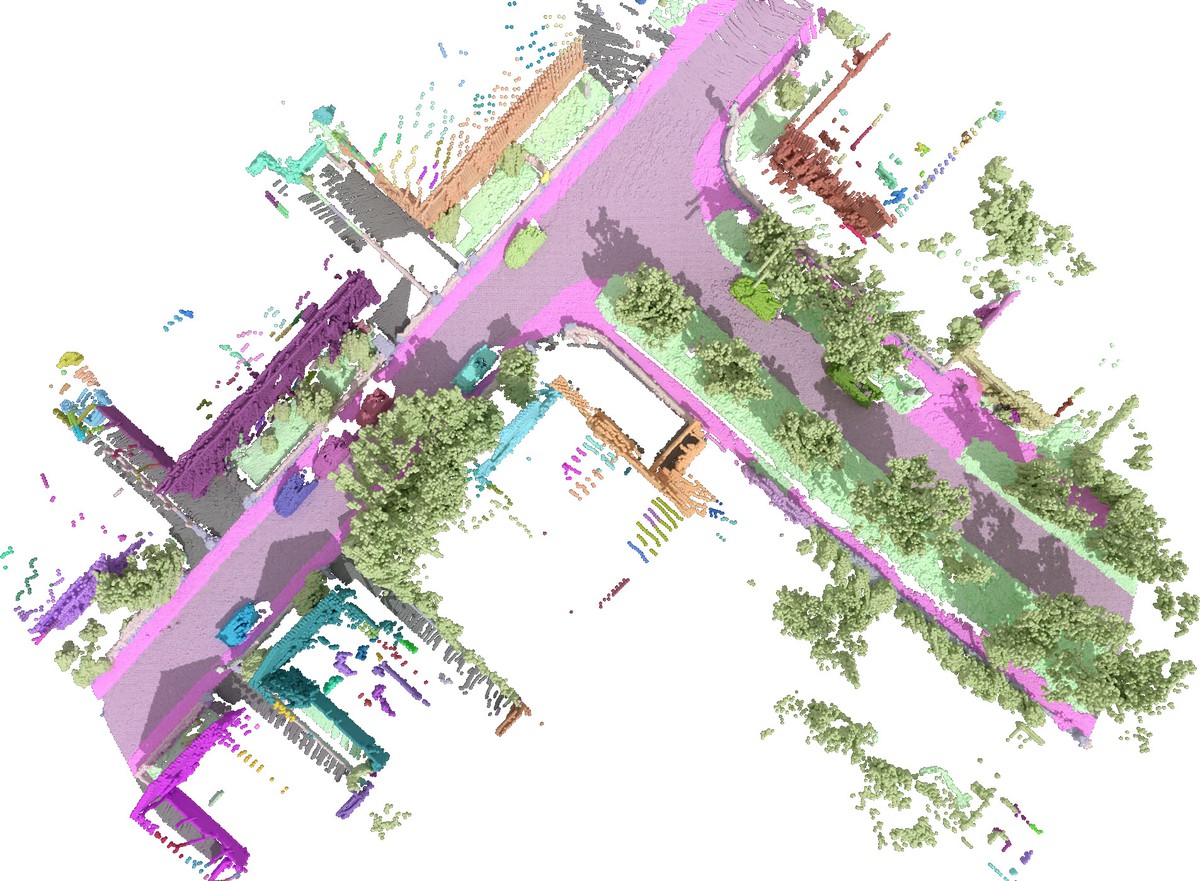}
    &  \includegraphics[width=.23\textwidth, height=.11\textheight]{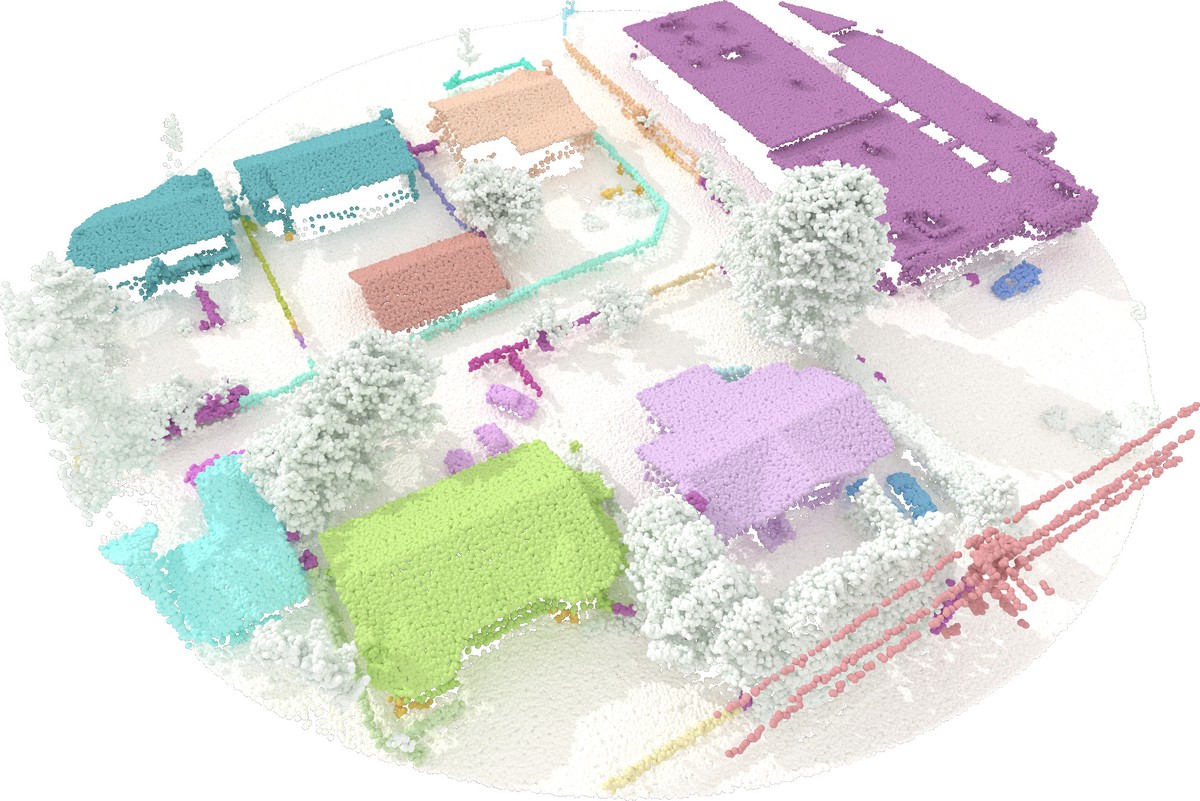}\\
      \end{tabular}
    \caption{{\bf Qualitative Results.} We present the panoptic predictions of our model for the four considered datasets. 
    The scenes' size corresponds to a single batch item during training. ``Stuff'' classes are represented with a lower opacity. \vspace{-5mm}}
    \label{fig:qualitative}
\end{figure*}

We compare our method quantitatively with state-of-the-art models in \tabref{tab:s3dis:f5} to \ref{tab:dales}.
We also report a runtime analysis in \tabref{tab:runtime} and qualitative illustrations in \figref{fig:qualitative}.

\begin{table}
 \caption{{\bf ScanNetv2 Val.} We report the Semantic Segmentation (\emph{SS}) and Panoptic Segmentation (\emph{PS}) performance for various methods on the open test set of ScanNetv2.  $\dagger$ code and models unavailable.}
  \label{tab:scannet}
 \resizebox{\columnwidth}{!}{
\begin{tabular}{llllll}
    \toprule
    & \multicolumn{1}{c}{Size}
    & \multicolumn{1}{c}{SS} & \multicolumn{3}{c}{PS}\\  
     \cmidrule(r{4pt}){2-2} \cmidrule(r{4pt}){3-3} \cmidrule(r{4pt}){4-6}
    &\multicolumn{1}{c}{\small $\times  10^6$} &\multicolumn{1}{c}{mIoU}  & PQ & RQ & SQ \\
     \midrule
    \multicolumn{5}{l}{  Semantic segmentation models}
    \\\midrule
    KPConv \cite{thomas2019kpconv} & 14.1 & 69.2 & - & - & - \\
    Point Trans \cite{zhao2021point} & 7.8 & 70.6 & - & - \\
    Point Trans. v2 \cite{wu2022point} & 11.3 & \bf 75.4 & - & - & -\\
    OctFormer \cite{Wang2023OctFormer} & 44.0 & \bf 75.7  & - & - & -\\ 
    \midrule
    \multicolumn{5}{l}{  Panoptic segmentation models}
    \\
    \midrule
    SceneGraphFusion \cite{wu2021scenegraphfusion,Wald2020} & 2.9
 & - & 31.5 & 42.2 & 72.9 \\
    PanopticFusion \cite{narita2019panopticfusion} & \;$\dagger$ & - & 33.5& 45.3 & 73.0 \\
    \bf SuperCluster (ours) & \bf 1.0 & 66.1 & \bf 58.7 & \bf  69.1 & \bf  84.1  \\
    \bottomrule
\end{tabular}
}
\end{table}

\paragraph{\bf S3DIS.}
We report in \tabref{tab:s3dis:f5} the performance of our algorithm evaluated for Area 5 of the S3DIS dataset.
Compared to several baselines for panoptic segmentation, our model shows a notable improvement with a PQ boost of $+7.8$ points and a mIoU increase of $+3.2$ points.
Remarkably, our model is more than $33$ times smaller than the highest performing model.
Furthermore, we compute panoptic metrics by treating \emph{wall}, \emph{ceiling}, and \emph{floor} as ``stuff'' classes to account for their arbitrary boundaries.
To the best of our knowledge, we are the first to report panoptic results evaluated with 6-Fold cross-validation on S3DIS, in \tabref{tab:s3dis:6f}.

Despite its smaller size, our model achieves high semantic segmentation performance, 
reaching near state-of-the-art performance on the Area~5 and $6$-fold evaluations.

\paragraph{\bf ScanNet.}
As shown in \tabref{tab:scannet}, SuperCluster significantly improves the state-of-the-art of panoptic segmentation by $25.2$ PQ points.
Our model does not perform as well as large networks designed for semantic segmentation but provides decent results with a small backbone of only $1$M parameters.

\begin{table}
 \caption{{\bf KITTI-360.} We report the Semantic Segmentation (\emph{SS}) and Panoptic Segmentation (\emph{PS}) performance for various methods on the open test set of KITTI-360. No panoptic methods were evaluated on this dataset to the best of our knowledge.}
 \label{tab:kitti}
 \resizebox{\columnwidth}{!}{
\begin{tabular}{llllll}
    \toprule
    & \multicolumn{1}{c}{Size}
    & \multicolumn{1}{c}{SS} & \multicolumn{3}{c}{PS}\\  
    \cmidrule(r{4pt}){2-2} \cmidrule(r{4pt}){3-3} \cmidrule(r{4pt}){4-6}
    &\multicolumn{1}{c}{\small $\times 10^6$}&\multicolumn{1}{c}{mIoU}  & PQ & RQ & SQ \\
     \midrule
    \multicolumn{5}{l}{  Semantic segmentation models}
    \\\midrule
    Minkowski  \cite{choy20194minko} & 37.9 & 58.3 & - & - & - \\
    DeepViewAgg \cite{zhao2021point} & 41.2 & 62.1 & - & - & - \\  
    SPT \cite{robert2023efficient} & \bf 0.78  & \bf 63.5 & - & - & - \\  
    \midrule
    \multicolumn{5}{l}{   Panoptic segmentation models}
    \\\midrule
    \bf SuperCluster (ours) & \bf 0.79 & 62.1 & \bf  48.3 & \bf  58.4 & \bf  75.1 \\
    \bottomrule
\end{tabular}
}
\end{table}

\begin{table}
\caption{{\bf DALES.} We report the Semantic Segmentation (\emph{SS}) and Panoptic Segmentation (\emph{PS}) performance for various methods on the open test set of DALES. No panoptic methods were evaluated on this dataset to the best of our knowledge.}
 \label{tab:dales}
 \resizebox{\columnwidth}{!}{
\begin{tabular}{llllll}
    \toprule
    & \multicolumn{1}{c}{Size}
    & \multicolumn{1}{c}{SS} & \multicolumn{3}{c}{PS}\\  
    \cmidrule(r{4pt}){2-2} \cmidrule(r{4pt}){3-3} \cmidrule(r{4pt}){4-6}
    &\multicolumn{1}{c}{\small $\times 10^6$}&\multicolumn{1}{c}{mIoU}  & PQ & RQ & SQ \\
     \midrule
    \multicolumn{6}{l}{\textit{Semantic segmentation models}}
    \\\midrule
    ConvPoint \cite{boulch2020convpoint} & 4.7 & 67.4 & - & - & - \\
    PointNet++ \cite{qi2017pointnetpp} & 3.0 & 68.3 & - & - & - \\
    SPT \cite{robert2023efficient} & \bf 0.21 & 79.6 & - & - & - \\
    KPConv \cite{thomas2019kpconv} & 14.1 & \bf  81.1 & - & - & - \\
    \midrule
    \multicolumn{6}{l}{\textit{Panoptic segmentation models}}
    \\\midrule
    \bf SuperCluster (ours) &\bf  0.21 & 77.3 & \bf  61.2 & \bf  68.6 & \bf  87.1 \\
    \bottomrule
\end{tabular}
}
\end{table}

\paragraph{\bf DALES and KITTI-360.}
SuperCluster is the first capable of processing the large tiles of the DALES and KITTI-360 datasets, thus establishing the first panoptic state-of-the-art for these datasets given in \tabref{tab:kitti} and \tabref{tab:dales}.

\paragraph{\bf Inference and Training Speed.}
In \tabref{tab:runtime}, we compare the inference speed of our approach with state-of-the-art instance and panoptic segmentation algorithms.
As we use a 1080Ti GPU to replicate the setting used to measure most of the approaches' speed (a Titan-X), the values are not entirely comparable.
Still, our model is on par with the fastest methods and offers superior scalability.

None of the reported runtimes include the method's pre-processing times.
Thanks to SPT's efficiency, our entire pre-processing, including the superpoint partition, is faster or equivalent to all existing 3D segmentation methods~\cite{robert2023efficient}.

Our model can be trained in an amount of time comparable to its backbone SPT for semantic segmentation \cite{robert2022learning}.
One fold of S3DIS takes just under $4$ hours, which is substantially quicker than most existing semantic, instance, or panoptic segmentation models.
For example, PointTransformer~\cite{zhao2021point} trains for $63$~h and Stratified Transformer~\cite{lai2022stratified} $216$~GPU-h.
SuperCluster trains on $6$~h on ScanNet, compared to $78$~h for Mask3D~\cite{schult2023mask3d} and $20$~h for ISBNet \cite{ngo2023isbnet}.

\begin{table}
\caption{{\bf Runtime.} We compare the speed of our model to various instance and panoptic segmentation models. %
We report the time spent in the backbone network (first number) and performing panoptic segmentation (second number) on ScanNet Val. scans. $\star$ optional CRF post-processing.}
\label{tab:runtime}
 \resizebox{\columnwidth}{!}{
\begin{tabular}{lll}
    \toprule
    & Hardware & Runtime in ms \\
    \midrule
    \textit{Instance segmentation methods} & & \textit{average per scan on Val} \\
    \midrule
    PointGroup \cite{jiang2020pointgroup} & Titan X & 452 = 128 + 324 \\
    SoftGroup \cite{vu2022softgroup} & Titan X & 345 = 152 + 148 \\
    HAIS \cite{chen2021hierarchical} & Titan X & 339 = 154 + 185 \\
    Mask3D \cite{schult2023mask3d} & Titan X & 339 \\
    ISBNet \cite{ngo2023isbnet} & Titan X & \textbf{237} = 152 + 85 \\
    \bf SuperCluster  (ours) & 1080Ti & \textbf{238} = 193 + 45\\
    \midrule
    \textit{Panoptic segmentation methods} & & \textit{for \texttt{scene0645\_01}}  \\
    \midrule
     PanopticFusion \cite{narita2019panopticfusion} & 2$\times$1080Ti & 485 =  317 + 168 (+ 4500$^\star$) \\
     \bf SuperCluster (ours) & 1080Ti & \textbf{482} = 376 + 106 \\
    \bottomrule
\end{tabular}
}
\end{table}

\subsection{Ablation Study}
\label{sec:ablation}

We evaluate the impact of our design choice by performing several experiments whose results are given in \tabref{tab:ablation}.
More experiments are provided in the Appendix.

\begin{table}
    \caption{{\bf Ablation Study.} We report the performance of different experiments on S3DIS 
    {Area 5}
    with \emph{wall}, \emph{ceiling} and \emph{floor} as ``stuff''.}
    \label{tab:ablation}
    \centering
    \resizebox{.8\linewidth}{!}{
    \begin{tabular}{llll}
        \toprule
        \multirow{2}{*}{Experiment} & \multicolumn{3}{c}{PS} \\
        \cmidrule(r{4pt}){2-4}
        & PQ & RQ & SQ \\
        \midrule
        Best Model            & 58.4 & 68.4 & 77.8 \\
        \midrule
        Constant Edge Weights & 54.2 & 64.2 & 76.6 \\
        Offset Prediction     & 57.1 & 65.2 & 77.1 \\
        Smaller Superpoints   & 56.6 & 64.6 & 78.6 \\
        \midrule
        Superpoint Oracle     & 93.4 & 99.7 & 93.7 \\
        Clustering Oracle     & 83.6 & 91.7 & 90.8 \\
        \bottomrule
    \end{tabular}
    }
\end{table}

\vspace{-3mm}
\paragraph{Constant Edge Weights.}
Replacing all edge weights with a constant value of $1$ yields a drop of $4.2$ PQ points.
This experiment shows the benefit of learning object transitions.

\vspace{-3mm}
\paragraph{Offset Prediction.}
Several \emph{bottom-up} \cite{he2021deep} segmentation approaches \cite{jiang2020pointgroup,xiang2023review,lahoud20193d,han2020occuseg} propose clustering points by shifting their positions towards the predicted position of the object centroid.
To reproduce this strategy, we adjust the position of $x^\text{pos}_s$ in $x$ along a vector that predicts the center of the majority object.
We supervise this prediction with the L1 loss, as it produced the best results among several alternatives that we examined.
Despite our efforts, this approach did not improve the results: $-1.3$ PQ points.
We attribute this to the size diversity of objects observed in large-scale scenes (corridors, buildings), resulting in an unstable prediction.

\vspace{-3mm}
\paragraph{Smaller Superpoints.}
To demonstrate the benefits of using superpoints, we consider a finer partition with $\cS/\cP\sim 15$ instead of $30$.
This requires training with smaller $3$~m cylinders instead of $7$, decreasing the performance by $-1.8$ PQ points.
This result illustrates that the superpoint paradigm is central to our approach.

\vspace{-3mm}
\paragraph{Superpoint Oracle.} 
Using superpoints greatly improves the efficiency and scalability of SuperCluster.
However, since the predictions are made at the superpoint level and never for individual 3D points, the semantic and object purity of the superpoints can restrict the model's performance.
To evaluate this impact, we define the superpoint oracle, which assigns to each superpoint $s$ the class and index of its majority object $\obj(s)$.
The resulting performance provides an upper bound of what our model could potentially achieve.
The high performance of this oracle ($93.4$ PQ) indicates that very little precision is lost by working with superpoints.%

\vspace{-3mm}
\paragraph{Clustering Oracle.}
In a similar vein, we calculate the upper bound of our model by computing the results of the graph clustering with perfect network predictions: $x^\text{class}$ is set as the one-hot-encoding of the class of the majority object, and the object agreement is set to its true value: $a_{p,q}=\hat{a}_{p,q}$.
The performance of this oracle 
($83.6$ PQ) 
shows that our scalable clustering formulation does not significantly compromise the model's precision in its current regime.

\vspace{-3mm}
\paragraph{Limitations.} 
Our approach, while efficient, is not devoid of constraints.
The functional minimized in \eqref{eq:gmpp} is noncontinuous and nondifferentiable, which hinders the computation of gradients and the possibility of learning the partition.
Nevertheless, this aspect lends itself to the speed and simplicity of our training process.
Although our approach can run on diverse acquisition setups, the superpoint partition is sensitive to low point density and may fail for sparse scans as visible on the edge of some KITTI-360 acquisitions.

We use a lightweight SPT network to ensure maximum scalability.
This network, while expressive, is not the most powerful existing architecture.
There is a potential for improved results using more resource-intensive networks.%

Since our model does not improve the semantic segmentation performance of the backbone model (SPT) in any of our experiments, we conclude that our local panoptic supervision scheme does not help semantic segmentation.

\FloatBarrier
\section{Conclusion}

In this paper, we introduced SuperCluster, a novel approach for 3D panoptic segmentation of large-scale point clouds.
We propose a new formulation of this task as a scalable graph clustering problem, bypassing some of the most compute-intensive steps of current panoptic segmentation methods.
Our results across multiple benchmarks, including S3DIS, ScanNet, KITTI-360, and DALES, demonstrate that our model achieves state-of-the-art performance while being significantly smaller, scalable, and easier to train.

Despite the considerable industrial applications, large-scale panoptic segmentation has been relatively unexplored by the 3D computer vision community. We hope that our positive results and the state-of-the-art we established on new datasets and settings will encourage the development of future panoptic approaches for large-scale 3D scans.

\paragraph*{Acknowledgements.} 
This work was funded by ENGIE Lab CRIGEN
and ANR project READY3D ANR-19-CE23-0007.
It used HPC resources from GENCI–IDRIS (Grant 2023-AD011013388R1).

\FloatBarrier
\pagebreak
\pagebreak
{\small
\bibliographystyle{templates/3DV/ieeenat_fullname}
\bibliography{mybib}
}

\ARXIV{
    \FloatBarrier
    \pagebreak
    \section*{\centering \LARGE Appendix}
    \setcounter{section}{0}
    \setcounter{figure}{0}
    \setcounter{table}{0}
    \renewcommand*{\theHsection}{appendix.\the\value{section}}
    \renewcommand\thefigure{\arabic{figure}}
    \renewcommand\thetable{\arabic{table}}
    \renewcommand\thefigure{A-\arabic{figure}}
\renewcommand\thesection{A-\arabic{section}}
\renewcommand\thetable{A-\arabic{table}}
\renewcommand\theequation{A-\arabic{equation}}
\renewcommand\thealgorithm{A-\arabic{algorithm}} 

\ARXIV{
    \begin{figure*}
        \begin{center}
\centering

\begin{tabular}{@{}cccc@{}}
\includegraphics[width=.23\linewidth]{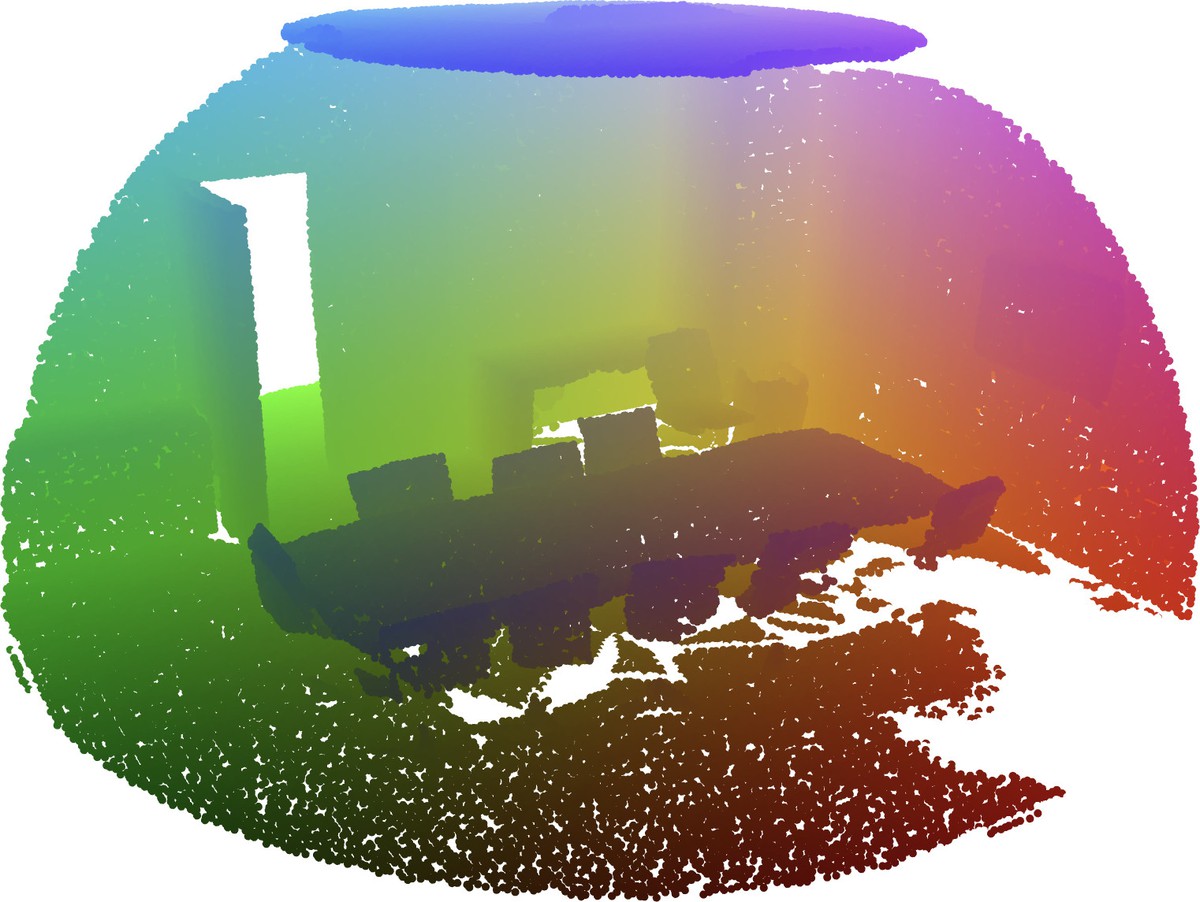}
&
\includegraphics[width=.23\linewidth]{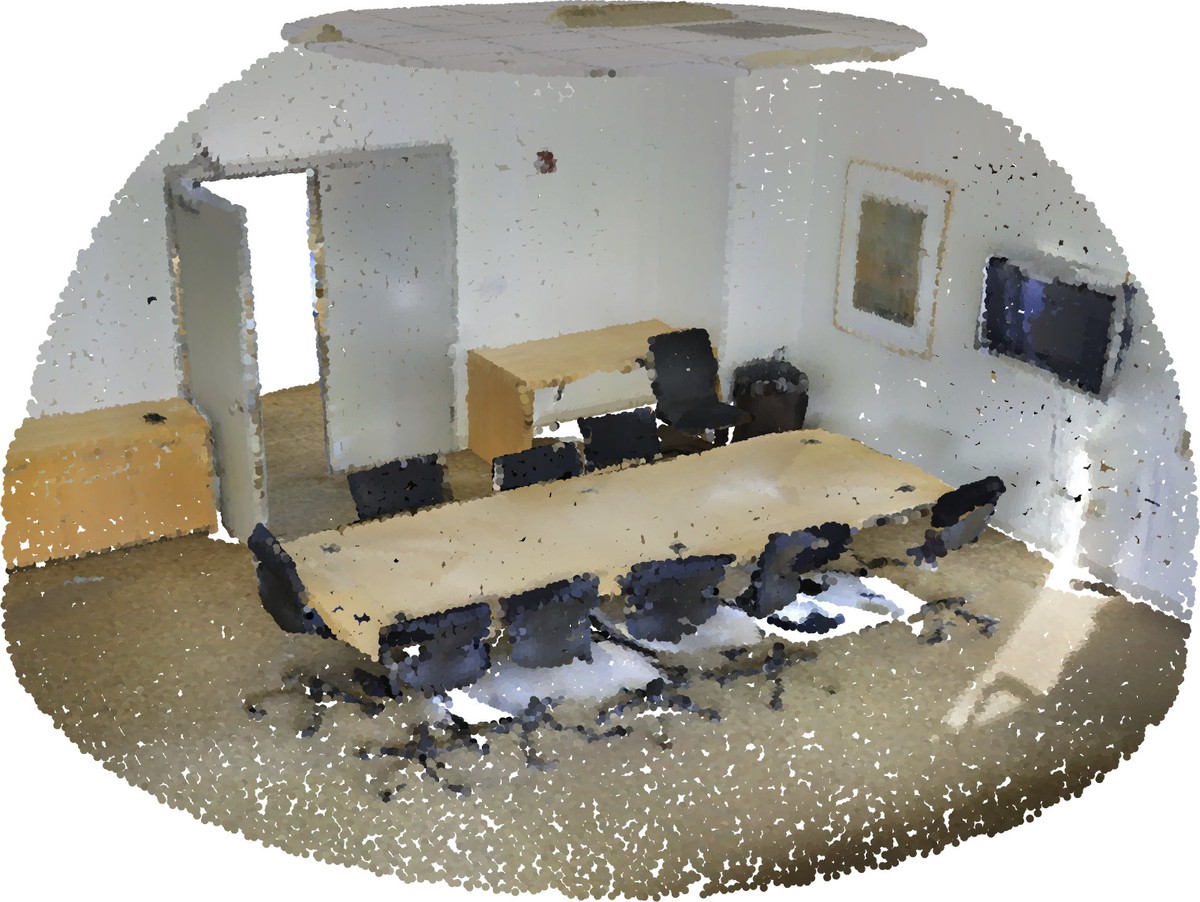}
&
\includegraphics[width=.23\linewidth]{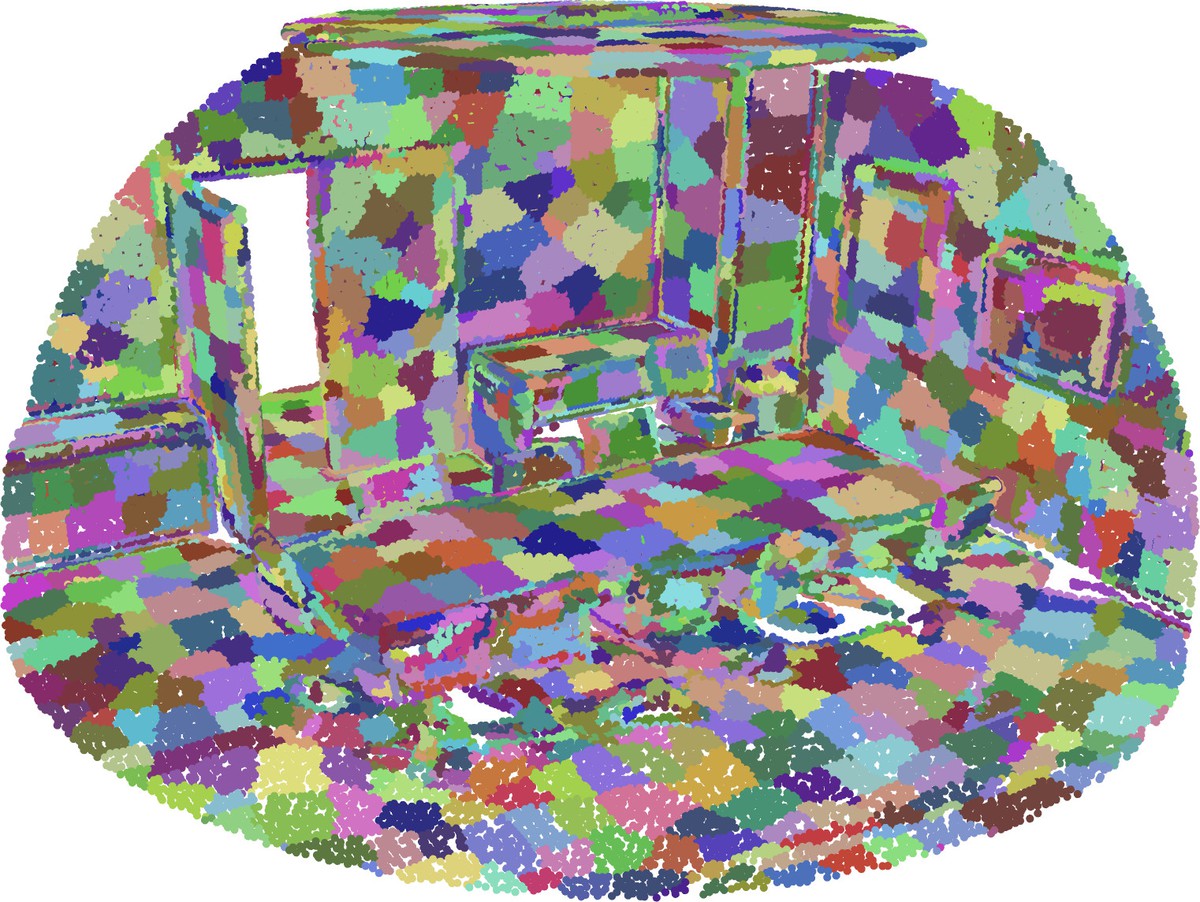}
&
\includegraphics[width=.23\linewidth]{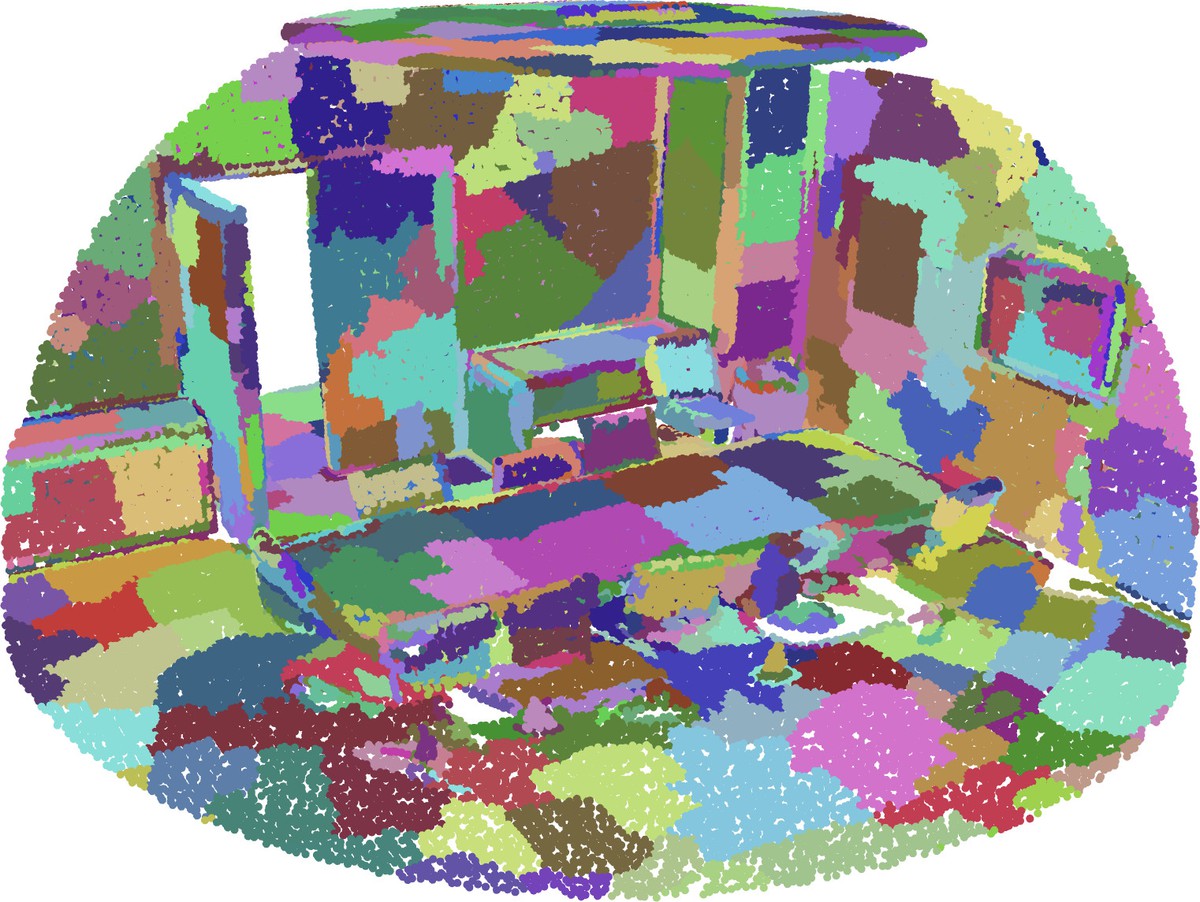}
\\
a) Position & b) RGB   & c) Level-1  & d) Level-2  
\\
\includegraphics[width=.23\linewidth]{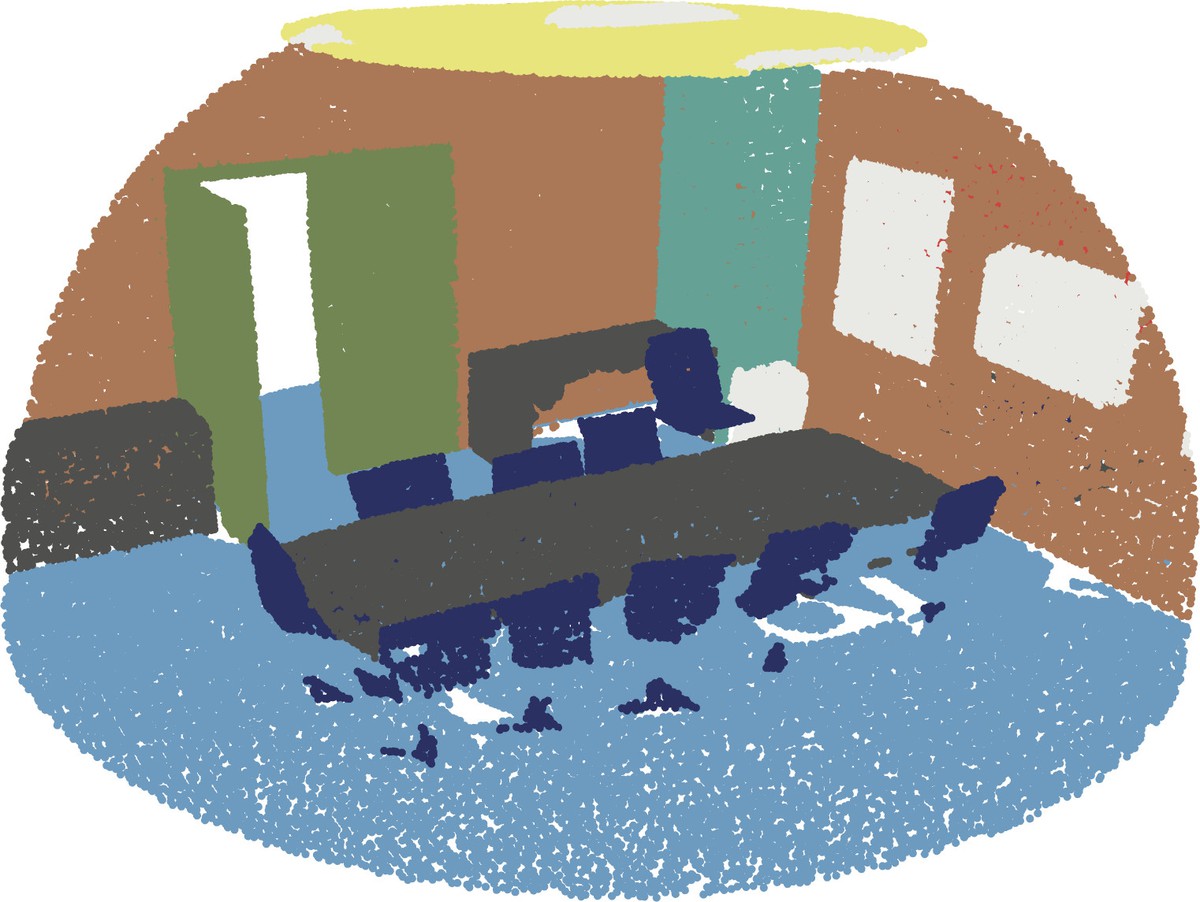}
&
\includegraphics[width=.23\linewidth]{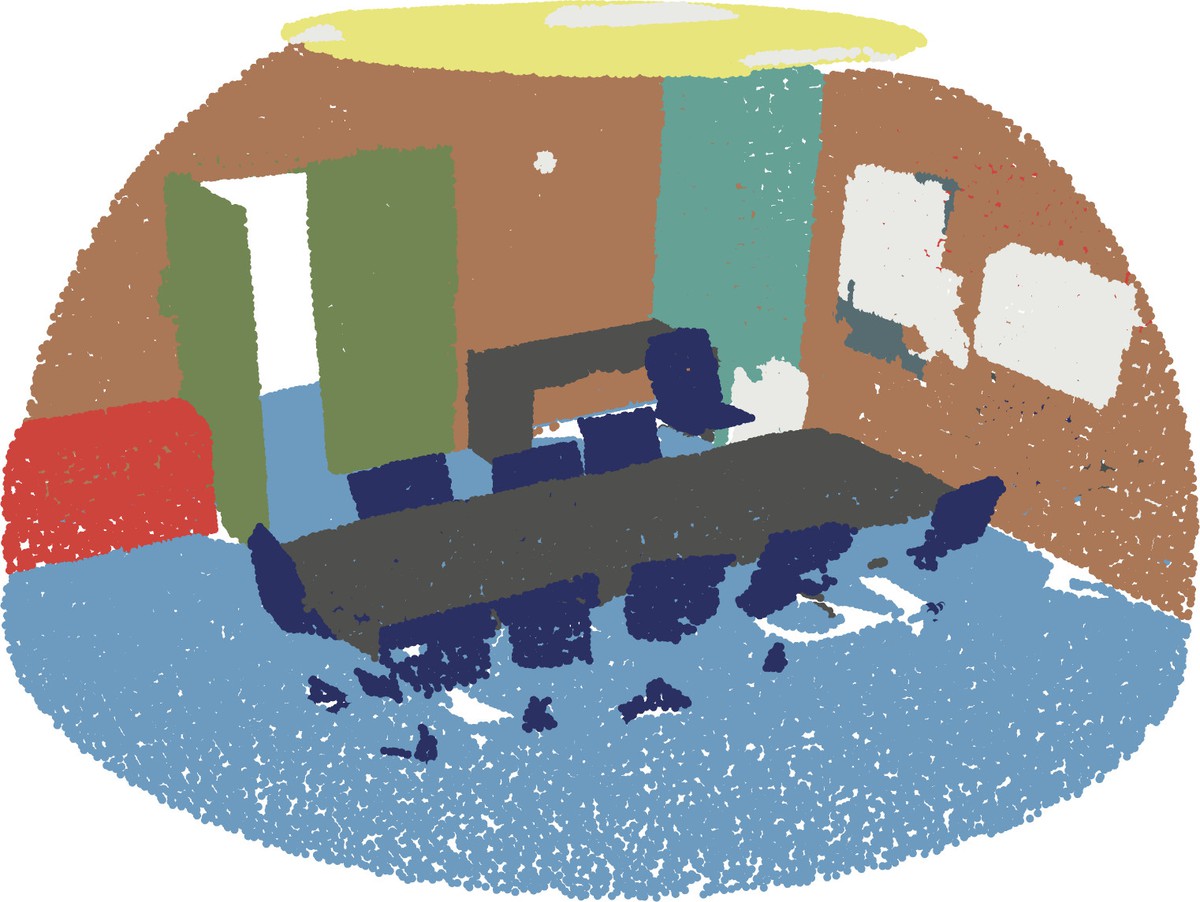}
&
\includegraphics[width=.23\linewidth]{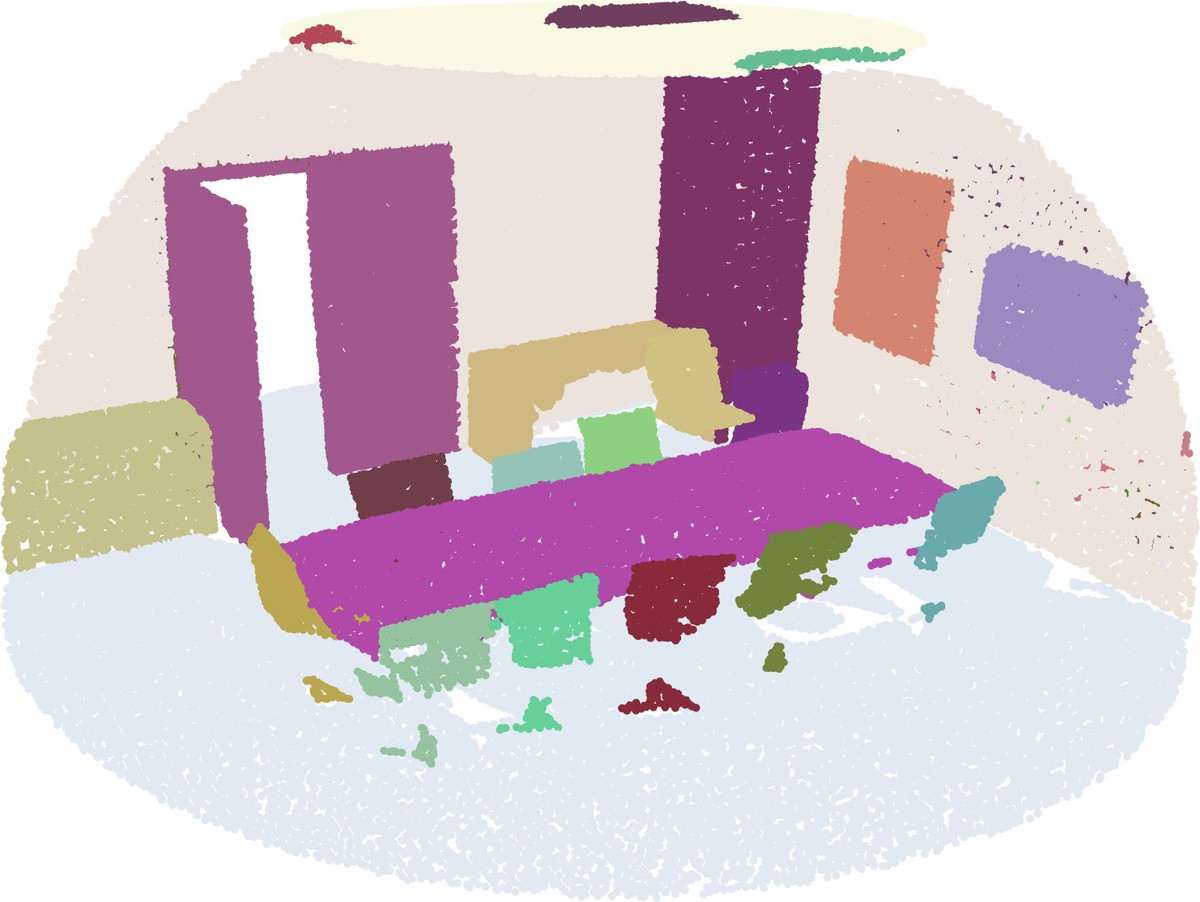}
&
\includegraphics[width=.23\linewidth]{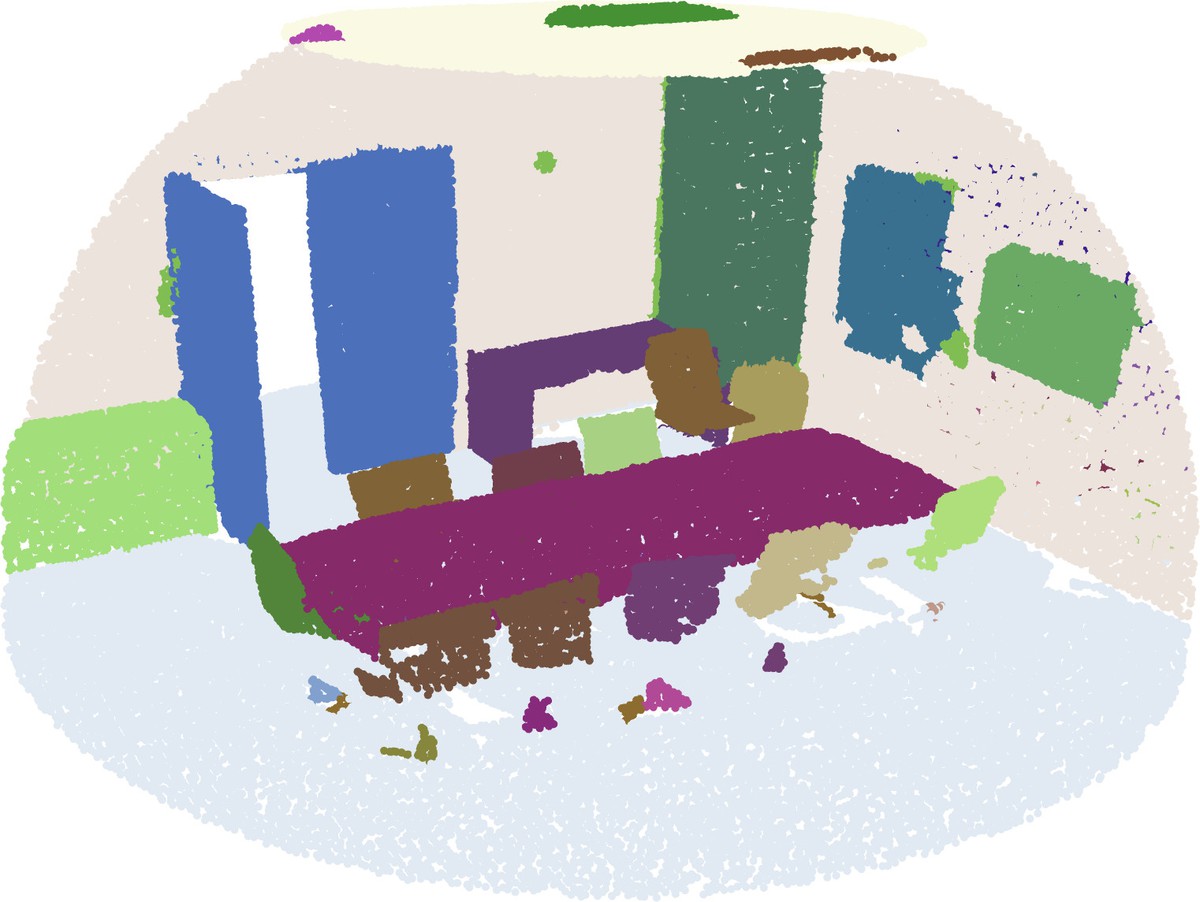}
\\
e) Semantic Annotations  & f) Semantic Predictions & g) Panoptic Annotations  & h) Panoptic Predictions
\end{tabular}
\captionsetup{type=figure}
\captionof{figure}{{\bf Interactive Visualization.} Our interactive viewing tool allows for the manipulation and visualization of point cloud samples colorized according to their position (a), radiometry (b), partition level (c,d), semantic annotations (e) and predictions (f), and panoptic annotations (g) and predictions (h).}
 \label{fig:visu}
\end{center}

    \end{figure*}
}{}

In this appendix, we introduce our interactive visualization tool in \secref{sec:visu}, our source code in \secref{sec:code}.
We provide details about the superpoint encoder backbone SPT in \secref{sec:spt}
and implementation details in \secref{sec:implem}.
In \secref{sec:instance}, we discuss the evaluation of \SHORTHAND on instance segmentation, 
before comparing in \secref{sec:hungarian} the scalability of the Hungarian algorithm with our graph clustering formulation.
We then provide an illustration of how many points can be segmented at once with SuperCluster in \secref{sec:max_batch}.
Finally, we provide detailed class-wise results and illustrate the colormaps of each dataset in \secref{sec:classwise}.

\section{Interactive Visualization}
\label{sec:visu}

Our project page \url{https://drprojects.github.io/supercluster} offers interactive visualizations of our method.
As shown in \figref{fig:visu}, we can visualize samples from the datasets with different point attributes and from any angle.
These visualizations were instrumental in designing and validating our model; we hope that they will also facilitate the reader's understanding.

\section{Source Code}
\label{sec:code}

 We make our source code publicly available at \GITHUB.
Our method is developed in PyTorch and relies on PyTorch Geometric, PyTorch Lightning, and Hydra.

\section{Superpoint-Based Backbone}
\label{sec:spt}

As mentioned in Section \textcolor{red}{$3.3$}, our panoptic segmentation method conveniently extends to superpoint-based methods.
In particular, we discuss here our choice of using Superpoint Transfomer~\cite{robert2023efficient} for both computing superpoint partitions and learning superpoint features.

\paragraph{Superpoint Transformer.} 
Superpoint Transformer (SPT) is a superpoint-based transformer architecture for the efficient semantic segmentation of large-scale 3D point clouds.
The authors propose a fast algorithm to build a hierarchical superpoint partition, whose implementation runs $7$ times faster than previous superpoint-based approaches.
Additionally, SPT relies on a self-attention mechanism to capture the relationships between superpoints at multiple scales, achieving state-of-the-art performance on S3DIS, KITTI-360, and DALES.

As a memory- and compute-efficient approach capable of producing superpoint representations for very large scenes, we found Superpoint Transformer to be the ideal backbone for scalable 3D panoptic segmentation endeavor.

\paragraph{Alternative Partitions.}
Alternative methods could be considered for computing the superpoint partition.  Clustering-based methods such as VCCS~\cite{papon2013voxel} draw inspiration from SLIC~\cite{achanta2012slic} and use k-means on point features, under local adjacency constraints.
However, these k-means-based methods rely on a fixed number of randomly initialized clusters, proscribing the processing of point clouds of arbitrary size and geometric complexity.

On the other hand, we use the implementation proposed in SPT, which itself derives from Landrieu~\etal~\cite{landrieu2018spg}.
These papers cast point cloud oversegmentation as a structured optimization problem and use the cut-pursuit~\cite{landrieu2016cut} algorithm to generate superpoints.
This scalable approach does not make any assumption on the number of superpoints and produces a partition whose granularity adapts to the 3D geometry.

\paragraph{Alternative Backbones.}
One may consider different architectures to produce superpoint-wise or point-wise representations.
Other methods for embedding superpoints exist~\cite{landrieu2018spg,landrieu2019ssp,hui2021superpoint,thyagharajan2022segment,kang2023region}, but Robert~\etal~\cite{robert2023efficient} demonstrates superior performance and efficiency.

Alternatively, one may choose to adopt a per-point paradigm and rely on established models such as KPConv~\cite{thomas2019kpconv}, MinkowskiNet~\cite{choy20194minko}, Stratified Transformer~\cite{lai2022stratified}, or PointNeXt~\cite{qian2022pointnext}.
Although expressive, these models are memory and compute-intensive and can only handle small point clouds at once.
For example, in an indoor setting such as S3DIS or ScanNet, SPT can process entire buildings as a whole, while these methods can only handle a few rooms simultaneously, which limits their applicability for large-scale panoptic segmentation.

\section{Implementation Details}
\label{sec:implem}

\begin{table}
\caption{\textbf{Graph Clustering Parameters.} We provide the graph clustering parameters used for each dataset.
}
\label{tab:graph_clustering_parameters}
\centering
\small{
\begin{tabular}{@{}lccc@{}}
    \toprule
    Dataset               & $\lambda$ & $\eta$ & $\epsilon$ \\
    \midrule
    S3DIS                 & $10$ & $5.10^{-2}$ & $10^{-4}$ \\
    S3DIS  - no ``stuff'' & $20$ & $5.10^{-2}$ & $10^{-4}$ \\
    ScanNet               & $20$ & $5.10^{-2}$ & $10^{-4}$ \\
    KITTI-360             & $10$ & $5.10^{-2}$ & $10^{-4}$ \\
    DALES                 & $20$ & $5.10^{-2}$ & $10^{-4}$ \\
    \bottomrule
\end{tabular}}
\end{table}

In this section, we provide the exact parameterization of the \SHORTHAND architecture used for our experiments. For simplicity, we represent an MLP by the list of its layer widths: $[\text{in\_channels}, \text{hidden\_channels}, \text{out\_channels}]$. 

\paragraph{Backbone.} Our backbone model is Superpoint Transformer \cite{robert2023efficient} with minor modifications, described below. We use SPT-64 for S3DIS and DALES and SPT-128 for KITTI360 and ScanNet.

For all datasets, we reduce the output dimension of the point encoder $\phi^0_\text{enc}$ from $128$ to $64$~\cite{robert2023efficient}. We find that this does not affect SPT performance while reducing its memory requirements.
For ScanNet, we find that using $32$ heads instead of $16$ and setting $D_\text{adj}=64$ instead of $32$~\cite{robert2023efficient} improves the performance. 

\paragraph{Object Agreement Head.} The object agreement prediction head $\phi^{\text{object}}$ is a normalization-free MLP with LeakyReLU activations~\cite{maas2013leakyrelu} and layers $[2D, 32, 16, 1]$, where $D$ is the output feature dimension of the backbone (\textit{i.e.} $64$ for S3DIS and DALES, and $128$ for ScanNet and KITTI-360).

\paragraph{Graph Clustering.}
As mentioned in Section \textcolor{red}{$4.1$}, since our supervision framework does not require solving the clustering problem of Equation 1 during training, we tune the graph clustering parameters on the train set only after training.
In \tabref{tab:graph_clustering_parameters} we detail the tuned parameters for each dataset. 

\section{Instance Segmentation Evaluation}
\label{sec:instance}

While methods predicting a panoptic segmentation could also provide an instance segmentation, we argue that their instance segmentation metrics are not directly comparable to the ones of methods dedicated to instance segmentation.

Firstly, instance segmentation metrics allow overlap between proposals, and not all points need to be in a predicted instance.
Thus, instance segmentation methods can predict multiple instances per true object and avoid predicting in ambiguous or complex areas.
In contrast, panoptic segmentation methods assign exactly one object label to each point~\cite{kirillov2019panoptic}.

Secondly, instance segmentation metrics require a confidence score for each proposal, which has a substantial influence on performance \cite[4.2.2]{jiang2020pointgroup}.
Typically, instance segmentation methods learn this score with a dedicated network \cite{jiang2020pointgroup}. 
While Mask3D~\cite{schult2023mask3d} derives this score from the semantic and mask confidences of the prediction, these are supervised by an explicit matching between the true and proposed instances.
\SHORTHAND is precisely designed to avoid this matching step and never explicitly builds instances during training.

In summary, while it is technically possible to evaluate the predictions of \SHORTHAND with instance segmentation metrics, their comparison with dedicated methods would not be fair.
A more equitable evaluation of \SHORTHAND on instance segmentation would require a dedicated post-processing step and an instance scoring mechanism, which falls outside the scope of this paper.

\section{Scalability of Matching Step}
\label{sec:hungarian}

In this section, we provide an experimental evaluation of the cost of the matching step using the Hungarian algorithm.

\paragraph{Experimental Protocol.} We measure the time it takes our method to perform the panoptic segmentation step (Equation 1) for scenes of various sizes.
Our goal is to compare this processing time with the matching step of conventional approaches.
Given that we deal with scenes containing a large number of objects, far beyond what the encoders of these methods can handle in inference, we generate synthetic cost matrices to simulate the matching process.
We compute these matrices for different combinations of the number of true objects $\texttt{n\_true}$ and of proposals $\texttt{n\_pred}$.
To generate realistic cost matrices, each proposed instance has a nonzero random cost for at most $3$ true objects.
We then measure the time taken by the Hungarian algorithm to solve the assignment.

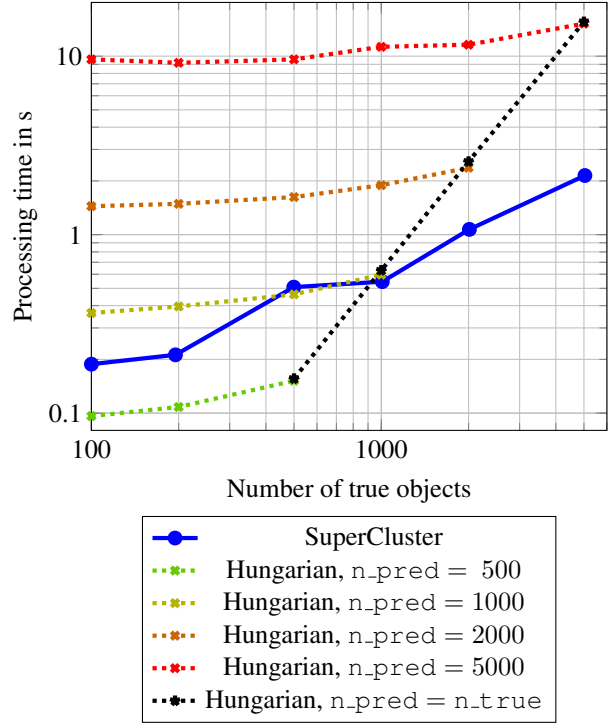
\begin{figure}

\begin{tikzpicture}
\begin{axis}[
    xlabel={Number of true objects},
    ylabel={Processing time in s},
    ylabel style={yshift=-10pt}, 
    xmode=log,
    ymode=log,
    xtick={100, 1000, 10000},
    xticklabels={100, 1000, 10000},
    ytick={0.1, 1, 10},
    yticklabels={0.1, 1, 10},
    xmin=100,
    xmax=6000,
    ymin=0.08,
    ymax=20,
    grid=both,
    legend style={at={(0.5,-0.2)},anchor=north},
]

 \addplot[mark=*,color=blue, ultra thick] coordinates {
    (100, 0.188) 
    (195, 0.212) 
    (499, 0.508) 
    (1008, 0.545) 
    (2013, 1.070) 
    (5039, 2.144)
};
\addlegendentry{\SHORTHAND}

\definecolor{HUNGARIANCOLORA}{RGB}{0, 204, 0}
\definecolor{HUNGARIANCOLORB}{RGB}{102, 204, 0}
\definecolor{HUNGARIANCOLORC}{RGB}{180, 180, 0}
\definecolor{HUNGARIANCOLORD}{RGB}{204, 102, 0}
\definecolor{HUNGARIANCOLORE}{RGB}{255, 0, 0}
\definecolor{HUNGARIANCOLORF}{RGB}{0, 0, 0}

\addplot[mark=x,color=HUNGARIANCOLORB, dotted, ultra thick,mark options=solid] coordinates {
    ( 100, 0.0962)
    ( 200, 0.1082)
    ( 500, 0.1518)
};
\addlegendentry{Hungarian, $\texttt{n\_pred}=~500$}

\addplot[mark=x,color=HUNGARIANCOLORC, dotted, ultra thick,mark options=solid] coordinates {
    ( 100, 0.3638)
    ( 200, 0.3958)
    ( 500, 0.4625)
    (1000, 0.5964)
};
\addlegendentry{Hungarian, $\texttt{n\_pred}=1000$}

\addplot[mark=x,color=HUNGARIANCOLORD, dotted, ultra thick,mark options=solid] coordinates {
    ( 100, 1.4425)
    ( 200, 1.4874)
    ( 500, 1.6234)
    (1000, 1.8910)
    (2000, 2.3705)
};
\addlegendentry{Hungarian, $\texttt{n\_pred}=2000$}

\addplot[mark=x,color=HUNGARIANCOLORE, dotted, ultra thick,mark options=solid] coordinates {
    ( 100, 9.5925)
    ( 200, 9.1767)
    ( 500, 9.5936)
    (1000, 11.2798)
    (2000, 11.5908)
    (5000, 15.2139)
};
\addlegendentry{Hungarian, $\texttt{n\_pred}=5000$}

\addplot[mark=star,color=HUNGARIANCOLORF, dotted, ultra thick,mark options=solid] coordinates {
    ( 500, 0.1560)
    (1000, 0.6307)
    (2000, 2.5663)
    (5000, 15.5615)
};
\addlegendentry{Hungarian, $\texttt{n\_pred}=\texttt{n\_true}$}

\end{axis}
\end{tikzpicture}
\caption{{\bf Cost of Matching Step.}
We plot the time taken by our method to perform the panoptic segmentation step for scenes with various numbers of true objects.
We also show the time necessary for the Hungarian algorithm to perform the matching for different numbers of maximum instance proposals $\texttt{n\_pred}$ and true instances $\texttt{n\_true}$.}
\label{fig:sup:hungarian}
\end{figure}

\paragraph{Analysis.} We report the results of this experiment in \figref{fig:sup:hungarian}.
One significant advantage of our approach, denoted \SHORTHANDX, is that it does not require a predefined maximum number of detected objects.
In contrast, the processing time of the Hungarian algorithm is strongly affected by this parameter. Attempting to predict too many instances can lead to significantly prolonged training times, regardless of the number of true objects.
Even in the idealized setting of $\texttt{n\_pred}$ equals to $\texttt{n\_true}$, the Hungarian algorithm becomes much slower than our method when the number of true objects exceeds $1000$.
We also remind the reader that approaches relying on matching-based supervision must perform this step at each training iteration, while \SHORTHAND only solves Equation $1$ once during inference and never during training.

\section{Large-scale Inference}
\label{sec:max_batch}

\begin{figure*}
    \centering
    \begin{tikzpicture}
        \node[anchor=south west,inner sep=0] at (0,0) {\includegraphics[width=0.95\linewidth]{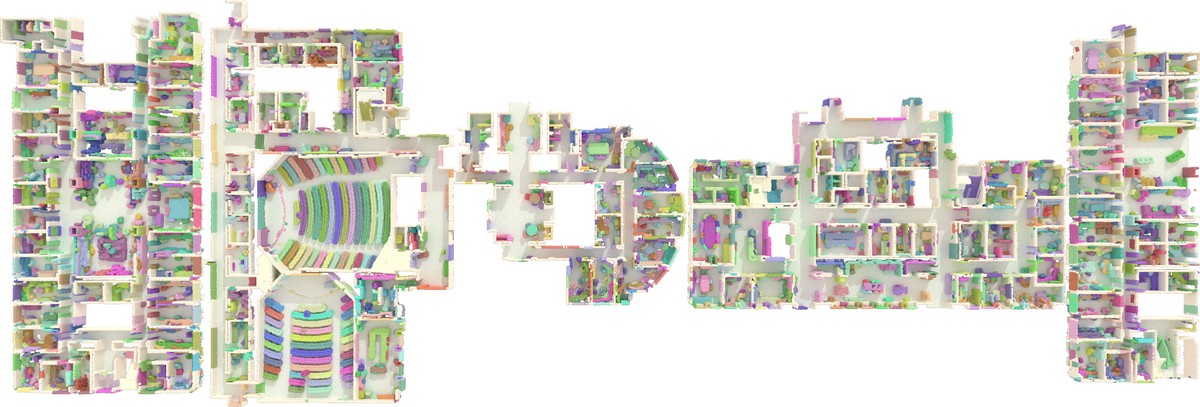}};
         \node[draw=none] at (1.6, -0.3) {\large\bf Area~1};
         \node[draw=none] at (4.5, -0.3) {\large\bf Area~2};
         \node[draw=none] at (8.1, -0.3) {\large\bf Area~3};
         \node[draw=none] at (12.2, -0.3) {\large\bf Area~4};
         \node[draw=none,text width=1cm] at (15.6, -0.3) {\large\bf Area~6 (65\%)};
    \end{tikzpicture}
    \caption{{\bf Large-Scale Inference on S3DIS.} 
    Largest scan that SuperCluster can segment in one inference on an A40 GPU: \\ 
    $\textbf{4.6}$ areas, $\mathbf{21.3}$\textbf{M} points, $\mathbf{646}$\textbf{k} superpoints, $\mathbf{5298}$ target objects, and $\mathbf{4565}$ predicted objects. Inference takes $\bf 7.4$ \textbf{seconds}.}
    \label{fig:max_batch_s3dis}
\end{figure*}

\begin{figure*}
    \centering
    \includegraphics[width=\linewidth]{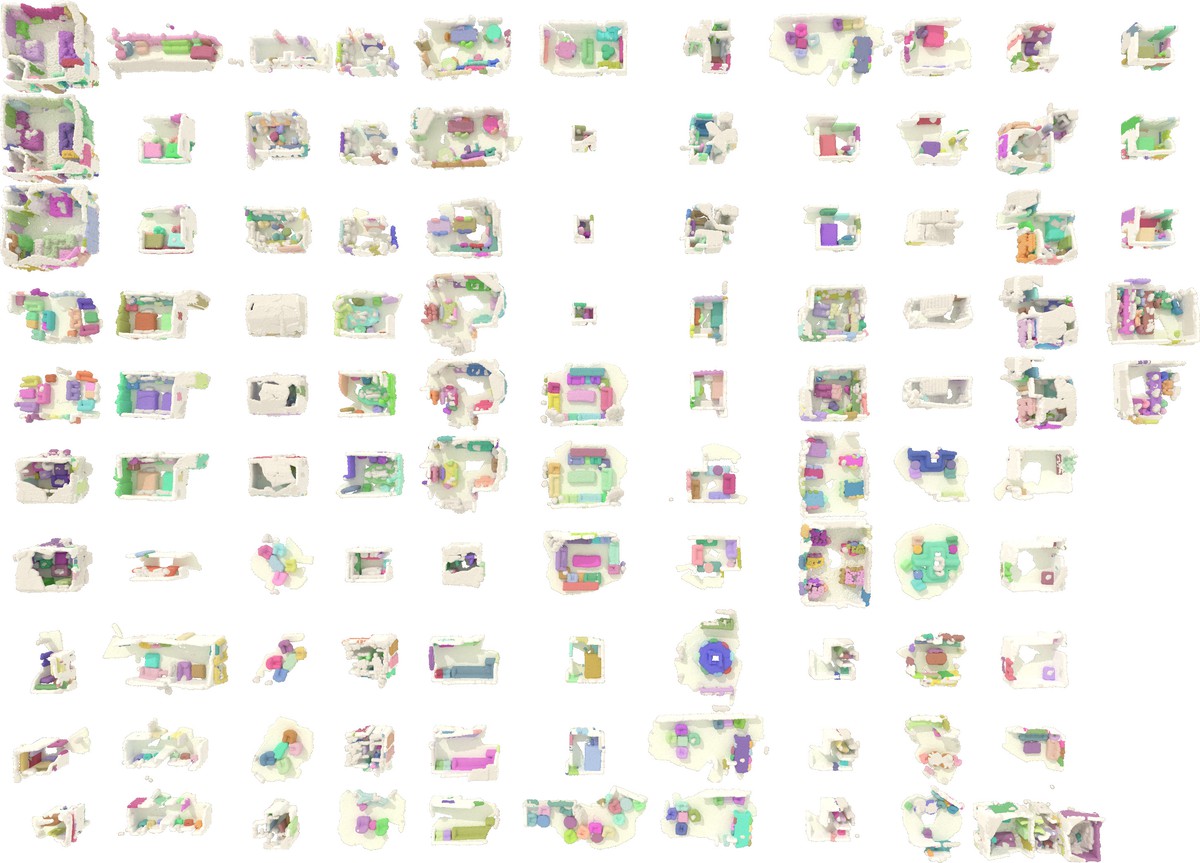}
    \caption{{\bf Large-Scale Inference on ScanNet.} Largest number of scans that SuperCluster can segment in one inference on an A40 GPU: \\
    $\textbf{105}$ scans , $\textbf{10.9}$\textbf{M} points, $\textbf{398}$\textbf{k} superpoints , $\textbf{1683}$ target objects, and $\textbf{2148}$ predicted objects. The inference takes $\bf 6.8$ \textbf{seconds}.}
    \label{fig:max_batch_scannet}
\end{figure*}

\begin{figure*}
    \centering
    \includegraphics[width=\linewidth]{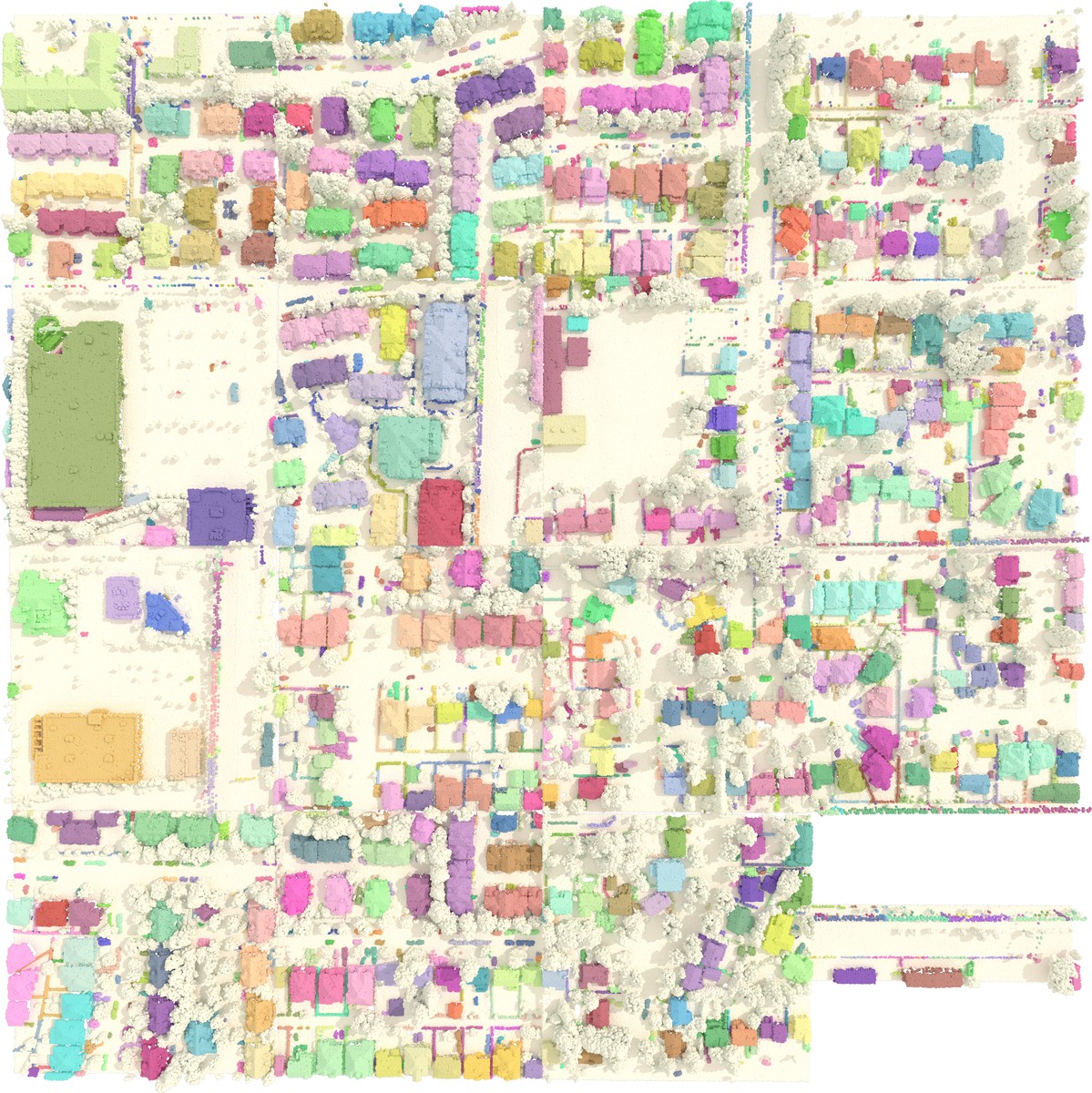}
    \caption{{\bf Large-Scale Inference on DALES.} {
    Largest scan that SuperCluster can segment in one inference on an A40 GPU:\\
    $\textbf{15.3}$ tiles, $\textbf{7.8}$~\textbf{km}$\bf^2$, $\textbf{18.0}$\textbf{M} points, $\textbf{589}$\textbf{k} superpoints, $\textbf{1727}$ target objects, and $\textbf{1559}$ predicted objects. Inference takes $\bf 10.1$ \textbf{seconds}.
 }}
    \label{fig:max_batch_dales}
\end{figure*}

\begin{figure*}
    \centering
    \includegraphics[width=\linewidth]{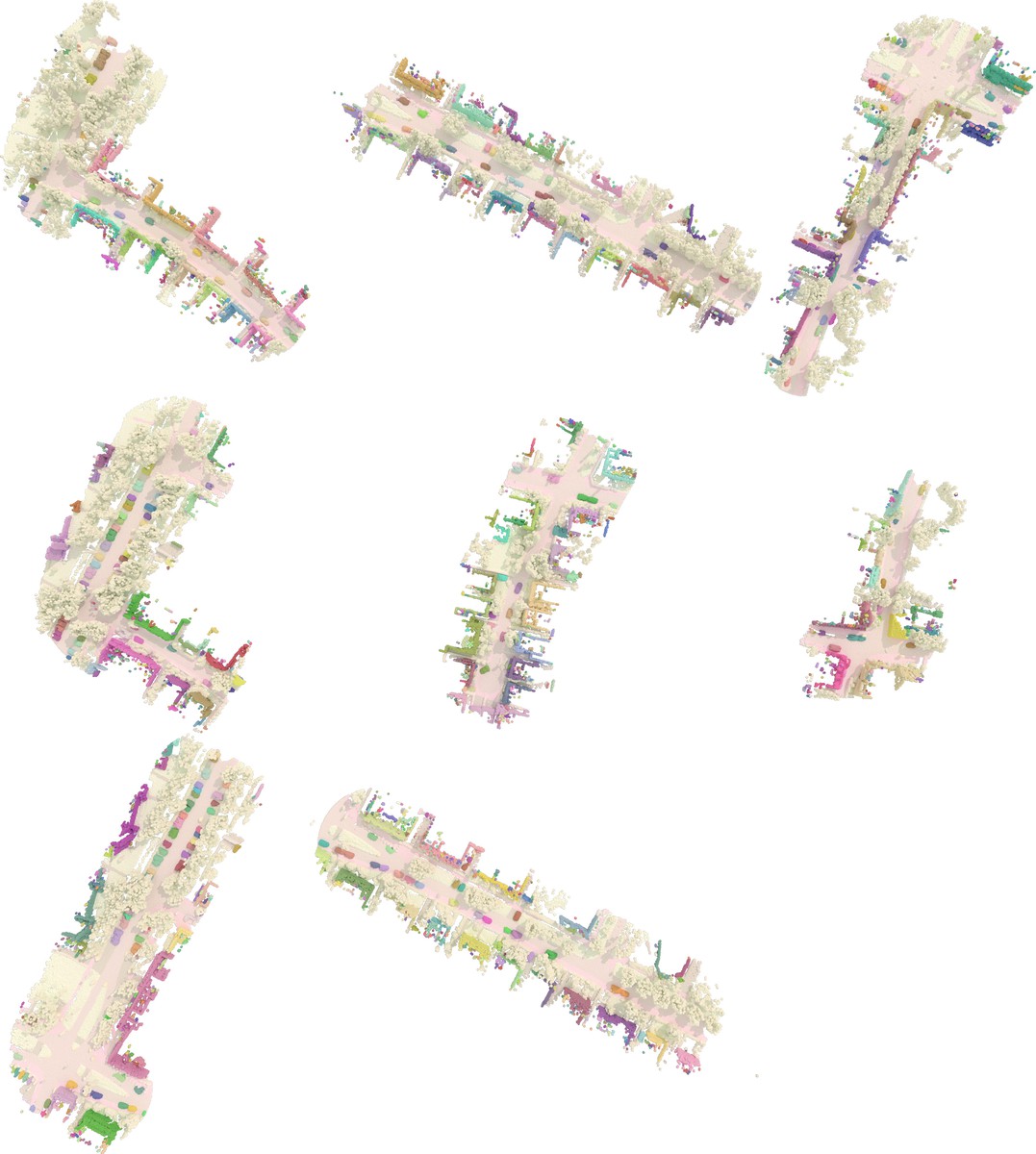}
    \caption{{\bf Large-Scale Inference on KITTI-360.} {Largest scan that SuperCluster can segment in one inference on an A40 GPU: \\
    $\bf 7.5$ tiles, $\bf 11.0$\textbf{M} points, $\bf 414$\textbf{k} superpoints, $\bf 602$ target objects, and $\bf 1947$ predicted objects. Inference takes $\bf 6.6$ \textbf{seconds}.
    }}
    \label{fig:max_batch_kitti360}
\end{figure*}

Our method can process large 3D point clouds with just one inference. In this section, we represent the largest portion of each dataset that \SHORTHAND can handle in one inference with an A40 GPU (48G of VRAM).
Results for each dataset are presented in \figref{fig:max_batch_s3dis}, \figref{fig:max_batch_scannet}, \figref{fig:max_batch_dales}, and \figref{fig:max_batch_kitti360}.

\section{Detailed Results}
\label{sec:classwise}

\begin{table*}[t]
\caption{{\bf S3DIS Class-wise Performance.} We report the average and per-class panoptic quality (PQ), recognition quality (RQ), segmentation quality (SQ), precision (Prec), and recall (Rec) performance of \SHORTHAND on S3DIS. 
We indicate ``stuff'' classes with $\dagger$.}
\label{tab:classwise_s3dis} 
\begin{center}
\footnotesize{

    \begin{tabular}{lc*{14}{c}}
        
        \toprule
        \multicolumn{15}{c}{S3DIS Area~5} \\
        Metric & Avg. & ceiling $\dagger$ & floor $\dagger$ & wall $\dagger$ & beam & column & window & door & chair & table & bookcase & sofa & board & clutter \\
        \midrule

        PQ     & 58.4 & 93.8 & 96.2 & 84.0 & 0.0 & 48.5 & 64.7 & 45.3 & 64.3 & 40.1 & 47.7 & 62.9 & 70.5 & 40.6 \\
        RQ     & 68.4 & 100.0 & 100.0 & 100.0 & 0.0 & 61.0 & 77.2 & 57.4 & 76.4 & 52.6 & 55.8 & 78.3 & 81.1 & 49.7 \\
        SQ     & 77.8 & 93.8 & 96.2 & 84.0 & 0.0 & 79.5 & 83.9 & 78.9 & 84.1 & 76.2 & 85.4 & 80.4 & 86.9 & 81.7 \\
        Prec.  & 71.4 & 100.0 & 100.0 & 100.0 & 0.0 & 64.2 & 79.6 & 59.8 & 75.7 & 53.3 & 66.0 & 75.0 & 93.8 & 60.5 \\
        Rec.   & 66.2 & 100.0 & 100.0 & 100.0 & 0.0 & 58.1 & 75.0 & 55.1 & 77.1 & 52.0 & 48.4 & 81.8 & 71.4 & 42.2 \\
        \\\midrule

        \multicolumn{15}{c}{S3DIS 6-FOLD} \\
        \midrule~\\

        PQ     & 62.7 & 93.8 & 95.2 & 84.1 & 58.9 & 64.7 & 70.2 & 41.1 & 48.0 & 45.5 & 45.8 & 55.7 & 64.3 & 47.2 \\
        RQ     & 73.2 & 100.0 & 100.0 & 100.0 & 67.9 & 78.2 & 81.8 & 55.3 & 57.6 & 58.6 & 54.9 & 65.3 & 74.7 & 56.8 \\
        SQ     & 84.7 & 93.8 & 95.2 & 84.1 & 86.8 & 82.7 & 85.8 & 74.4 & 83.4 & 77.7 & 83.4 & 85.3 & 86.0 & 83.2 \\
        Prec.  & 77.8 & 100.0 & 100.0 & 100.0 & 67.9 & 80.2 & 84.7 & 61.7 & 69.0 & 55.9 & 66.2 & 74.4 & 80.0 & 71.1 \\
        Rec.   & 69.8 & 100.0 & 100.0 & 100.0 & 67.9 & 76.4 & 79.2 & 50.1 & 49.4 & 61.5 & 46.9 & 58.2 & 70.1 & 47.2 \\
        \midrule

        \multicolumn{15}{c}{S3DIS Area~5 - no ``stuff''} \\
        Metric & Avg. & ceiling & floor & wall & beam & column & window & door & chair & table & bookcase & sofa & board & clutter \\
        \midrule

        PQ     & 50.1 & 46.9 & 69.5 & 39.0 & 0.0 & 45.7 & 68.1 & 47.9 & 64.2 & 41.1 & 48.6 & 66.2 & 74.8 & 39.1 \\
        RQ     & 60.1 & 52.0 & 78.4 & 49.3 & 0.0 & 58.6 & 80.8 & 60.2 & 75.9 & 53.5 & 57.6 & 81.8 & 85.7 & 47.8 \\
        SQ     & 76.6 & 90.3 & 88.6 & 79.1 & 0.0 & 78.0 & 84.2 & 79.6 & 84.5 & 76.8 & 84.3 & 80.9 & 87.3 & 81.8 \\
        Prec.  & 63.6 & 45.5 & 70.6 & 43.7 & 0.0 & 62.1 & 90.5 & 68.7 & 76.7 & 58.5 & 74.4 & 81.8 & 94.3 & 59.9 \\
        Rec.   & 58.4 & 60.5 & 88.2 & 56.6 & 0.0 & 55.4 & 73.1 & 53.5 & 75.2 & 49.3 & 47.0 & 81.8 & 78.6 & 39.8 \\
        \\\midrule
        
        \multicolumn{15}{c}{S3DIS 6-FOLD - no ``stuff''} \\
        \midrule~\\

        PQ     & 55.9 & 68.6 & 64.1 & 40.0 & 65.6 & 64.0 & 70.1 & 42.7 & 48.0 & 48.3 & 43.7 & 55.4 & 69.4 & 46.7 \\
        RQ     & 66.3 & 74.8 & 72.6 & 50.8 & 74.3 & 76.7 & 81.8 & 57.1 & 57.3 & 62.8 & 52.5 & 64.6 & 80.5 & 56.0 \\
        SQ     & 83.8 & 91.8 & 88.2 & 78.6 & 88.3 & 83.4 & 85.8 & 74.7 & 83.7 & 76.9 & 83.2 & 85.7 & 86.2 & 83.4 \\
        Prec.  & 72.8 & 76.4 & 69.8 & 50.2 & 77.9 & 78.3 & 86.7 & 68.8 & 70.7 & 63.9 & 67.0 & 72.7 & 90.8 & 73.1 \\
        Rec.   & 61.7 & 73.3 & 75.7 & 51.4 & 71.1 & 75.2 & 77.4 & 48.8 & 48.2 & 61.8 & 43.1 & 58.2 & 72.3 & 45.4 \\
        \midrule
    \end{tabular}
    
}\end{center}
\end{table*}

\begin{table*}[t]
\caption{{\bf ScanNetv2 Val. Class-wise Performance.} We report the average and per-class panoptic quality (PQ), recognition quality (RQ), segmentation quality (SQ), precision (Prec), and recall (Rec) performance of \SHORTHAND on ScanNet. 
We indicate ``stuff'' classes with $\dagger$.}
\label{tab:classwise_scannet} 
\begin{center}
\resizebox{\linewidth}{!}{
    \begin{tabular}{lc*{21}{c}}
        \toprule
        Metric & Avg. & \rotatebox{90}{wall $\dagger$} & \rotatebox{90}{floor $\dagger$} & \rotatebox{90}{cabinet} & \rotatebox{90}{bed} & \rotatebox{90}{chair} & \rotatebox{90}{sofa} & \rotatebox{90}{table} & \rotatebox{90}{door} & \rotatebox{90}{window} & \rotatebox{90}{bookshelf} & \rotatebox{90}{picture} & \rotatebox{90}{counter} & \rotatebox{90}{desk} & \rotatebox{90}{curtain} & \rotatebox{90}{refrigerator} & \rotatebox{90}{shower} & \rotatebox{90}{toilet} & \rotatebox{90}{sink} & \rotatebox{90}{bathtub} & \rotatebox{90}{otherfurniture} \\
        \midrule

        PQ     & \normalsize{58.7} & \normalsize{73.3} & \normalsize{91.8} & \normalsize{50.5} & \normalsize{70.3} & \normalsize{61.3} & \normalsize{69.0} & \normalsize{58.9} & \normalsize{42.5} & \normalsize{44.5} & \normalsize{65.9} & \normalsize{27.7} & \normalsize{42.9} & \normalsize{49.6} & \normalsize{40.9} & \normalsize{64.1} & \normalsize{72.0} & \normalsize{88.6} & \normalsize{51.0} & \normalsize{61.7} & \normalsize{46.7} \\
        RQ     & \normalsize{69.1} & \normalsize{88.7} & \normalsize{99.3} & \normalsize{61.9} & \normalsize{77.9} & \normalsize{70.7} & \normalsize{79.2} & \normalsize{68.9} & \normalsize{52.9} & \normalsize{54.6} & \normalsize{72.1} & \normalsize{34.7} & \normalsize{58.4} & \normalsize{62.6} & \normalsize{49.6} & \normalsize{71.7} & \normalsize{84.2} & \normalsize{99.2} & \normalsize{64.5} & \normalsize{75.4} & \normalsize{56.0} \\
        SQ     & \normalsize{84.1} & \normalsize{82.6} & \normalsize{92.4} & \normalsize{81.6} & \normalsize{90.2} & \normalsize{86.8} & \normalsize{87.2} & \normalsize{85.5} & \normalsize{80.3} & \normalsize{81.5} & \normalsize{91.4} & \normalsize{79.7} & \normalsize{73.5} & \normalsize{79.1} & \normalsize{82.6} & \normalsize{89.4} & \normalsize{85.5} & \normalsize{89.3} & \normalsize{79.1} & \normalsize{81.8} & \normalsize{83.4} \\
        Prec.  & \normalsize{76.7} & \normalsize{93.1} & \normalsize{100.0} & \normalsize{69.7} & \normalsize{81.1} & \normalsize{79.0} & \normalsize{80.0} & \normalsize{68.0} & \normalsize{65.9} & \normalsize{63.4} & \normalsize{75.7} & \normalsize{64.9} & \normalsize{68.4} & \normalsize{60.1} & \normalsize{58.0} & \normalsize{94.3} & \normalsize{82.8} & \normalsize{98.3} & \normalsize{86.0} & \normalsize{76.7} & \normalsize{69.5} \\
        Rec.   & \normalsize{64.3} & \normalsize{84.6} & \normalsize{98.7} & \normalsize{55.7} & \normalsize{75.0} & \normalsize{63.9} & \normalsize{78.4} & \normalsize{69.9} & \normalsize{44.2} & \normalsize{48.0} & \normalsize{68.8} & \normalsize{23.7} & \normalsize{51.0} & \normalsize{65.4} & \normalsize{43.3} & \normalsize{57.9} & \normalsize{85.7} & \normalsize{100.0} & \normalsize{51.6} & \normalsize{74.2} & \normalsize{46.8} \\
        \midrule
    \end{tabular}
}
\end{center}
\end{table*}

\begin{table*}[t]
\caption{{\bf KITTI-360 Val. Class-wise Performance.} We report the average and per-class panoptic quality (PQ), recognition quality (RQ), segmentation quality (SQ), precision (Prec), and recall (Rec) performance of \SHORTHAND on the KITTI-360 Validation set. 
We indicate ``stuff'' classes with $\dagger$.}
\label{tab:classwise_kitti360} 
\begin{center}
\footnotesize{

    \begin{tabular}{lc*{16}{c}}
        \toprule
        Metric & Avg. & \rotatebox{90}{road $\dagger$} & \rotatebox{90}{sidewalk $\dagger$} & \rotatebox{90}{building} & \rotatebox{90}{wall $\dagger$} & \rotatebox{90}{fence $\dagger$} & \rotatebox{90}{pole $\dagger$} & \rotatebox{90}{traffic lig. $\dagger$} & \rotatebox{90}{traffic sig. $\dagger$} & \rotatebox{90}{vegetation $\dagger$} & \rotatebox{90}{terrain $\dagger$} & \rotatebox{90}{person $\dagger$} & \rotatebox{90}{car} & \rotatebox{90}{truck $\dagger$} & \rotatebox{90}{motorcycle $\dagger$} & \rotatebox{90}{bicycle $\dagger$} \\
        \midrule

        PQ     & 48.3 & 94.3 & 75.6 & 44.8 & 31.5 & 20.0 & 43.4 & 0.0 & 30.5 & 89.5 & 49.5 & 19.4 & 84.2 & 81.5 & 55.5 & 5.4 \\
        RQ     & 58.4 & 100.0 & 95.3 & 53.4 & 51.4 & 33.3 & 65.6 & 0.0 & 40.8 & 100.0 & 68.5 & 20.7 & 89.1 & 83.3 & 66.7 & 7.1 \\
        SQ     & 75.1 & 94.3 & 79.3 & 83.9 & 61.2 & 60.0 & 66.2 & 0.0 & 74.6 & 89.5 & 72.3 & 93.6 & 94.4 & 97.8 & 83.3 & 75.5 \\
        Prec.  & 60.3 & 100.0 & 96.2 & 48.4 & 51.9 & 33.3 & 65.6 & 0.0 & 43.5 & 100.0 & 76.0 & 23.1 & 92.4 & 90.9 & 73.3 & 10.0 \\
        Rec.   & 56.9 & 100.0 & 94.4 & 59.5 & 50.9 & 33.3 & 65.6 & 0.0 & 38.5 & 100.0 & 62.3 & 18.8 & 86.1 & 76.9 & 61.1 & 5.6 \\
        \midrule
    \end{tabular}
    
}\end{center}
\end{table*}

\begin{table*}[t]
\caption{{\bf DALES Class-wise Performance.} We report the average and per-class panoptic quality (PQ), recognition quality (RQ), segmentation quality (SQ), precision (Prec), and recall (Rec) performance of \SHORTHAND on DALES. 
We indicate ``stuff'' classes with $\dagger$.}
\label{tab:classwise_dales} 
\begin{center}
\footnotesize{
   
    \begin{tabular}{lc*{10}{c}}
        \toprule
        Metric & Avg. & ground $\dagger$ & vegetation $\dagger$ & car & truck & power line & fence & pole & building \\
        \midrule

        PQ     & 61.2 & 95.6 & 90.3 & 70.9 & 45.0 & 18.8 & 23.5 & 64.3 & 81.5 \\
        RQ     & 68.6 & 100.0 & 99.0 & 78.4 & 51.1 & 23.1 & 31.3 & 79.6 & 86.6 \\
        SQ     & 87.1 & 95.6 & 91.2 & 90.4 & 88.2 & 81.3 & 75.0 & 80.8 & 94.1 \\
        Prec.  & 68.5 & 100.0 & 99.0 & 87.3 & 55.1 & 16.3 & 24.2 & 81.5 & 84.7 \\
        Rec.   & 71.0 & 100.0 & 99.0 & 71.1 & 47.5 & 39.7 & 44.3 & 77.8 & 88.5 \\
        \midrule
    \end{tabular}\\
    
}\end{center}
\end{table*}

We report in \tabref{tab:classwise_s3dis}, \tabref{tab:classwise_scannet}, \tabref{tab:classwise_kitti360}, and \tabref{tab:classwise_dales} the average and per-class performances of \SHORTHAND on each dataset.

\balance

\begin{figure*}[!b]
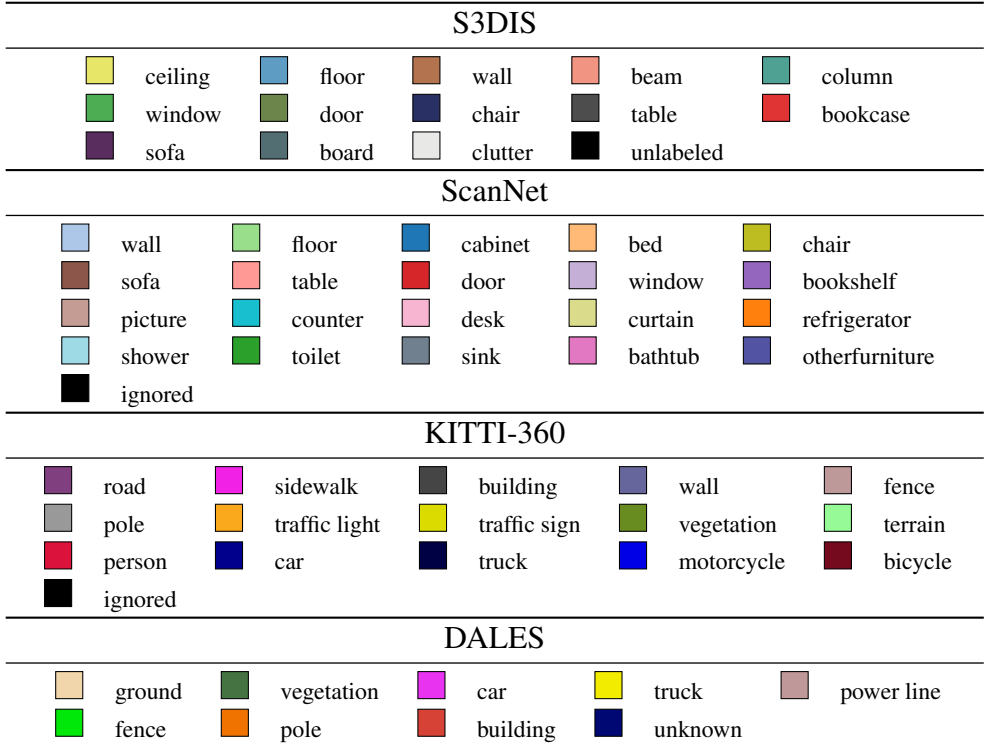

    \centering
    \begin{tabular}{c}
    
        \toprule
        \large{S3DIS}
        \\\midrule
        \begin{tabular}{@{}rlrlrlrlrl@{}}
            \definecolor{tempcolor}{rgb}{0.91,0.90,0.41}
            \tikz \fill[fill=tempcolor, scale=0.3, draw=black] (0, 0) rectangle (1.2, 1.2);
            & \small{ceiling} 
            &
            \definecolor{tempcolor}{rgb}{.37,0.61,0.77}
            \tikz \fill[fill=tempcolor, scale=0.3, draw=black] (0, 0) rectangle (1.2, 1.2); 
            & \small{floor}
            &
            \definecolor{tempcolor}{rgb}{0.70,0.45,0.31}
            \tikz \fill[fill=tempcolor, scale=0.3, draw=black] (0, 0) rectangle (1.2, 1.2); 
            & \small{wall}
            &
            \definecolor{tempcolor}{rgb}{0.95,.58,0.51}
            \tikz \fill[fill=tempcolor, scale=0.3, draw=black] (0, 0) rectangle (1.2, 1.2);
            & \small{beam} &
            \definecolor{tempcolor}{rgb}{0.31,0.63,.58}
            \tikz \fill[fill=tempcolor, scale=0.3, draw=black] (0, 0) rectangle (1.2, 1.2);
            & \small{column} 
            \\
            \definecolor{tempcolor}{rgb}{0.30,0.68,.32}
            \tikz \fill[fill=tempcolor, scale=0.3, draw=black] (0, 0) rectangle (1.2, 1.2); 
            & \small{window}
            &
            \definecolor{tempcolor}{rgb}{.42,0.52,0.29}
            \tikz \fill[fill=tempcolor, scale=0.3, draw=black] (0, 0) rectangle (1.2, 1.2); 
            & \small{door}
            &
            \definecolor{tempcolor}{rgb}{.16,0.19,0.39}
            \tikz \fill[fill=tempcolor, scale=0.3, draw=black] (0, 0) rectangle (1.2, 1.2);
            & \small{chair}
            &
            \definecolor{tempcolor}{rgb}{.30,0.30,0.30}
            \tikz \fill[fill=tempcolor, scale=0.3, draw=black] (0, 0) rectangle (1.2, 1.2);
            & \small{table} 
            &
            \definecolor{tempcolor}{rgb}{.88,0.20,0.20}
            \tikz \fill[fill=tempcolor, scale=0.3, draw=black] (0, 0) rectangle (1.2, 1.2); 
            & \small{bookcase}
            \\
            \definecolor{tempcolor}{rgb}{0.35,.18,0.37}
            \tikz \fill[fill=tempcolor, scale=0.3, draw=black] (0, 0) rectangle (1.2, 1.2); 
            & \small{sofa}
            &
            \definecolor{tempcolor}{rgb}{.32,0.43,.45}
            \tikz \fill[fill=tempcolor, scale=0.3, draw=black] (0, 0) rectangle (1.2, 1.2);
            & \small{board}
            &
            \definecolor{tempcolor}{rgb}{0.91,0.91,.90}
            \tikz \fill[fill=tempcolor, scale=0.3, draw=black] (0, 0) rectangle (1.2, 1.2);
            & \small{clutter} 
            &
            \definecolor{tempcolor}{rgb}{0.,0.,0}
            \tikz \fill[fill=tempcolor, scale=0.3, draw=black] (0, 0) rectangle (1.2, 1.2);
            & \small{unlabeled} 
        \end{tabular}
        \\

        \toprule
        \large{ScanNet}
        \\\midrule
        \begin{tabular}{rlrlrlrlrl}
            \definecolor{tempcolor}{rgb}{0.68, 0.78, 0.91}
            \tikz \fill[fill=tempcolor, scale=0.3, draw=black] (0, 0) rectangle (1.2, 1.2);
            & \small{wall} 
            &
            \definecolor{tempcolor}{rgb}{0.6 , 0.87, 0.54}
            \tikz \fill[fill=tempcolor, scale=0.3, draw=black] (0, 0) rectangle (1.2, 1.2);
            & \small{floor} 
            &
            \definecolor{tempcolor}{rgb}{0.12, 0.47, 0.71}
            \tikz \fill[fill=tempcolor, scale=0.3, draw=black] (0, 0) rectangle (1.2, 1.2);
            & \small{cabinet} 
            &
            \definecolor{tempcolor}{rgb}{1.  , 0.73, 0.47}
            \tikz \fill[fill=tempcolor, scale=0.3, draw=black] (0, 0) rectangle (1.2, 1.2);
            & \small{bed} 
            &
            \definecolor{tempcolor}{rgb}{0.74, 0.74, 0.13}
            \tikz \fill[fill=tempcolor, scale=0.3, draw=black] (0, 0) rectangle (1.2, 1.2);
            & \small{chair} 
            \\
            \definecolor{tempcolor}{rgb}{0.55, 0.34, 0.29}
            \tikz \fill[fill=tempcolor, scale=0.3, draw=black] (0, 0) rectangle (1.2, 1.2);
            & \small{sofa} 
            &
            \definecolor{tempcolor}{rgb}{1.  , 0.6 , 0.59}
            \tikz \fill[fill=tempcolor, scale=0.3, draw=black] (0, 0) rectangle (1.2, 1.2);
            & \small{table} 
            &
            \definecolor{tempcolor}{rgb}{0.84, 0.15, 0.16}
            \tikz \fill[fill=tempcolor, scale=0.3, draw=black] (0, 0) rectangle (1.2, 1.2);
            & \small{door} 
            &
            \definecolor{tempcolor}{rgb}{0.77, 0.69, 0.84}
            \tikz \fill[fill=tempcolor, scale=0.3, draw=black] (0, 0) rectangle (1.2, 1.2);
            & \small{window} 
            &
            \definecolor{tempcolor}{rgb}{0.58, 0.4 , 0.74}
            \tikz \fill[fill=tempcolor, scale=0.3, draw=black] (0, 0) rectangle (1.2, 1.2);
            & \small{bookshelf} 
            \\
            \definecolor{tempcolor}{rgb}{0.77, 0.61, 0.58}
            \tikz \fill[fill=tempcolor, scale=0.3, draw=black] (0, 0) rectangle (1.2, 1.2);
            & \small{picture} 
            &
            \definecolor{tempcolor}{rgb}{0.09, 0.75, 0.81}
            \tikz \fill[fill=tempcolor, scale=0.3, draw=black] (0, 0) rectangle (1.2, 1.2);
            & \small{counter} 
            &
            \definecolor{tempcolor}{rgb}{0.97, 0.71, 0.82}
            \tikz \fill[fill=tempcolor, scale=0.3, draw=black] (0, 0) rectangle (1.2, 1.2);
            & \small{desk} 
            &
            \definecolor{tempcolor}{rgb}{0.86, 0.86, 0.55}
            \tikz \fill[fill=tempcolor, scale=0.3, draw=black] (0, 0) rectangle (1.2, 1.2);
            & \small{curtain} 
            &
            \definecolor{tempcolor}{rgb}{1.  , 0.5 , 0.05}
            \tikz \fill[fill=tempcolor, scale=0.3, draw=black] (0, 0) rectangle (1.2, 1.2);
            & \small{refrigerator} 
            \\
            \definecolor{tempcolor}{rgb}{0.62, 0.85, 0.9 }
            \tikz \fill[fill=tempcolor, scale=0.3, draw=black] (0, 0) rectangle (1.2, 1.2);
            & \small{shower} 
            &
            \definecolor{tempcolor}{rgb}{0.17, 0.63, 0.17}
            \tikz \fill[fill=tempcolor, scale=0.3, draw=black] (0, 0) rectangle (1.2, 1.2);
            & \small{toilet} 
            &
            \definecolor{tempcolor}{rgb}{0.44, 0.5 , 0.56}
            \tikz \fill[fill=tempcolor, scale=0.3, draw=black] (0, 0) rectangle (1.2, 1.2);
            & \small{sink} 
            &
            \definecolor{tempcolor}{rgb}{0.89, 0.47, 0.76}
            \tikz \fill[fill=tempcolor, scale=0.3, draw=black] (0, 0) rectangle (1.2, 1.2);
            & \small{bathtub} 
            &
            \definecolor{tempcolor}{rgb}{0.32, 0.33, 0.64}
            \tikz \fill[fill=tempcolor, scale=0.3, draw=black] (0, 0) rectangle (1.2, 1.2);
            & \small{otherfurniture} 
            \\
            \definecolor{tempcolor}{rgb}{0.  , 0.  , 0.  }
            \tikz \fill[fill=tempcolor, scale=0.3, draw=black] (0, 0) rectangle (1.2, 1.2);
            & \small{ignored} 
            &
        \end{tabular}\\
        
        \toprule
        \large{KITTI-360}
        \\\midrule
        \begin{tabular}{rlrlrlrlrl}
            \definecolor{tempcolor}{rgb}{0.50, 0.25, 0.50}
            \tikz \fill[fill=tempcolor, scale=0.3, draw=black] (0, 0) rectangle (1.2, 1.2);
            & \small{road} 
            &
            \definecolor{tempcolor}{rgb}{0.95, 0.13, 0.90}
            \tikz \fill[fill=tempcolor, scale=0.3, draw=black] (0, 0) rectangle (1.2, 1.2);
            & \small{sidewalk} 
            &
            \definecolor{tempcolor}{rgb}{0.27, 0.27, 0.27}
            \tikz \fill[fill=tempcolor, scale=0.3, draw=black] (0, 0) rectangle (1.2, 1.2);
            & \small{building} 
            &
            \definecolor{tempcolor}{rgb}{0.4 , 0.4 , 0.61}
            \tikz \fill[fill=tempcolor, scale=0.3, draw=black] (0, 0) rectangle (1.2, 1.2);
            & \small{wall} 
            &
            \definecolor{tempcolor}{rgb}{0.74, 0.6 , 0.6 }
            \tikz \fill[fill=tempcolor, scale=0.3, draw=black] (0, 0) rectangle (1.2, 1.2);
            & \small{fence} 
            \\
            \definecolor{tempcolor}{rgb}{0.6 , 0.6 , 0.6 }
            \tikz \fill[fill=tempcolor, scale=0.3, draw=black] (0, 0) rectangle (1.2, 1.2);
            & \small{pole} 
            &
            \definecolor{tempcolor}{rgb}{0.98, 0.66, 0.11}
            \tikz \fill[fill=tempcolor, scale=0.3, draw=black] (0, 0) rectangle (1.2, 1.2);
            & \small{traffic light} 
            &
            \definecolor{tempcolor}{rgb}{0.86, 0.86, 0.  }
            \tikz \fill[fill=tempcolor, scale=0.3, draw=black] (0, 0) rectangle (1.2, 1.2);
            & \small{traffic sign} 
            &
            \definecolor{tempcolor}{rgb}{0.41, 0.55, 0.13}
            \tikz \fill[fill=tempcolor, scale=0.3, draw=black] (0, 0) rectangle (1.2, 1.2);
            & \small{vegetation} 
            &
            \definecolor{tempcolor}{rgb}{0.59, 0.98, 0.59}
            \tikz \fill[fill=tempcolor, scale=0.3, draw=black] (0, 0) rectangle (1.2, 1.2);
            & \small{terrain} 
            \\
            \definecolor{tempcolor}{rgb}{0.86, 0.07, 0.23}
            \tikz \fill[fill=tempcolor, scale=0.3, draw=black] (0, 0) rectangle (1.2, 1.2);
            & \small{person} 
            &
            \definecolor{tempcolor}{rgb}{0.  , 0.  , 0.55}
            \tikz \fill[fill=tempcolor, scale=0.3, draw=black] (0, 0) rectangle (1.2, 1.2);
            & \small{car} 
            &
            \definecolor{tempcolor}{rgb}{0.  , 0.  , 0.27}
            \tikz \fill[fill=tempcolor, scale=0.3, draw=black] (0, 0) rectangle (1.2, 1.2);
            & \small{truck} 
            &
            \definecolor{tempcolor}{rgb}{0.  , 0.  , 0.90}
            \tikz \fill[fill=tempcolor, scale=0.3, draw=black] (0, 0) rectangle (1.2, 1.2);
            & \small{motorcycle} 
            &
            \definecolor{tempcolor}{rgb}{0.46, 0.04, 0.12}
            \tikz \fill[fill=tempcolor, scale=0.3, draw=black] (0, 0) rectangle (1.2, 1.2);
            & \small{bicycle} 
            \\
            \definecolor{tempcolor}{rgb}{0.  , 0.  , 0.  }
            \tikz \fill[fill=tempcolor, scale=0.3, draw=black] (0, 0) rectangle (1.2, 1.2);
            & \small{ignored} 
            &
        \end{tabular}\\
        
        \toprule
        \large{DALES}
        \\\midrule
        \begin{tabular}{rlrlrlrlrl}
            \definecolor{tempcolor}{rgb}{0.95, 0.84, 0.67}
            \tikz \fill[fill=tempcolor, scale=0.3, draw=black] (0, 0) rectangle (1.2, 1.2);
            & \small{ground} 
            &
            \definecolor{tempcolor}{rgb}{0.27, 0.45, 0.26}
            \tikz \fill[fill=tempcolor, scale=0.3, draw=black] (0, 0) rectangle (1.2, 1.2); 
            & \small{vegetation}
            &
            \definecolor{tempcolor}{rgb}{0.91, 0.20, 0.94}
            \tikz \fill[fill=tempcolor, scale=0.3, draw=black] (0, 0) rectangle (1.2, 1.2); 
            & \small{car}
            &
            \definecolor{tempcolor}{rgb}{0.95, 0.93, 0.}
            \tikz \fill[fill=tempcolor, scale=0.3, draw=black] (0, 0) rectangle (1.2, 1.2);
            & \small{truck} &
            \definecolor{tempcolor}{rgb}{0.75, 0.6,0.6  }
            \tikz \fill[fill=tempcolor, scale=0.3, draw=black] (0, 0) rectangle (1.2, 1.2);
            & \small{power line} 
            \\
            \definecolor{tempcolor}{rgb}{0., 0.91,  0.04}
            \tikz \fill[fill=tempcolor, scale=0.3, draw=black] (0, 0) rectangle (1.2, 1.2); 
            & \small{fence}
            &
            \definecolor{tempcolor}{rgb}{0.94,  0.45, 0.}
            \tikz \fill[fill=tempcolor, scale=0.3, draw=black] (0, 0) rectangle (1.2, 1.2); 
            & \small{pole}
            &
            \definecolor{tempcolor}{rgb}{0.84, 0.26, 0.21}
            \tikz \fill[fill=tempcolor, scale=0.3, draw=black] (0, 0) rectangle (1.2, 1.2);
            & \small{building}
            &
            \definecolor{tempcolor}{rgb}{0., 0.03, 0.45}
            \tikz \fill[fill=tempcolor, scale=0.3, draw=black] (0, 0) rectangle (1.2, 1.2);
            & \small{unknown}
            
        \end{tabular}\\
        \midrule~\\
    \end{tabular}
    \caption{{\bf Colormaps.} Throughout all visualization in the main paper, the appendix, and the interactive visualization, we use this colormaps to represent the semantic of each point. }
    \label{fig:colormaps}
\end{figure*}

}{}

\end{document}